\documentclass[letterpaper]{article} % DO NOT CHANGE THIS
\usepackage{aaai24}  % DO NOT CHANGE THIS
\usepackage{amsmath,amsthm,amssymb}
\usepackage{times}  % DO NOT CHANGE THIS
\usepackage{helvet}  % DO NOT CHANGE THIS
\usepackage{courier}  % DO NOT CHANGE THIS
\usepackage[hyphens]{url}  % DO NOT CHANGE THIS
\usepackage{graphicx} % DO NOT CHANGE THIS
\usepackage{natbib}  % DO NOT CHANGE THIS AND DO NOT ADD ANY OPTIONS TO IT
\usepackage{caption} % DO NOT CHANGE THIS AND DO NOT ADD ANY OPTIONS TO IT
\usepackage[utf8]{inputenc} % allow utf-8 input
\usepackage[T1]{fontenc}    % use 8-bit T1 fonts
\usepackage{url}            % simple URL typesetting
\usepackage{booktabs}       % professional-quality tables
\usepackage{amsfonts}       % blackboard math symbols
\usepackage{nicefrac}       % compact symbols for 1/2, etc.
\usepackage{microtype}      % microtypography
\usepackage{xcolor}         % colors
\usepackage{resizegather}

\usepackage{algorithm}
\usepackage{newfloat}
\usepackage{listings}

\usepackage{multirow}
\usepackage[algo2e]{algorithm2e}

\usepackage{cancel}
\usepackage[noend]{algpseudocode}
\usepackage{enumerate}
\usepackage{mathtools, cuted}

\def\ci{\perp\!\!\!\perp}
 
\newcommand{\X}{X}
\newcommand{\Y}{Y}
\newcommand{\Z}{Z}
\newcommand{\V}{W}
\newcommand{\x}{x}
\newcommand{\y}{y}
\newcommand{\z}{z}
\newcommand{\vpoint}{w}
\newcommand{\Xspace}{\mathcal{X}}
\newcommand{\Yspace}{\mathcal{Y}}
\newcommand{\Zspace}{\mathcal{Z}}
\newcommand{\Vspace}{\mathcal{W}}
\newcommand{\mX}{m_X}
\newcommand{\mY}{m_Y}
\newcommand{\mZ}{m_Z}
\newcommand{\mV}{m_W}
\newcommand{\PXYZ}{P_{XYZ}}
\newcommand{\PXYconditionalZ}{P_{XY|Z}}
\newcommand{\PXconditionalZ}{P_{X|Z}}
\newcommand{\PYconditionalZ}{P_{Y|Z}}

\newcommand{\pmf}{p}
\newcommand{\pdf}{f}
\newcommand{\OurEstimator}{MS$_{0-\infty}$}

\theoremstyle{plain}
\newtheorem{mythm}{Theorem}

\newtheorem{mylemma}[mythm]{Lemma}

\newtheorem{myassump}[mythm]{Assumption}

\newcommand{\MSEstimatorMaxNullWithPrime}{\hat{I}^{MS}(\X^\prime;\Y^\prime|\Z^\prime)}

\newcommand{\OursEstimatorMaxNull}{\hat{I}^{0-\infty}(\X;\Y|\Z)}

\newcommand{\nLimit}{\lim_{n\to\infty}}
\newcommand{\E}[1]{\mathbb{E}\left[#1\right]}

\DeclareCaptionStyle{ruled}{labelfont=normalfont,labelsep=colon,strut=off} % DO NOT CHANGE THIS
\floatstyle{ruled}
\newfloat{listing}{tb}{lst}{}
\floatname{listing}{Listing}
\title{Non-parametric Conditional Independence Testing for Mixed Continuous-Categorical Variables: A Novel Method and Numerical Evaluation}
\author{
    Oana-Iuliana Popescu\textsuperscript{\rm 1},
    Andreas Gerhardus\textsuperscript{\rm 1},
     Jakob Runge\textsuperscript{\rm 1,2}
}
\affiliations{
    \textsuperscript{\rm 1}Institute of Data Science, German Aerospace Center,
     \textsuperscript{\rm 2}Technische Universität Berlin
}

\begin{document}

\maketitle

\begin{abstract}
Conditional independence testing (CIT) is a common task in machine learning, e.g.~for variable selection, and a main component of constraint-based causal discovery. While most current CIT approaches assume that all variables are numerical or all variables are categorical, many real-world applications involve mixed-type datasets that include numerical and categorical variables. Non-parametric CIT can be conducted using conditional mutual information (CMI) estimators combined with a local permutation scheme. Recently, two novel CMI estimators for mixed-type datasets based on k-nearest-neighbors (k-NN) have been proposed. As with any k-NN method, these estimators rely on the definition of a distance metric. One approach computes distances by a one-hot encoding of the categorical variables, essentially treating categorical variables as discrete-numerical, while the other expresses CMI by entropy terms where the categorical variables appear as conditions only. In this work, we study these estimators and propose a variation of the former approach that does not treat categorical variables as numeric. Our numerical experiments show that our variant detects dependencies more robustly across different data distributions and preprocessing types.
\end{abstract}

\section{Introduction}\label{sec:intro}

Conditional independence testing (CIT) is a central component of constraint-based causal discovery frameworks; e.g., in algorithms such as PC and FCI \citep{spirtes2000causation}, and is used to infer causal relations from purely observational data. The performance of causal discovery algorithms depends heavily on the performance of the CIT and its robustness toward different types, distributions, and sample sizes of the data. A good CIT approach achieves high statistical power to detect true conditional dependence while simultaneously controlling false positives at the desired level. Most current CIT approaches assume that either all variables are numerical or that all variables are categorical. Still, many real-world applications involve mixed-type datasets, e.g., datasets with variables such as gender and height in medicine or weather regime types and continuous temperature in climate science.

In this work, we consider the two recent $k$-NN estimators of CMI for mixed-type data. The estimator of \citet{mesner_shalizi} transforms categorical variables by one-hot encoding and then measures distances on the resulting product space of mixed continuous-discrete variables. \citet{Zan2022ACM} frame mixed-type datasets as consisting of quantitative and qualitative variables and rewrite the CMI as a linear combination of entropies where the qualitative variables appear as conditions only, thus requiring distance notions on the quantitative subspaces. CMI is of relevance for non-parametric CIT since the conditional independence $\X \ci \Y ~|~ \Z$ holds if and only if $I(\X;\Y|\Z) = 0$ \cite{gray2011entropy} and thus is a non-parametric measure for conditional independence. To construct CIT, CMI estimators can be combined with a local permutation scheme (similar to \citet{runge2018conditional}) and be formulated as a statistical test of the null hypothesis $H_0: \, \X \ci \Y \mid \Z$ as done in \citet{Zan2022ACM}. No general analytical results for the finite or asymptotic distribution of CMI under conditional independence are known. 

We study CIT based on the previously mentioned estimators from an empirical perspective. We first outline the challenges these estimators face, some of which have been discussed in the respective works, for example, the curse of dimensionality that leads to increased bias. We briefly discuss why the CMI estimation performance can affect the outcomes of the corresponding CIT. We investigate how the two estimators and their respective CITs perform under different choices of hyperparameters, data distributions, and combinations of variable types and dimensionalities. 
To reduce the effect of the challenges that the two estimators face on CIT, we propose a variant based on the estimator of \citet{mesner_shalizi} that does not rely on one-hot encoding of categorical variables.
In summary, our main contributions are (1) a new $k$-NN estimator for the CMI of mixed-type data that is a variant of the estimator of \citet{mesner_shalizi}, (2) an empirical evaluation of the three CMI estimators, and (3) an extensive and systematic numerical evaluation of CIT performance based on the three CMI estimators in combination with a local permutation scheme.

% The choice of hyperparameters is critical as, e.g., in causal discovery, hyperparameter tuning is not easily possible (as there is no validation set with known ground truth). 

\section{Background and related work}

After the preliminaries, we give an extensive summary of related work in order to highlight the subtle differences between the different estimators and to build upon further below.

\subsection{Preliminaries}
Let $\X: \Omega \to \Xspace$, $\Y: \Omega \to \Yspace$ and $\Z: \Omega \to \mathcal{Z}$ be (vectors of) random variables with $dim(\Xspace)=\mX$, $dim(\Yspace)=\mY$, and $dim(\Zspace)=\mZ$. We demand that $\Xspace = \Xspace_1 \times \ldots \Xspace_{\mX}$ where $\Xspace_a$ with $1 \leq a \leq \mX$ is $\mathbb R$ or a discrete set; similarly for $\Yspace$ and $\Zspace$. Let $\PXYZ$ be the probability measure on $\Xspace \times \Yspace \times \Zspace$ induced by the joint vector $(\X, \Y, \Z)$. We assume that the conditional probability measure $\PXYconditionalZ$ exists and is absolutely continuous wrt to the product measure $\PXconditionalZ \times \PYconditionalZ$. As discussed in \citet{mesner_shalizi}, these assumptions are fulfilled if every component of $(\X, \Y, \Z)$ is either discrete (i.e., absolutely continuous wrt to the counting measure) or non-singular continuous (i.e., absolutely continuous wrt to the Lebesgue measure) or a mixture of these two cases (for simplicity, from here on we refer to ``non-singular continuous'' as ``continuous''). The CMI $I(\X;\Y|\Z)$ of $\X$ and $\Y$ given $\Z$ can be defined as, see \citet{gray2011entropy},
\begin{equation}\label{eq:CMI-general}
I(\X;\Y|\Z) = \int \log \left(\frac{d\PXYconditionalZ}{d(\PXconditionalZ \times \PYconditionalZ)}\right) d\PXYconditionalZ \, ,
\end{equation} 
where the argument of the $\log$ is the Radon-Nikodym derivative of $\PXYconditionalZ$ wrt $\PXconditionalZ \times \PYconditionalZ$. If all components of $(\X, \Y, \Z)$ are discrete, then the rhs of eq.~\eqref{eq:CMI-general} reduces to the familiar form
\begin{equation}\label{eq:CMI-discrete}
\sum_{\x,\y,\z} \pmf_{\X\Y\Z}(\x,\y,\z)\log \frac{\pmf_{\X\Y|\Z}(\x,\y|\z)}{\pmf_{\X|\Z}(\x|\z)\pmf_{\Y|\Z}(\y|\z)} 
\end{equation}
in terms of the probability mass functions (pmfs) $\pmf_{\cdot}$. If all components are continuous, then in eq.~\eqref{eq:CMI-discrete} integrals replace sums and probability density functions (pdfs) $\pdf_{\cdot}$ replace pmfs.

We distinguish three types of discrete random variables $V$: First, the values of $V$ can be a discrete subset of $\mathbb R$ such that distance notions on $\mathbb R$ (e.g., $L^p$-distance) other than the discrete metric are semantically meaningful (``discrete numeric''). Second and third, the values of $V$ can be on an ordinal or a nominal/categorical scale (``non-numeric''). From a measure-theory perspective, these cases are equivalent. When speaking of ``mixed continuous-categorical variables'' (in short: ``mixed variables''), we refer to any of the following cases: First, all of $\X$, $\Y$, $\Z$ are either fully discrete or fully continuous, with at least one of them fully discrete and another fully continuous. Second, at least one of $\X$, $\Y$, $\Z$ contains both a discrete and a continuous component, but no component of $(\X, \Y, \Z)$ is a mixture variable. Third, at least one component $V$ of $(\X, \Y, \Z)$ is a mixture variable; that is, this component $V$ itself is neither discrete nor continuous. In this paper, we focus on the first two of these cases.

\subsection{CMI estimation in the fully continuous case}

\textbf{KL estimator for differential entropy} Let $\V: \Omega \to \Vspace$ be a (vector of) continuous random variables with $\Vspace = \mathbb R^{\mV}$ and let $\vpoint_1, \, \ldots, \vpoint_n$ be \emph{iid} observations of $\V$. The \citet{kozachenko_leonenko_1987} (KL) estimator of the differential entropy $H(\V)$ is the sample average
\begin{equation}
\label{eq:kl-estimator-raw}
    \hat{H}^{KL}(\V) = -\frac{1}{n} \sum_{i\,=\,1}^n \log \hat{\pdf}_{\V}(\vpoint_i).
\end{equation}
The local density estimates$\hat{\pdf}_{\V}(\vpoint_i)$ are calculated under the assumption that $\pdf_{\V}$ is locally constant within an $L^p$-ball $B(\vpoint_i, \rho_i)$ of radius $\rho_i$ around $\vpoint_i$ where $\rho_i$ is the $L^p$-distance of $\vpoint_i$ to its $k$-th nearest neighbor (not counting $\vpoint_i$ itself) for some positive integer $k$. Since $\V$ is continuous, this $k$-th nearest neighbor is unique with probability one. The local constancy assumption implies that the probability $P_i$ of the event $\vpoint \in B(\vpoint_i, \rho_i)$ is $P_i \equiv p_{\vpoint}(\vpoint_i) \cdot V_{\mV, p} \cdot \rho_i^{\mV}$, where $V_{\mV, p}$ is the volume of the unit-ball in the $L^p$-metric. Using that $\mathbb E[\log P_i] = \psi(k) - \psi(n)$ with the digamma function $\psi(x)$, see \citet{Kraskov2004EstimatingMI}, and approximating $E[\log \rho_i]$ by a sample average, eq.~\eqref{eq:kl-estimator-raw} then takes the form
\begin{equation} 
\label{eq:kl_estimator}
    \hat{H}^{KL}(\V) = \psi(n) - \psi(k) + \log V_{\mV, p} + \frac{\mV}{n} \sum_{i\,=\,1}^n \log \rho_i \, .
\end{equation} 

\textbf{KSG estimator for mutual information}\citet{Kraskov2004EstimatingMI} estimate the MI $I(\X;\Y) = H(\X) + H(\Y) - H(\X,\Y)$ by estimating the three individual entropies with the KL estimator~\eqref{eq:kl_estimator}. The authors heuristically argue that the errors incurred by the local constancy assumptions approximately cancel out in the combined estimator if all entropy estimates use the same local length scales $\rho_i$. Thus, they equip $\Xspace \times \Yspace$ with the maximum metric $d_{\Xspace \times \Yspace}(\cdot, \cdot) = \max \{ d_\Xspace(\cdot, \cdot), \, d_{\Yspace}(\cdot, \cdot)\}$ and define $\rho_i$ as in the KL estimate~\eqref{eq:kl_estimator} of $H(\X, \Y)$, and use the same radii $\rho_i$ for estimating $H(\X)$ and $H(\Y)$. The estimator takes the form
\begin{multline}\label{eq:mi_ksg-2}
    \hat{I}^{KSG}(\X; \Y) = \frac{1}{n} \sum_{i\,=\,1}^n [ \psi(k) + \psi(n) - \psi(k_{\X,i}+1)-\\- \psi(k_{\Y,i}+1) ] \, ,
\end{multline}
\\
where $k_{\X,i}$ and $k_{\Y,i}$ are defined by ($W$ is placeholder for $\X$ and $\Y$ and $w$ is placeholder for $\x$ and $\y$)
\begin{equation}\label{eq:number-points-in-ball}
k_{W, i} = |\{ w_j ~|~ \Vert w_j - w_i \Vert < \rho_i, \, j \neq i \}| \, ,
\end{equation}
as the number of points $\x_j \neq \x_i$ (resp.~$\y_j \neq \y_i$) within the \emph{open} ball $B(\x_i, \rho_i)$ (resp.~$B(\y_i, \rho_i)$) in $\Xspace$ (resp.~$\Yspace$). The terms with $\log \rho_i$ cancel out due to using the same radii $\rho_i$ in all three entropy estimates. Since $\Xspace \times \Yspace$ is equipped with the maximum metric, the volume terms cancel out too.

\textbf{FP estimator of conditional mutual information}
Using the same rationale, \citep{Frenzel2007PartialMI} extend the KSG estimator to CMI, with $k_{\Z, i}$, $k_{\X\Z,i}, k_{\Y\Z,i}$ as in eq.~\eqref{eq:number-points-in-ball} and $\rho_i$ as in the KL estimate of $H(\X,\Y,\Z)$:
\begin{multline}
\label{eq:fp_cont} 
\hat{I}^{FP}(\X;\Y|\Z) = \frac{1}{n}\sum_{i\,=\,1}^{n} [ \psi(k) + \psi(k_{\Z,i}+1) - \\ -\psi(k_{\X,i}+1) - \psi(k_{\Y,i}+1) ] \, .    
\end{multline}

\subsection{CMI estimation in the mixed variables case}

\textbf{GKOV estimator of mutual information} \citet{gao_17} propose an estimator for mixed MI $I(\X;\Y)$ under the assumption that both $\Xspace$ and $\Yspace$ are Euclidean spaces, thus implicitly requiring that either the discrete values are numeric with a semantically meaningful notion of distance or that they have been mapped to a real space (that is, ignoring the conceptual problem of a semantically non-meaningful $L^p$-distance). The GKOV estimator builds on KSG and the observation that, in the mixed case, the distance $\rho_i$ of $(\x_i, \y_i)$ to its $k$-th nearest neighbor in $\Xspace \times \Yspace$ can be $\rho_i = 0$ with non-zero probability. \citet{gao_17} consider the event $\rho_i = 0$ to indicate that point $(\x_i, \y_i)$ is ``discrete''. Their estimator takes the form
\begin{multline}
\label{eq:gkov} 
    \hat{I}^{GKOV}(\X;\Y) = \frac{1}{n}\sum_{i\,=\,1}^{n} [\psi(\tilde{k}^{\prime}_i) + \log (n) - \\-\log(\tilde{k}_{\X,i}+1) - \log(\tilde{k}_{\Y,i}+1)  ] \, ,
\end{multline}
where $\tilde{k}^{\prime}_i = k$ if $\rho_i > 0$ and $\tilde{k}^{\prime}_i = \tilde{k}_{\X\Y,i}$ if $\rho_i = 0$ with
\begin{equation}
\label{eq:number-points-in-ball-with-boundary}
\tilde{k}_{W, i} = |\{ w_j ~|~ \Vert w_j - w_i \Vert \leq \rho_i, \, j \neq i \}| \, .
\end{equation}
As opposed to eq.~\eqref{eq:number-points-in-ball}, eq.~\eqref{eq:number-points-in-ball-with-boundary} uses the non-strict inequality $\Vert w_j - w_i \Vert \leq \rho_i$. The combination of $\psi(\cdot)$ and $\log(\cdot)$ terms is ad-hoc and ultimately justified by their consistency proof.

\textbf{MS estimator of conditional mutual information} \citet{mesner_shalizi} propose an estimator that slightly modifies the CMI generalization of the GKOV estimator. The modifications are motivated by the observation that, besides $\rho_i = 0$, there is also a non-zero probability that different pairs of points have the same distance. Thus, the $k$-th nearest neighbor of $(\x_i, \y_i, \z_i)$ is non-unique with non-zero probability, and a non-unique $k$-th nearest neighbour is equivalent to $k < \tilde{k}_{\X\Y\Z,i}$ with $\tilde{k}_{\X\Y\Z,i}$ as defined by eq.~\eqref{eq:number-points-in-ball-with-boundary} for $W = \X\Y\Z$. Instead of $\rho_i = 0$, the event $k < \tilde{k}_{\X\Y\Z,i}$ is considered to indicate that $(\x_i, \y_i, \z_i)$ is a ``discrete'' point. Specifically, their estimator takes the form
\begin{multline}
\label{eq:ms} 
    \hat{I}^{MS}(\X;\Y|\Z) =  \frac{1}{n} \cdot \\ \sum_{i\,=\,1}^{n}\underbrace{\left[g(\tilde{k}_{\X\Y\Z,i}) + g (\tilde{k}_{\Z,i}) -g(\tilde{k}_{\X\Z,i}) - g(\tilde{k}_{\Y\Z,i})\right]}_{\equiv\, \hat{\xi}^{MS}_i(X; Y \vert Z)} \,
%     \hat{I}^{MS}(\X;\Y|\Z) = \max\Bigg\{0, \frac{1}{n} \cdot \\ \sum_{i\,=\,1}^{n}\underbrace{\left[g(\tilde{k}_{\X\Y\Z,i}) + g (\tilde{k}_{\Z,i}) -g(\tilde{k}_{\X\Z,i}) - g(\tilde{k}_{\Y\Z,i})\right]}_{\equiv\, \hat{\xi}^{MS}_i(X; Y \vert Z)} \,\Bigg\} 
\end{multline}
where $g(\cdot) = \psi(\cdot)$ if $\tilde{k}_{\X\Y\Z,i} = k$ and $g(\cdot) = \log(\cdot)$ if $\tilde{k}_{\X\Y\Z,i} > k$.\footnote{In their paper, \citet{mesner_shalizi} define their estimator by additionally computing a maximum of the estimate with $0$, motivated by the fact that $I(\X;\Y|\Z) \geq 0$. However, their implementation does seem not actually apply this, and preliminary experiments of ours show that the maximum with 0 can be detrimental for CIT. We do not apply the maximum with $0$ in any experiments.} The authors prove consistency of their estimator, and also show that it suffers from the curse of dimensionality: For fixed $\mX$ and $\mY$, if the dimension $\mZ$ of $\Zspace$ increases to infinity and $H(\Z)/\mZ$ is non-zero in this limit, then $\hat{I}^{MS}(\X;\Y|\Z)$ converges to $0$ in probability as $\mZ \to \infty$. The MS estimator equips the discrete components of $\Xspace$, $\Yspace$, $\Zspace$ with the discrete metric, which is equivalent to a one-hot encoding of the components and again raises the conceptual problem that the corresponding distance notions might not be semantically meaningful. In their experiments, the authors heuristically set $k=n/10$ where $n$ is the sample size.

\textbf{ZMADG estimator of conditional mutual information}
\cite{Zan2022ACM} assumes the absence of mixture variables and proposes an estimator for this case of mixed variables CMI $I(\X;\Y|\Z)$ that avoids defining a distance between qualitative components. They split $\X$, $\Y$, and $\Z$ in their respective quantitative components $\X^t, \Y^t, \Z^t$ and qualitative components $\X^l, \Y^l, \Z^l$ and express the CMI as
\begin{multline}\label{eq:cmi_cond_expression}
I(\X;\Y|\Z) = H(\X^t,\Z^t|\X^l, \Z^l) + H(\Y^t,\Z^t|\Y^l, \Z^l)- \\- H(\X^t, \Y^t,\Z^t|\X^l, \Y^l, \Z^l) - H(\Z^t|\Z^l) + H(\X^l, \Z^l) + \\ + H(\Y^l, \Z^l) - H(\X^l, \Y^l, \Z^l) - H(\Z^l) \, ,
\end{multline}
where the first four terms on the rhs are (conditional) differential entropies and the last four are (conditional) entropies. The (conditional) entropies are estimated with the standard plug-in estimator using empirical frequencies, while (conditional) differential entropies are calculated using the KL estimator on each subset of the samples defined by fixed values of the qualitative components and then averaging according to the empirical frequencies of the qualitative values. The parameter $k$ of the KL estimates is set to $k = \max \left\{\lfloor n_{cluster}/10\rfloor, \,1\right\}$ with $n_{cluster}$ the number of samples in the respective subsets determined by the values of the qualitative components (i.e., $k$ is separately chosen for each subset of samples). As a sum of consistent estimators, the estimator is consistent. The ZMADG estimator does not seem to suffer from the curse of dimensionality as the MS estimator, but, as we will further discuss below, we believe it incurs higher variance.

\subsection{Non-parametric CIT using CMI and a local permutation scheme}
\label{subsec:hyp}

To statistically test the null hypothesis $H_0:\, \X \ci \Y ~|~ \Z$ of conditional independence from finite samples $(\x_i, \y_i, \z_i)$, a distribution of the estimate $\hat{I}(\X;\Y|\Z)$ under the null hypothesis (the so-called null distribution) or an approximation thereof is needed. If $\X \ci \Y ~|~ \Z$, then the component values $\x_i$ and $\y_i$ within the subset of samples determined by the value $\z_i$ can be permuted arbitrarily without changing the distribution of the estimated CMI, i.e., setting $\tilde{\x}_i = \x_{\sigma(i)}$ with a permutation $\sigma$ such that for all $i$ both $\x_i$ and $\x_{\sigma(i)}$ are in the subset of samples determined by $\z_i$, the estimators $\hat{I}(\X;\Y|\Z)$ and $\hat{I}(\tilde{\X};\Y|\Z)$ have the same distribution. Since this equality holds for any such permutation, a null distribution can be obtained. For fully discrete $\Z$, the subset of samples determined by $\z_i$ are all samples $(\x_j, \y_j, \z_j)$ with $\z_j = \z_i$. For fully continuous $\Z$, \citet{runge2018conditional} uses a $k$-NN approach to determine the subsets of samples for which $\z_j \approx \z_i$ according to the $L^{\infty}$-distance. \citet{Zan2022ACM} adapt this method to the mixed data case: The sample $(\x_j, \y_j, \z_j)$ with $\z_j = (\z^t_j, \z^l_j)$, where $\z^t_j$ is the quantitative and $\z^l_j$ the qualitative component, is part of the subset of samples determined by $\z_i = (\z^t_i, \z^l_i)$ if and only if $\z^l_j = \z^l_i$ and $\z^t_j \approx \z^t_i$. Supplementaryi2 Material (SM) Sec. A describes how p-values are obtained.

\section{Proposed novel estimator}

We first discuss the problems of the MS and ZMADG estimators that motivate us to introduce a novel estimator. We then formally define this novel estimator and presents its theoretical guarantees.

\subsection{Motivation: Problems of the MS and ZMADG estimators}

% Such sets of $k$-NNs are not only conceptually problematic but can also be expected to negatively impact statistical power. 

We highlight three issues of the \textbf{MS estimator} \cite{mesner_shalizi}. First, it suffers from the conceptual problem that---because the $k$-NNs can come from different \emph{clusters} (defined as the subsets of samples points with equal values of the discrete variable)---it implicitly assumes local constancy across different clusters. However, different clusters might be entirely unrelated to each other. For example, it could be that dependence exists in only one of the clusters. Despite this fact, the MS estimator might estimate the local contribution of a point by combining neighbours from both the cluster with and without dependence. Not only does this give rise to conceptual complications, but it can also be expected to negatively affect statistical power. Second, due to the one-hot encoding of non-numeric discrete variables, the MS estimator is (as opposed to CMI) not invariant under scaling all variables with a common factor. Third, as discussed in \citet{mesner_shalizi}, the MS estimator is biased towards $0$ in high-dimensional settings. To exemplify, say the continuous and numeric discrete variables are scaled to $[0, 1]$ in preprocessing. Then, due to one-hot encoding and the $L^\infty$-metric, the maximum distance between any two sample points is $1$. Thus, if the clusters of the $i$-th sample point $w_i$ contains at most $k$ points, then $\rho_i = 1$ (since there are not enough points in the cluster), which in turn implies $k_{i, XZ} = k_{i, YZ}=k_{i, XYZ}=k_{i, Z} = n$ and hence $\xi^{MS}_i(X; Y \vert Z) = 0$. \citet{Zan2022ACM} discuss further cases in which the MS estimators suffers from local zero estimates. Generally, a bias towards zero can affect CI test performance because it can lead to false conclusions of independence. 

The \textbf{ZMADG estimator} \citet{Zan2022ACM} reduces these problems by considering each discrete cluster individually and adaptively reducing $k$ (in the estimation of entropies). However, this approach can lead to another issue that, unfortunately, has not yet been discussed or investigated in detail: Since the CMI estimator is a sum of up to $8$ entropy estimators, the CMI estimator might suffer from higher variance than the MS estimator, leading to increased CIT error rates.

\subsection{Definition and intuition of the proposed novel estimator}

To address these problems, we introduce a \textbf{novel CMI estimator \OurEstimator{}} that combines ideas from the MS and ZMADG estimators. Specifically, \OurEstimator{} can be understood as a variant of MS with the following two modifications.

First, instead of one-hot encoding non-numeric variables, we keep the original space $\Xspace \times \Yspace \times \Zspace$ and equip it with the $0-\infty$ ``metric''\footnote{Formally, the $0-\infty$ ``metric'' is not a metric due to the value $+\infty$. However, we use this formulation to highlight the similarity with MS.} defined as
\begin{equation}\label{eq:0infty}
\vert\vert w_i-w_j\vert\vert_{0-\infty}=
    \begin{cases}
        \vert\vert w_{i, c} - w_{j, c}\vert\vert_{L^{\infty}} & \text{if } w_{i, d}=w_{j,d}\\
        \infty & \text{otherwise}
    \end{cases} \, ,
\end{equation}
where we split the point $w_k = (x_k, y_k, z_k)$ into its numeric component $w_{k, c} = (x_{k,c}, y_{k,c}, z_{k,c})$ and its non-numeric component $w_{k, d} = (x_{k,d}, y_{k,d}, z_{k,d})$; similarly for the subspaces $\Xspace \times \Zspace$, $\Yspace \times \Zspace$, and $\Zspace$. That is, if $w_i$ and $w_j$ are in the same cluster (i.e., $w_{i, d} = w_{j,d}$), then their distance is finite and measured by the $L^\infty$-distance, else their distance is $\infty$. 

Second, we adopt the heuristic to adaptively set $k = \lfloor k_c \cdot (n_{cl, \min}-1)\rfloor$,  where $0 < k_c < 1$ is a hyperparameter and $n_{cl, \min}=\min_{i\in[[1,n]]}|\{w_j: \vert\vert w_i-w_j\vert\vert_{L^{\infty}} \neq \infty\}|$ is the number of points in the ``smallest'' cluster. The necessity of such a heuristic stems from the fact that, unlike in the infinite sample case, in practice some clusters might contain less than $k+1$ points. In Sec. D of the SM, we compare multiple heuristics and motivate our final choice. 

To formally specify our estimator, we first define the counts
\begin{equation}
\label{eq:number-points-in-ball-with-boundary-ours}
\tilde{k}_{W, i}^{0-\infty} = |\{ w_j ~|~ \Vert w_j - w_i \Vert_{0-\infty} \leq \rho_i, \, j \neq i \}| \, ,
\end{equation}
where $\rho_i < \infty$ is the $0-\infty$ distance of $w_i$ to its $k$-th nearest neighbour (here, since $\rho_i$ is finite, this distance equals the $L^\infty$-distance) and $W$ stands for $XYZ$, $XZ$, $YZ$ or $Z$. In terms of these counts, our estimator reads
\begin{multline}
      \hat{I}^{0-\infty}(\X;\Y|\Z) = \frac{1}{n} \cdot \\ \sum_{i\,=\,1}^{n}\underbrace{\left[g(\tilde{k}_{\X\Y\Z,i}^{0-\infty}) + g (\tilde{k}_{\Z,i}^{0-\infty}) - g(\tilde{k}_{\X\Z,i}^{0-\infty}) - g(\tilde{k}_{\Y\Z,i}^{0-\infty})\right]}_{\equiv \, \hat{\xi}^{0-\infty}_i(X; Y \vert Z)}  ,
\end{multline}
where $g(\cdot) = \psi(\cdot)$ if $\tilde{k}^{0-\infty}_{\X\Y\Z,i} = k$ and $g(\cdot) = \log(\cdot)$ if $\tilde{k}^{0-\infty}_{\X\Y\Z,i} > k$. 

% Theoretically, CMI is non-negative, yet the estimation can take negative values. We define our estimator without an additional operation of maximum with 0. We note that, while the MS estimator is formally defined as non-negative, this is not implemented in practice.

While the modifications that define our estimator \OurEstimator{} might appear minor at first, they indeed address the above explained problems of the MS and ZMADG estimators: First, our estimator by construction restricts all nearest neighbours of a point to the cluster of that point. Hence, our estimator does not assume local constancy across different clusters. Second, our estimator is invariant under a common scaling of all variables. Third, there seem to be fewer cases than for the MS estimator in which our estimator has local zero estimates: For example, in the case discussed for the third problem of the MS estimator. A discussion of all cases in which MS and \OurEstimator{} have local zero estimates is, however, out of scope. Thus, an empirical evaluation of the bias towards zero is called for. Fourth, unlike the ZMADG estimator, our estimator is not the sum of up to $8$ entropy terms but retains the same general form as the MS estimator. Thus, our estimator is not expected to incur increased variance, which is another hypothesis subject to empirical evaluation.

\subsection{Theoretical guarantees}

We provide theoretical guarantees of our estimator under the \textbf{assumptions} that (1) there are at most finitely many clusters as defined by the non-numeric components of $XYZ$ and (2) all numeric components of $XYZ$ have a finite range. We are confident that the theoretical guarantees also hold without the second assumption and that, to prove them, only mild adaptions of the corresponding proofs in \citet{mesner_shalizi} are needed. However, we consider such an adaption to be out of scope here and leave it to future work.

The presented theoretical guarantees concern the $k$-NN limit $\nLimit$ with $k \to \infty$ and $\tfrac{k}{n} \to 0$,\footnote{Note that our above heuristic choice of $k$ does not lead to $k/n \to 0$ as $n\to \infty$. However, that above choice should be considered as heuristic for how to choose $k$ for finite $n$, whereas the convergence results for $n \to \infty$ require $k/n \to 0$.} where $n$ is the sample size, and are based on the following Lemma.
\begin{mylemma}\label{lemma:Lq-convergence-to-MS}
Let $q$ be a positive integer, and let $X^\prime Y^\prime Z^\prime$ be obtained by applying a common non-constant affine function $h: \mathbb R \to \mathbb R$ to all numeric components of $X Y Z$ such that the ranges of all numeric components of $X^\prime Y^\prime Z^\prime$ are contained within the open interval $(0, 1)$.\footnote{Such a function $h$ exists due to the second assumption.} Then, the difference $\OursEstimatorMaxNull - \MSEstimatorMaxNullWithPrime$ converges to the constant $0$ in $L^q$-norm, that is,
\begin{equation}
\nLimit \, \E{\left\vert\OursEstimatorMaxNull - \MSEstimatorMaxNullWithPrime\right\vert^q } = 0 \, .
\end{equation}
\end{mylemma}
\textit{Proof sketch.}
Let $i_1, \, \ldots, i_q$ be arbitrary integers within $[[1, n]]$. It suffices to show that $\mathbb{E}[\chi]$ with $\chi = \prod_{\alpha \, = \, 1}^q \, \vert\hat{\xi}^{0-\infty}_{i_\alpha}(X; Y \vert Z) - \hat{\xi}^{MS}_{i_\alpha}(X^\prime; Y^\prime \vert Z^\prime)\vert$ converges to $0$. To this end, we first note that $\vert\hat{\xi}^{0-\infty}_{i}(X; Y \vert Z)\vert$ and $\vert\hat{\xi}^{(MS)}_{i}(X^\prime; Y^\prime \vert Z^\prime)\vert$ are bounded by $2 \log(n)$ and, thus, $\vert\chi\vert \leq 4^q \log(n)^q$. Next, consider the $i$-th sample point $w_i = (x_i, y_i, z_i)$. This point belongs to a certain cluster $C(w_i)$ as determined by the non-numeric part $w_{i,d}$ (if $w_i$ has only numeric components, then that cluster is the entire space). The probability $\mathbb{P}(C(w_i))$ that an arbitrary point lies in $C(w_i)$ is non-zero because else $w_i$ would not have been in $C(w_i)$. Then, since $k/n \to 0$, the probability that there are at least $k+1$ many points in $C(w_i)$ exponentially converges to one as $n \to \infty$ according to the Chernoff Bound. Thus, using the union bound, also the probability $\mathbb{P}(A)$ of the event $A$ that there are at least $k+1$ many points in all of the (not necessarily distinct) $p$ clusters $C(w_{i_1}), \, \ldots , \, C(w_{i_p})$ exponentially converges to one. Equivalently, $\mathbb{P}(A^c)$ is exponentially suppressed as $n \to \infty$. Finally, one can show that $\hat{\xi}^{0-\infty}_{i_\alpha}(X; Y \vert Z) = \hat{\xi}^{MS}_{i_\alpha}(X^\prime; Y^\prime \vert Z^\prime)$ conditioned on $A$, and thus $\chi = 0$ conditioned on $A$. Thus $\mathbb{E}[\chi] = \mathbb{E}[\chi ~|~ A^c] \cdot \mathbb{P}(A^c) \leq 4^q \log(n)^q \cdot \mathbb{P}(A^c)$, which implies $\nLimit \mathbb{E}[\chi] = 0$ due to the exponential suppression of $\mathbb{P}(A^c)$. \hfill $\square$

Writing the difference $\OursEstimatorMaxNull - I(\X;\Y|\Z)$ as $[\OursEstimatorMaxNull - \MSEstimatorMaxNullWithPrime] + [\MSEstimatorMaxNullWithPrime - I(\X;\Y|\Z)]$ and using $I(\X;\Y|\Z) = I(\X^\prime;\Y^\prime|\Z^\prime)$, Lemma~\ref{lemma:Lq-convergence-to-MS} transfers the convergence results of the MS estimator to our estimator. Specifically, we get the following.

\begin{mythm}
Our CMI estimator $\OursEstimatorMaxNull$ is $L^1$-consistent in the $k$-NN limit, that is
\begin{equation}
\nLimit \, \E{\left\vert\OursEstimatorMaxNull - I(\X;\Y|\Z)\right\vert } = 0 \, .
\end{equation}
\end{mythm}

\begin{mythm}
Assume that, in addition to the requirements of the $k$-NN limit, $\tfrac{\left[k \cdot \ln(n)\right]^2}{n} \to 0$ as $n \to \infty$. Then, our CMI estimator is $L^2$-consistent, that is,
\begin{equation}
\nLimit \, Var\left[\OursEstimatorMaxNull\right] = 0 \, .
\end{equation}
\end{mythm}

In particular, our estimator is asymptotically unbiased and converges in probability to the true CMI. We provide all proofs in the SM Sec. B.

\section{Numerical evaluation of the CMI estimators}
\label{sec:eval_estim}
Here, we empirically study and compare bias and variance of the MS, ZMADG and our \OurEstimator{} estimator. 
% Theoretically, CMI is always positive, but we allow for negative estimates.

\textbf{Experimental setup} We consider four models, partly taken from \citet{mesner_shalizi} and \citet{Zan2022ACM} for reproducibility. Since \citet{mesner_shalizi} and \citet{Zan2022ACM} do not mention any transformations of the continuous variables, we do not apply any preprocessing. We evaluate the mean and variances of the estimates on $100$ realizations qualitatively using violin plots that capture the estimates' mean and range, and quantitatively using statistical tests. Here, we present and study one model. Further results and an evaluation of computational runtimes are presented in Sec. E and F of the SM.

\underline{``Independent $Z$'' (\cite{mesner_shalizi}):} $X$ is discrete uniform $X\sim\mathcal{U}(\{0,\ldots,c\})$ with $1 \leq c \in \mathbb{N}$, $Y\sim\mathcal{U}\left([X, X+2]\right)$ is continuous uniform, and $Z = (Z_1, \ldots, Z_d)$ is discrete with $Z_i\sim Ber\left(0.5\right)$ for $1 \leq i \leq d$. The ground truth is $I(X;Y| Z)=\ln{c}-\frac{c-1}{c}\cdot \ln{2}$. In our experiments, we set the $c = 5$, vary the sample size $n \in \{300, 600, 1000, 2000\}$ and $d \in\{1, 3\}$. We vary the $k$ parameter using $k_c \in \{0.01, 0.1, 0.2, 0.3\}$: for MS, $k=k_c \cdot n$; for MS$_{0{-}\infty}$, $k=k_c \cdot n_{cl,\min}$; for ZMADG, $k=n_{subset} \cdot k_c$.

\begin{figure}[ht]
\centering
\includegraphics[width=0.39\linewidth]{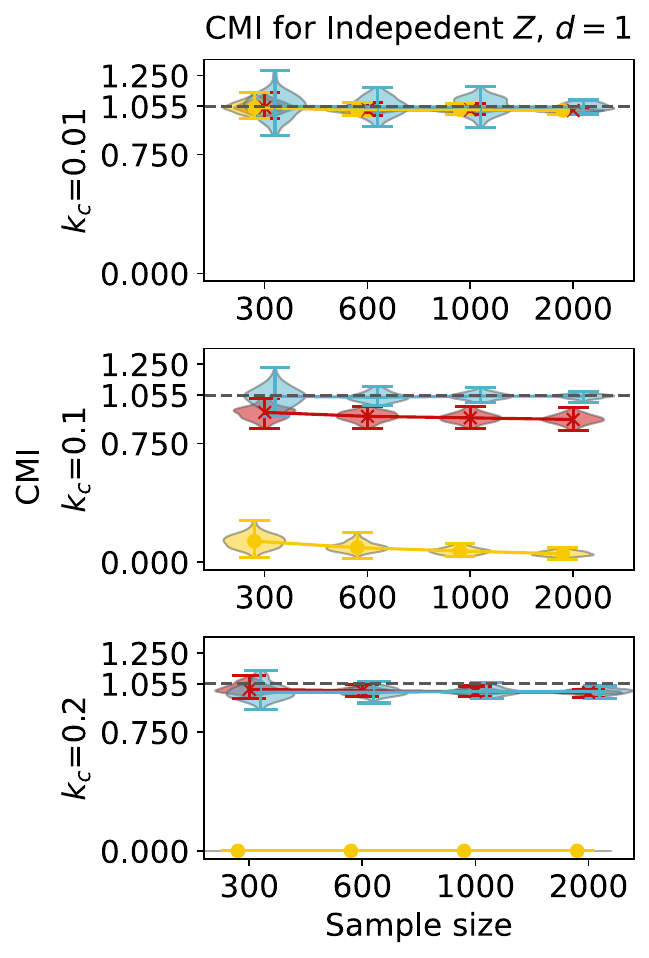}
\includegraphics[width=0.57\linewidth]{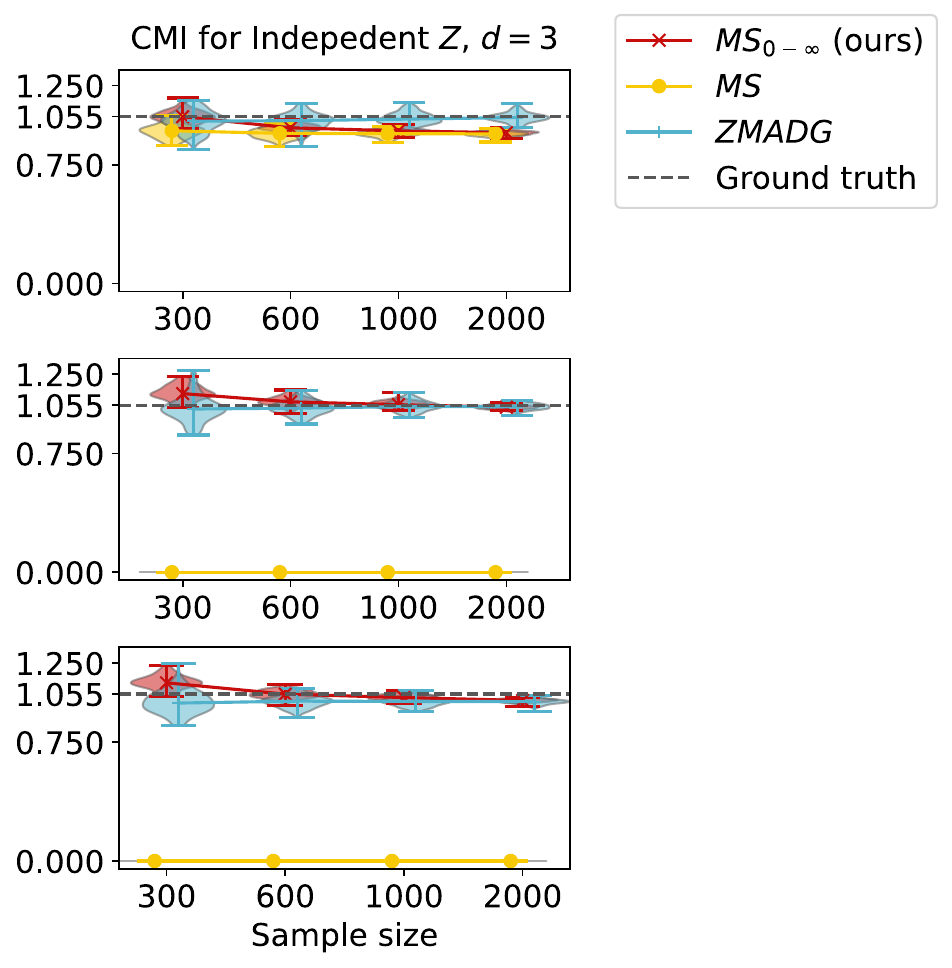}
\caption{Distribution of the CMI estimates for the \underline{"Independent $Z$"} model. Each row shows the results for different $k_c$. The ground truth is shown as the dashed line.}
\label{fig:cmi_t1_c1}
\end{figure}

\textbf{Results:} In the presented violin plots for the "Independent $Z$" model (Figure~\ref{fig:cmi_t1_c1}), we observe that for both dimensionalities  $d=1$ or $d=3$, all estimators perform comparably well only for $k_c=0.01$. For $k_c > 0.01$, the MS estimates show a bias towards zero, in contrast to the MS$_{0{-}\infty}$ estimates. The ZMADG estimator performs well, but, as anticipated, it has the highest variance among the estimators, especially for small $k_c$ and small $n$, and we observe marginal bias for $k_c>0.1$. Our approach also suffers from slight bias, e.g., for $d=1$ and $k_c=0.1$, and $d=3$ and smaller $n$ or $k_c \neq 0.1$. \underline{For the other three models}, we briefly summarize: MS has good performance except for the case when $X, Y$ are Gaussian and confounded by $Z$, while ZMADG performs well except when $X, Y$ are uniformly distributed and confounded by $Z$. ZMADG generally has higher variance than MS$_{0{-}\infty}$ and MS, especially for smaller $n$ and higher dimensionality. MS$_{0{-}\infty}$ has stable performance across all models but suffers from slight bias for large $k_c$ and small $n$. \underline{In summary}, we confirm the above expectations and observe that MS$_{0{-}\infty}$ draws a compromise between the strengths and weaknesses of MS and ZMADG: It alleviates bias towards zero at the expense of higher bias in cases with many discrete dimensions and small sample size. While not our primary focus, our MS$_{0{-}\infty}$ estimator can also handle mixture variables, i.e., variables that contain partially continuous and categorical samples, more robustly across $k_c$ values compared to the MS estimator, as shown by preliminary results in SM Sec. D.

\section{Evaluation of Conditional Independence Tests}
\label{sec:eval_sig}

\begin{figure*}[ht]
% \centering
\includegraphics[width=0.255\linewidth]{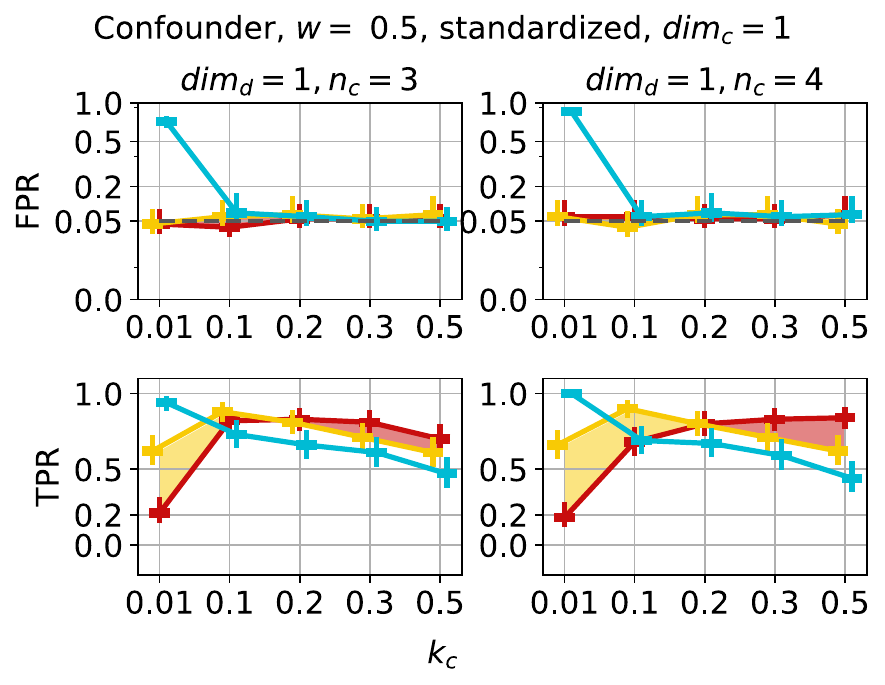}%
\includegraphics[width=0.255\linewidth]{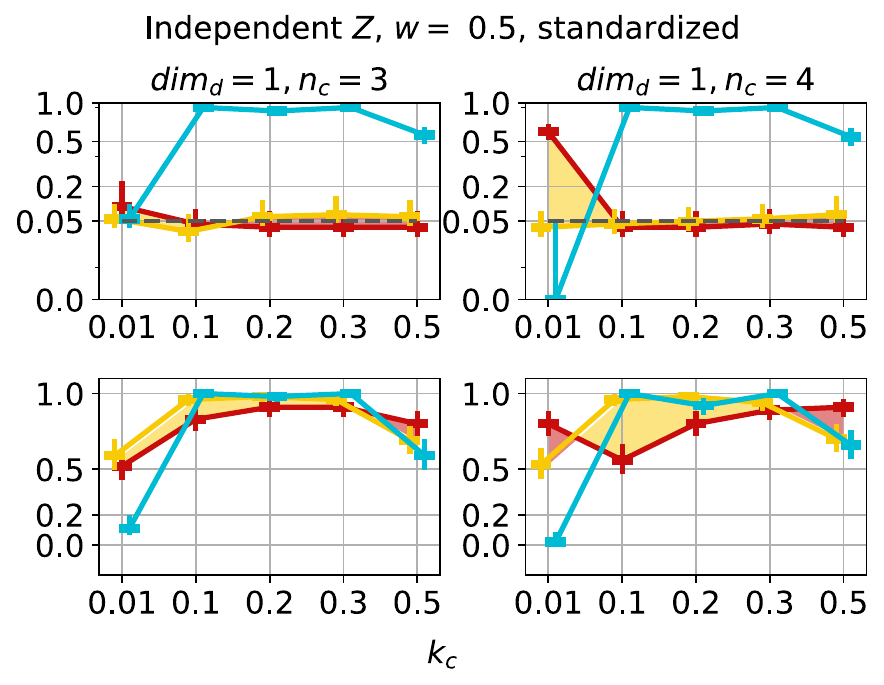}%
\includegraphics[width=0.49\linewidth]{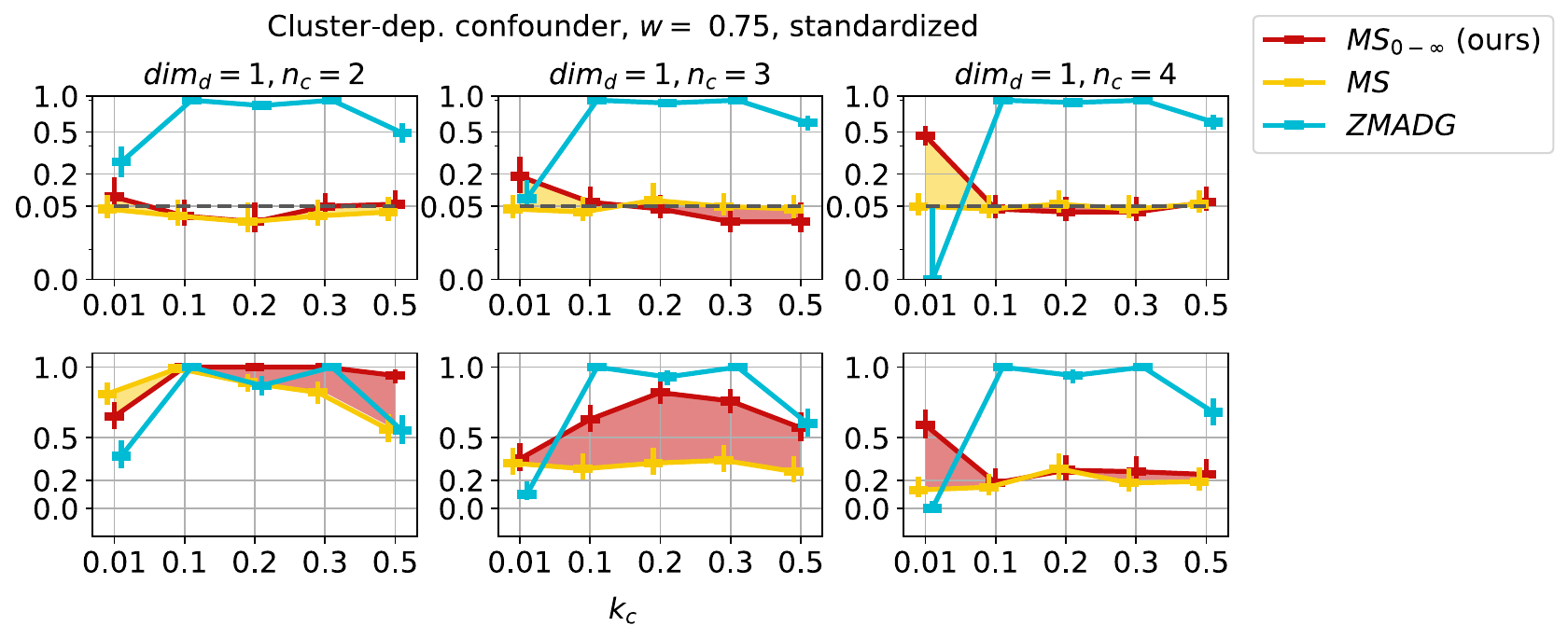}%

% \centering
\includegraphics[width=0.255\linewidth]{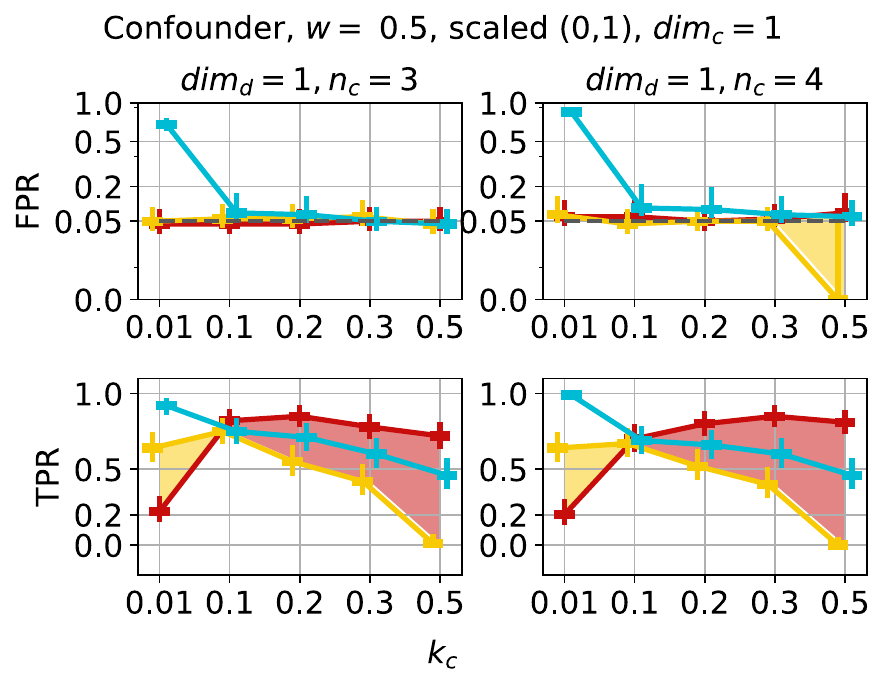}%
\includegraphics[width=0.255\linewidth]{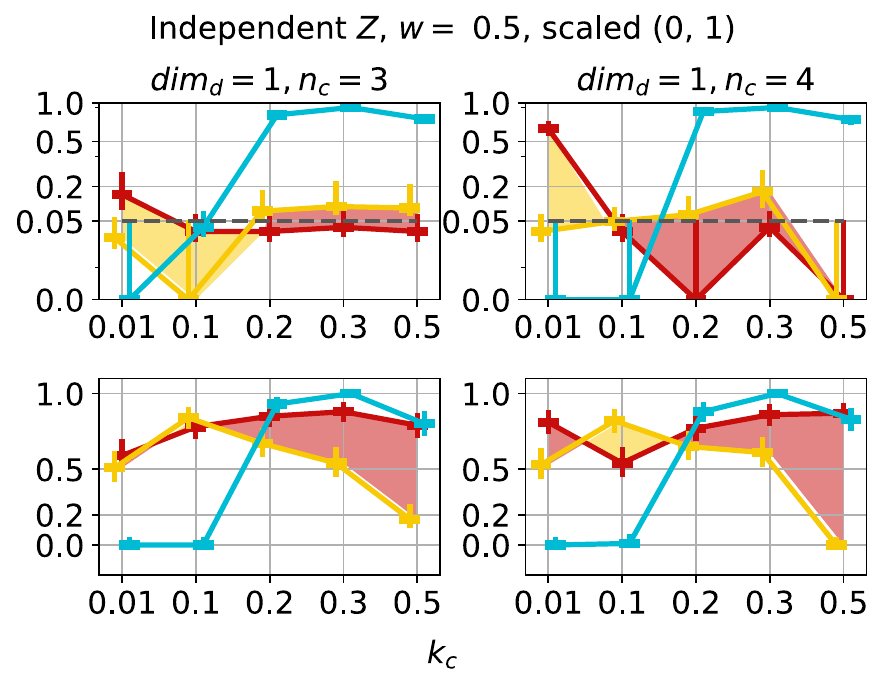}%
\includegraphics[width=0.39\linewidth]{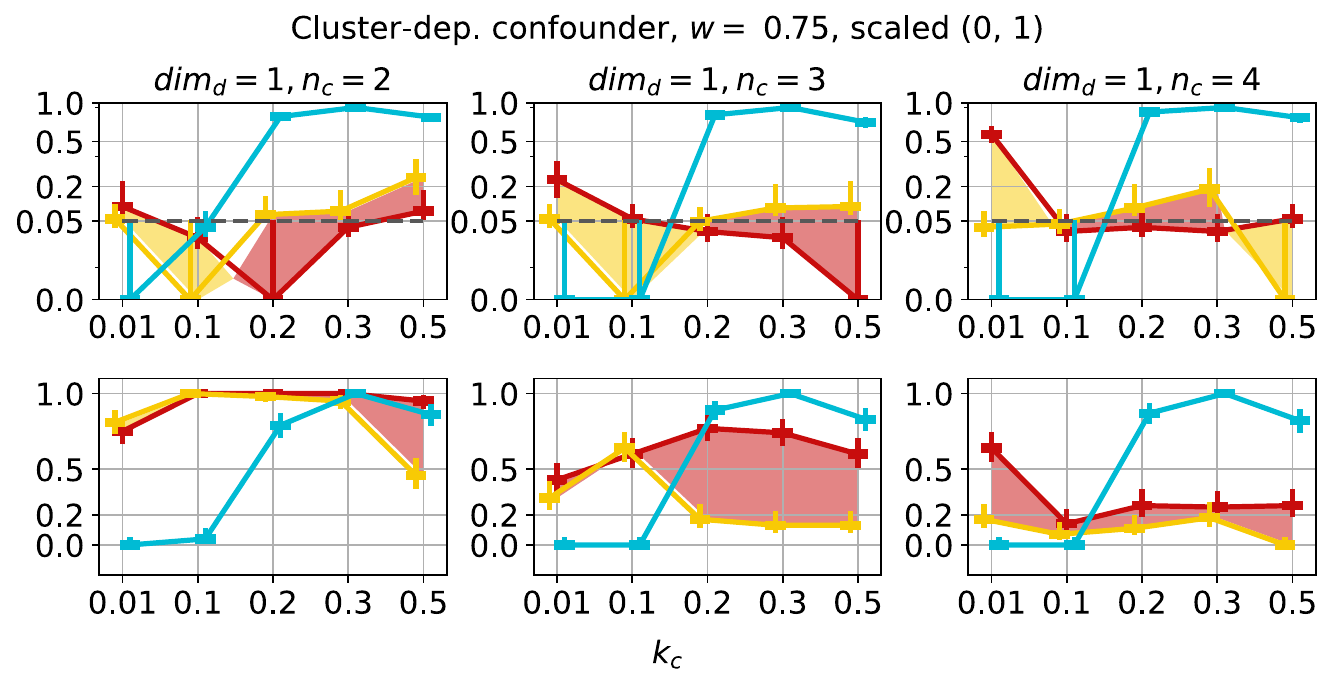}%

\caption{False positive rate (FPR, ideally under 0.05, log-scale) and true positive rate (TPR, higher is better, with 1 best) with standard error bars as the $95\%$ confidence interval (computed as described in SM Sec. E) for the following models with standardization (upper two rows) and scaling to $(0, 1)$ (bottom two rows) for the continuous variables and coupling factor $w$ as depicted: \underline{"Confounder"} (left), \underline{"Independent Z" } (center) and \underline{"Cluster-dependent confounder"}. The yellow areas indicate an advantage of the MS estimator, and the red areas indicate an advantage of the MS$_{0{-}\infty}$ estimator. }

\label{fig:sig}
\end{figure*}

We study whether the CIT controls the false positive rate (FPR) and retains statistical power, measured by the true positive rate (TPR) at a fixed significance level of $0.05$ in various mixed-type data setups. We focus on a controlled environment to obtain evaluation metrics with low error, which are difficult to obtain otherwise. \citet{Zan2022ACM} evaluate CIT for the ZMADG and MS estimators on synthetic and real-world data, computing acceptance rates using simulated fork, chain, and collider structures to evaluate FPR and TPR. The evaluation is done using models resembling those used in Sec.~\ref{sec:eval_estim}, involving different combinations of univariate mixed and discrete $X, Y, Z$. However, these models are not ideal for systematical CIT evaluation in a scenario close to the real world, where difficulties such as weak dependence occur. Moreover, the authors use rank transformations on the continuous variables, which, as they mention, can put the MS estimator at a disadvantage due to scaling. We apply standardization, re-scaling to $(0, 1)$, and rank transformation of the continuous variables for a comprehensive comparison.

\textbf{Experimental setup} Our data-generating models are inspired by the post-nonlinear model \cite{zhang2009identifiability}. We consider multiple causal structures where $Z$ is a confounder, part of a chain structure, or independent of $X$ and $Y$. We describe the individual data generation functions below. In all models, the coefficients $\beta_{\cdot}$ are randomly drawn as $\beta_{\cdot} \sim \mathcal{U}([-1, 1])$. We introduce dependence between $X$ and $Y$ using an additional noise term $\eta_W$ that influences both $X$ and $Y$, where $\eta_w \sim \mathcal{N}(0, 1)$ and the coupling factor $w$ that defines the dependency strength is $w=0$ for independence and $w > 0$ for dependence. The random variable $Z=(Z_1,\ldots, Z_m)$ is mixed-type with $dim_c$ continuous dimensions and $dim_d$ discrete dimensions, and each discrete component has $n_{c}$ categories. The noise terms $\eta_x, \eta_Y$ of $X$ and $Y$ follow $\mathcal{N}(0, 1)$.
% Discrete variables are drawn from a multinomial distribution with probabilities computed using the softmax function $\sigma(x)=\frac{e^{x_i}}{\sum_{i \in {1,...,n}e^{x_i}}}$, a process denoted $\sigma_{\sim}$. 

\underline{"Confounder"}:
Here, $Z_j \sim Bin(n_c-1,0.5)$ for the discrete components $Z_1, \ldots, Z_{dim_d}$ and $Z_j \sim \mathcal{N}(0, 1)$ for the continuous components $Z_{dim_d+1}, \ldots, Z_{m}$, and $V\in \{X,Y\}$ is calculated as follows:
\begin{equation}
         V = \!\sum_{j \in {1, ..., dim_d}} \beta_i \cdot l^{-1}(Z_j) + \!\sum_{j \in {dim_c+1, ... ,m}} \!\beta_j \cdot Z_j + \eta_v + w \cdot \eta_w
\end{equation}

Here, $l^{-1}(x)=\frac{e^x}{1 + e^x}$ is the inverse logit function. 

\underline{"Independent $Z$"}: Here, $Z$ has only discrete components $Z_1, \ldots, Z_{dim_d=m}$ where $Z_j \sim Bin(n_c-1,0.5)$. $X$ and $Y$ are continous univariate, and are computed as follows:
\begin{equation}
     X = \eta_x \!+ w \cdot \eta_w, \quad Y = \eta_y \!+ w \cdot \eta_w.
\end{equation}

\underline{"Cluster-dependent confounder"}: Here, $Z$ is discrete univariate $Z\sim Bin (n_c-1, 0.5)$. $X$ and $Y$ are continous univariate. For $Z=0$, $X$ and $Y$ are generated according to the "Confounder"-model with coupling factor $w > 0$, while for $Z \neq 0$ the same model is used with $w=0$. Thus, the variables are only dependent for the cluster formed by $Z=0$.

\underline{"Chain"}: $X$ and $Y$ are continuous, and $Z$ is discrete, and all variables are univariate. The model is defined as follows:
\begin{equation}
    \begin{aligned}
        & X = \eta_x + w \cdot \eta_w,  \\
        & Z = \left[\sigma_{\sim}(\beta_x \cdot X, (n_c - 1)) + \eta_z\right] \mod (n_c - 1), \\
        & Y = \beta_y \cdot l^{-1}(Z) + \eta_y + w \cdot \eta_w
    \end{aligned}
\end{equation}

Here, $\sigma_{\sim}(\beta_x \cdot X, n_c-1)$ denotes sampling from the multinomial distribution with $n_c-1$ categories where the $a$-th category (starting the count at 0) has probability $\tfrac{e^{a \cdot x}}{\sum_{a'=0}^{n_c-2} e^{a' \cdot x}}$. The noise $\eta_z$ follows $\eta_z \sim Bin(2, 0.7)$. 

\textbf{Results} We present results for a sample size of $1000$, with coupling factor for all models $w=0.5$ except for "Cluster-dependent confounder" where $w=0.75$ for $Z=0$. The number of classes is $n_c=3,\, 4$ for the "Confounder," "Independent $Z$," and "Chain" models, and $n_c=2, 3, 4$ for "Cluster-dependent confounder". We evaluate $k_c=0.01,\,0.1,\,0.2, \,0.3, \,0.5$ to set $k$ as in Sec.~\ref{sec:eval_estim}. For "Confounder" and  "Independent Z," we vary $dim_d=1,2$. We generate p-values with $300$ permuted surrogates using $k_{perm}=5$ and repeat each experiment $100$ times. Here, we present results with standardization and scaling to $(0, 1)$, and postpone the results with rank transformation to SM Sec.E. All code and experimental results for the CMI and CIT evaluation will be made public upon acceptance (see Sec. G in SM). 

For the \underline{"Confounder"}-model, we show results where the confounder $Z$ has one continuous and one discrete dimension (further results in SM. Sec. E). For the standardized case, we found that MS and MS$_{0{-}\infty}$ perform best (Fig.~\ref{fig:sig}, upper left). The scaling-related problem of MS becomes apparent when variables are scaled to $(0,1)$, which leads to a decrease in TPR as $k_c$ increases (Fig.~\ref{fig:sig}, bottom left). We observe another effect of this problem for $n_c=4$ and $k_c=0.5$: concurrently with the low TPR, the FPR of MS suddenly drops to 0 due to the observed and permutation statistics both being equal to 0, which results in a p-value equal to or close to 1. Our MS$_{0{-}\infty}$ CIT performs robustly even in this case. ZMADG typically gives satisfying results yet has lower TPR than MS and MS$_{0{-}\infty}$ and does not control FPR for $k_c=0.01$. For the \underline{"Independent $Z$"}-model (Fig.~\ref{fig:sig}, center), ZMADG has either low TPR or high FPR. MS and MS$_{0{-}\infty}$ perform better for both preprocessing types, yet we sometimes observe an elevated FPR. The scaling-related issues of MS persist for this model with scaling to $(0, 1)$ as well, while MS$_{0{-}\infty}$ again performs consistently well (except for $k_c=0.01$, where FPR is not controlled). Notably, we observe that the FPR of MS$_{0{-}\infty}$ drops to $0$ for $k_c=0.2$ and $k_c=0.5$. However, the corresponding increase in TPR indicates that these drops are not stemming from bias towards $0$. For all previous models, MS and MS$_{0{-}\infty}$ show similar performance for the standardized and rank transformation case, even as $k_c$ increases, since, in these models, there is dependence in each cluster. The results for the \underline{"Cluster-dependent confounder"}-model (Fig.~\ref{fig:sig}, right) show how MS performance suffers if data distributions differ between clusters. In this case, MS$_{0{-}\infty}$ identifies dependence more accurately than MS while controlling the FPR (with some exceptions) irrespective of the preprocessing method, especially for $n_c=3$. ZMADG again suffers from either high FPR or low TPR. \underline{Summarizing}, we observe that among the three estimators, MS and MS$_{0{-}\infty}$ have superior performance. Nonetheless, \OurEstimator{} seems to be the most robust estimator across the different models and preprocessing types. However, the hyperparameter $k_c$ has a crucial influence: In general, our approach benefits from higher $k_c$ (and thus higher $k$), possibly due to a reduction in variance, while for MS, the opposite holds. We provide recommendations on setting $k_c$ in the SM.

\section{Discussion and Conclusion}

Understanding the performance of CIT on heterogeneous data is pivotal for causal discovery and relevant across many fields, such as medicine or Earth Sciences. In this work, we evaluated the $k$-nearest neighbor CMI estimators of \citet{mesner_shalizi} (MS) and \citet{Zan2022ACM} (ZMADG) for mixed-type data and discussed their challenges: ZMADG suffers from high variance while MS treats categorical variables as numeric via one-hot encoding, leading to the conceptual problem of mixing categories. We proposed a modification of the MS estimator that mitigates the latter issue. We compared the bias and variance of the estimators and the corresponding CIT performance on synthetic models replicating realistic settings occurring in causal discovery. 

As anticipated, the ZMADG estimator has a low bias but high variance, particularly for small $k$ and $n$. The MS estimator has low variance but suffers from bias towards zero for larger $k$ and larger numbers of discrete variables. Our MS$_{0{-}\infty}$ reduces bias compared to MS and variance compared to ZMADG. Surprisingly, the ZMADG CIT test does not perform consistently, obtaining satisfying results only for the "Confounder" model. For other data models, it does not control false positives, and we suspect these problems occur in case of weaker dependencies. MS and \OurEstimator{} perform robustly, with MS slightly outperforming when dependence holds in all clusters, and the dimensionality is higher. However, their performance remains comparable considering the individual optimal $k_c$ value. When the continuous variables are scaled to $(0,1)$, the scaling-related problems of MS lead to underperformance for higher $k_c$, contrary to our estimator. Our method has superior performance for the case when data distributions differ between the clusters. Thus, both from a theoretical and an empirical perspective, we recommend our approach as the most robust estimator for mixed-type data scenarios. While our estimator suffers from the curse of dimensionality less than the MS estimator, the curse of dimensionality still applies to our estimator. Hence, sufficient samples per cluster are essential for reliable outcomes.

% Nevertheless, the user should still consider the curse of dimensionality, which also affects our estimator: Reliable results can only be obtained if there are sufficient samples per cluster. 

Lastly, while our analysis covered a range of scenarios, an evaluation of real-world data and causal discovery is beyond the scope of this paper and is left for future work.

\section{Acknowledgements}

This work was partly funded by the European Union’s Horizon 2020 research and innovation programme (project XAIDA, Grant No. 101003469) and by the European Research Council (ERC) (project Causal Earth, Grant No. 948112). We thank Tom Hochsprung for his valuable comments.

\bibliography{aaai24}

\twocolumn[
   \hsize\textwidth%
      \linewidth\hsize%
      \vskip 0.625in minus 0.125in%
      \centering%
      {\LARGE\bf  Non-parametric Conditional Independence Testing for Mixed Continuous-Categorical Variables: A Novel Method and Numerical Evaluation \\ SUPPLEMENTARY MATERIAL  \par}%
      \vskip 0.1in plus 0.5fil minus 0.05in%
      
]

% \begin{center}
% % \vbox{%
%       \hsize\textwidth%
%       \linewidth\hsize%
%       \vskip 0.625in minus 0.125in%
%       \centering%
%       {\LARGE\bf  Non-parametric Conditional Independence Testing for Mixed Continuous-Categorical Variables: A Novel Method and Numerical Evaluation \\ SUPPLEMENTARY MATERIAL  \par}%
%       % \vskip 0.1in plus 0.5fil minus 0.05in%
%   % }%
% \end{center}

% \appendix

\section{Nearest-Neighbor Permutation Test}

Algorithm \ref{alg:one} \citep{wasserman2010statistics} describes the procedure for generating the p-value from a set of $B$ permutations $\Pi=\{\pi_j | j=1,...,B\}$ of $n$ elements. Each permutation is generated using the nearest-neighbor scheme described in Sec. 2 of the main paper. Given a test statistic, in our case, the conditional mutual information (CMI) estimated using one of the estimators (MS, MS$_{0-\infty}$, or ZMADG), denoted as $T_{CMI}$, we obtain the p-value of the conditional independence test (CIT) of the respective estimator as follows:

\begin{algorithm}[H]
\caption{Local Permutation Test}\label{alg:one}
\SetKwInput{KwInput}{Input}                % Set the Input
\SetKwInput{KwOutput}{Output}              % set the Output
\DontPrintSemicolon
  \LinesNumbered
  \KwInput{samples $\{(x_i, y_i, z_i) | i=1,...,n\}$, permutation set $\Pi=\{\pi_j | j=1,...,B\}$, test statistic $T_{CMI}$, nominal level $\alpha$}
  \KwOutput{p-value}
For each permutation $\pi_j\in \Pi, j=1,...,B$: compute the statistic $T_{CMI}^{\pi_j}$ on the permuted samples

Compare the test statistic on the observed samples, $T_{CMI}$ with the permuted test statistics and calculate the p-value as 
\begin{equation}
    p= \frac{1}{B}\sum_{\pi_j \in \Pi} \mathbf{1}\{T_{CMI}^{\pi_j} \geq T_{CMI}\}
\end{equation}
\end{algorithm}

\section{Proofs}

In this part of the Supplementary Material, we formally prove the theoretical claims made in Sec.~3 of the main paper. Our proofs rely on the following two assumptions:

\begin{myassump}\label{assumption:finitly-many-clusters}
There are at most finitely many clusters as defined by the non-numeric components of $XYZ$
\end{myassump}

\begin{myassump}\label{assumption:finite-range}
All numeric components of $XYZ$ have a finite range.
\end{myassump}

Note that also \cite{mesner_shalizi} implicitly make Assumption~\ref{assumption:finitly-many-clusters} by assuming finite-dimensional random vectors in combination with one-hot encoding the non-numeric components. As we mentioned in the main text, we are confident that the theoretical guarantees of our estimator also hold without Assumption~\ref{assumption:finite-range} and that, to prove them in this case, only mild adaptions of the corresponding proofs in \citet{mesner_shalizi} are needed. However, we leave such an adaption of the proofs to future work and here do adopt Assumption~\ref{assumption:finite-range}.

\textbf{Proof of Lemma~1.}

We start by using the triangle inequality to get
% \begin{equation}
\begin{gather}
% \resizebox{\linewidth}{!}{
\begin{aligned}
&\nLimit \, \E{\left\vert \hat{I}^{0-\infty}(X;Y| Z) - \hat{I}^{MS}(X^\prime;Y^\prime| Z^\prime) \right\vert^q }\\
= &\nLimit \, \E{\left\vert\frac{1}{n} \sum_{i \,=\, 1}^n \hat{\xi}^{0-\infty}_i(X;Y|Z) - \frac{1}{n}\sum_{i \,=\, 1}^n \hat{\xi}^{MS}_i(X^\prime;Y^\prime|Z^\prime)\right\vert^q }\\
\leq &\nLimit \, \frac{1}{n^q}\, \E{\left( \sum_{i\,=\,1}^n \left\vert\hat{\xi}^{0-\infty}_i(X;Y|Z)  - \hat{\xi}^{MS}_i(X^\prime;Y^\prime|Z^\prime)\right\vert\right)^q\,} \\
= &\nLimit \, \frac{1}{n^q} \sum_{i_1 \,=\,1}^n\sum_{i_2 \,=\,1}^n \dots \\ & \sum_{i_q \,=\,1}^n \, \E{\prod_{\alpha \,=\,1}^q \left\vert\ \hat{\xi}^{0-\infty}_{i_\alpha}(X;Y|Z) - \hat{\xi}^{MS}_{i_\alpha}(X^\prime;Y^\prime|Z^\prime)\right\vert} \, . 
\end{aligned}
% }
\label{eq:proof-LP-convergence-to-MS-eq-1}
\end{gather}
% \end{equation}

Next, let $N_n(C_i)$ be the number of points other than the $i$-th point that are in the cluster $C_i$ of the $i$-th point. This random variable $N_n(C_i)$ follows the distribution $Bin(n-1, p_i)$, where $p_i = \mathbb{P}(C_i)$ is the probability that an arbitrary point belongs to the cluster $C_i$. We have $p_i > 0$ because else the $i$-th point would not have been in the cluster $C_i$. The Chernoff bound for the lower tail of the Binomial distribution then gives
\begin{equation}
%\mathbb{P}(N_n(C_i) \leq k -1) \leq e^{-\left(\tfrac12 n \cdot p_i - k\right)} \, .
\mathbb{P}(N_n(C_i) \leq k -1) \leq e^{-\tfrac{(n-1)\cdot p_i}{2}\cdot\left[1 - \tfrac{k-1}{(n-1)\cdot p_i}\right]^2} \, .
\end{equation}
Now let $A_{i_\alpha}$ be the event that $N_n(C_{i_\alpha}) \leq k-1$ and let $A_{i_1, \, \ldots, i_q} = \cup_{\alpha \,=\,1}^q A_{i_\alpha}$. The probability of $A_{i_1, \, \ldots, i_q}$ is upper bounded according to
\begin{equation}
\begin{aligned}
\mathbb{P}(A_{i_1, \, \ldots, i_q}) 
& \leq \sum_{\alpha \, = \, 1}^q \mathbb{P}(N_n(C_{i_\alpha}) \leq k -1) \\
& \leq \sum_{\alpha \, = \, 1}^q e^{-\tfrac{(n-1)\cdot p_{i_\alpha}}{2}\cdot\left[1 - \tfrac{k-1}{(n-1)\cdot p_{i_\alpha}}\right]^2} \, .
\end{aligned}
\end{equation}
Next, we use the law of total expectation to condition the expectation value on the right-hand-side of the last line of ineq.~\eqref{eq:proof-LP-convergence-to-MS-eq-1} on the events $A_{i_1, \, \ldots, i_q}$ and $A_{i_1, \, \ldots, i_q}^c$, which gives

\newpage 

\begin{strip}
\begin{equation}
\begin{aligned}
&\E{\prod_{\alpha \,=\,1}^q \left\vert\hat{\xi}^{0-\infty}_{i_\alpha}(X;Y|Z)  - \hat{\xi}^{MS}_{i_\alpha}(X^\prime;Y^\prime|Z^\prime)\right\vert} \\
&= \, \, \underbrace{\E{\prod_{\alpha \,=\,1}^q \left\vert\hat{\xi}^{0-\infty}_{i_\alpha}(X;Y|Z)  - \hat{\xi}^{MS}_{i_\alpha}(X^\prime;Y^\prime|Z^\prime)\right\vert ~\middle\vert~ A_{i_1, \, \ldots, i_q}}}_{\geq 0} \cdot \underbrace{\mathbb{P}(A_{i_1, \, \ldots, i_q})}_{\leq \,\sum_{\alpha \, = \, 1}^q \exp\left\{-\tfrac{(n-1)\cdot p_{i_\alpha}}{2}\cdot\left[1 - \tfrac{k-1}{(n-1)\cdot p_{i_\alpha}}\right]^2\right\}} \\
&\,\, + \mathbb{E}\Bigg[\,\,\underbrace{\prod_{\alpha \,=\,1}^q \left\vert\hat{\xi}^{0-\infty}_{i_\alpha}(X;Y|Z)  - \hat{\xi}^{MS}_{i_\alpha}(X^\prime;Y^\prime|Z^\prime)\right\vert ~\Bigg\vert~ A_{i_1, \, \ldots, i_q}^c}_{= \, 0}\,\,\Bigg] \cdot \underbrace{\mathbb{P}(A_{i_1, \, \ldots, i_q}^c)}_{\leq \, 1} \\
&\leq \E{\prod_{\alpha \,=\,1}^q \left\vert\hat{\xi}^{0-\infty}_{i_\alpha}(X;Y|Z)  - \hat{\xi}^{MS}_{i_\alpha}(X^\prime;Y^\prime|Z^\prime)\right\vert ~\middle\vert~ A_{i_1, \, \ldots, i_q}} \cdot \sum_{\alpha \, = \, 1}^q e^{-\tfrac{(n-1)\cdot p_{i_\alpha}}{2}\cdot\left[1 - \tfrac{k-1}{(n-1)\cdot p_{i_\alpha}}\right]^2}
\end{aligned}
\end{equation}
\end{strip}

Here, we have used $\hat{\xi}^{0-\infty}_{i_\alpha}(X;Y|Z) = \hat{\xi}^{MS}_{i_\alpha}(X^\prime;Y^\prime|Z^\prime)$ conditioned on the event $A_{i_1, \, \ldots, i_q}^c$. This equality holds for the following reason: Given $A_{i_1, \, \ldots, i_q}^c$, for all $1 \leq \alpha \leq q$ at least $k$ points other than the $i_\alpha$-point are in the cluster $C_{i_\alpha}$ of $i_\alpha$. Now fix some $\alpha$ with $1 \leq \alpha \leq q$ and let $v_0, v_1, \ldots, v_m$ be the points in cluster $C_{i_\alpha}$, ordered such that $v_0$ is the $i_\alpha$-th point and that $\Vert v_a - v_0\Vert_{L^\infty} \leq \Vert v_b - v_0\Vert_{L^\infty}$ if $a < b$. Then, $m \geq k$ and the distance $\rho_{i_\alpha}$ used by the estimate $\hat{\xi}^{0-\infty}_{i_\alpha}(X;Y|Z)$, here denoted as $\rho_{i_\alpha}^{0-\infty}$, equals $\Vert v_k - v_0\Vert_{L^\infty} < \infty$. Let $h(w) = \lambda \cdot w + c$ with $\lambda \neq 0$ be the non-constant affine function that transforms the numeric components of $XYZ$ to $X^\prime Y^\prime Z^\prime$, and for all $0 \leq a \leq m$ let $v_a^\prime$ denote the transformed version of $v_a$. Since the same transformation $h$ is applied to all numeric components of $XYZ$, we get that $\Vert v^\prime_a - v^\prime_0\Vert_{L^\infty} \leq \Vert v^\prime_b - v^\prime_0\Vert_{L^\infty}$ if $a < b$. Moreover, since the ranges of numeric components of $X^\prime Y^\prime Z^\prime$ are contained within $(0, 1)$, we get that $\Vert v^\prime_m - v^\prime_0\Vert_{L^\infty} < 1$. Consequently, for the purpose of $\hat{\xi}^{MS}_{i_\alpha}(X^\prime;Y^\prime|Z^\prime)$ the $k$-nearest neighbours of $v^\prime_0$ are $v^\prime_1, \, \ldots, v^\prime_k$ and the distance $\rho_{i_\alpha}$ used by the estimate $\hat{\xi}^{MS}_{i_\alpha}(X^\prime;Y^\prime|Z^\prime)$, here denoted as $\rho_{i_\alpha}^{MS}$, equals $\Vert v^\prime_k - v_0\Vert_{L^\infty} = \lambda \cdot \rho_{i_\alpha}^{0-\infty} < 1$. We thus see that also $\hat{\xi}^{MS}_{i_\alpha}(X^\prime;Y^\prime|Z^\prime)$ uses only points within the cluster $C_{i_\alpha}$ of the $i_\alpha$-th point. Let $W$ be a wildcard for $XYZ$, $XZ$, $YZ$ and $Z$ and consider the count $\tilde{k}^{0-\infty}_{W, i_\alpha}$ used by the estimate $\hat{\xi}^{0-\infty}_{i_\alpha}(X;Y|Z)$ as defined in eq.~(14) in the main text. By definition, see eqs.~(12) and (13) in the main text, this count is the number of points other than the $i_\alpha$-th point $v_0$ that in $W$-space have a $0-\infty$ ``distance'' of at most $\rho^{0-\infty}_{i_\alpha}$ to $v_0$. Since $\rho^{MS}_{i_\alpha} = \lambda \cdot \rho^{0-\infty}_{i_\alpha} < 1$ and the $L^\infty$-distance in $W^\prime$-space (where, for example, $W^\prime = X^\prime Y^\prime Z^\prime$ if $W = XYZ$) is $\lambda$ times the $L^\infty$-distance in $W$-space, the corresponding count $\tilde{k}_{W^\prime, i_\alpha}$  used by $\hat{\xi}^{MS}_{i_\alpha}(X^\prime;Y^\prime|Z^\prime)$, see eqs.~(9) and (10) in the main text, equals $\tilde{k}^{0-\infty}_{W, i_\alpha}$. From the equality $\tilde{k}^{0-\infty}_{W, i_\alpha} = \tilde{k}_{W^\prime, i_\alpha}$ we conclude that, as claimed, $\hat{\xi}^{0-\infty}_{i_\alpha}(X;Y|Z) = \hat{\xi}^{MS}_{i_\alpha}(X^\prime;Y^\prime|Z^\prime)$ conditioned on the event $A_{i_1, \, \ldots, i_q}^c$.

To upper bound the remaining conditional expectation, we use the triangle inequality and the fact that $\ln(a) \geq \psi(a) \geq 0$ for $a \geq 0$ to get
\begin{equation}
\begin{aligned}
& \left\vert\hat{\xi}^{0-\infty}_i(X;Y| Z) \right\vert \leq 2 \ln(n) \hspace{0.5em} \text{and} \\
& \hspace{0.5em} \left\vert\hat{\xi}^{MS}_i(X^\prime;Y^\prime| Z^\prime) \right\vert \leq 2 \ln(n) \, ,
\end{aligned}
\end{equation}
which implies
\begin{equation}
\left\vert\hat{\xi}^{0-\infty}_i(X;Y| Z) -\hat{\xi}^{MS}_i(X^\prime;Y^\prime| Z^\prime)\right\vert \leq 4 \ln(n) \, .
\end{equation}
We thus find
\begin{gather}
% \resizebox{\linewidth}{!}{
\begin{aligned}
\E{\prod_{\alpha \,=\,1}^q \left\vert\hat{\xi}^{0-\infty}_{i_\alpha}(X;Y|Z)  - \hat{\xi}^{MS}_{i_\alpha}(X^\prime;Y^\prime|Z^\prime)\right\vert ~\middle\vert~ A_{i_1, \, \ldots, i_q}} \\ \leq 4^q \ln(n)^q \, .
\end{aligned}
% }
\end{gather}

By combining the above results, we get

\begin{gather}
% \resizebox{\linewidth}{!}{
\begin{aligned}
&\nLimit \, \E{\left\vert\hat{I}^{0-\infty}(X;Y| Z) - \hat{I}^{MS}(X^\prime;Y^\prime| Z^\prime)\right\vert^q } \\ 
\leq & \nLimit \frac{4^q \cdot \ln(n)^q}{n^{q}} \cdot \\ 
& \,\left(\sum_{i_1 \,=\,1}^n\sum_{i_2 \,=\,1}^n \dots \sum_{i_q \,=\,1}^n\right)\, \sum_{\alpha \,=\,1}^q e^{-\tfrac{(n-1)\cdot p_{i_\alpha}}{2}\cdot\left[1 - \tfrac{k-1}{(n-1)\cdot p_{i_\alpha}}\right]^2} \, .
\end{aligned}
% }
\end{gather}

Let $C^{(1)}, \, \ldots, C^{(m)}$ be the list of all clusters with positive probability, where $m <  \infty$ because according to Assumption~\ref{assumption:finitly-many-clusters} there are at most finitely many clusters. Then $p_{min} = \min_{\gamma \in [[1,m]]} \mathbb{P}(C^{(\gamma)})$, which is independent of $k$ and $n$, exists and $p_{min} > 0$. Noting that $p_{i_\alpha} = \mathbb{P}(C_{i_{\alpha}}) > 0$ for all $i_\alpha$ because else the $i_{\alpha}$-th point would not have been in the cluster $C_{i_{\alpha}}$, we get that $1 \geq p_{i_\alpha} \geq p_{min} > 0$ for all $i_{\alpha}$ and hence
\begin{equation}
\frac{k-1}{(n-1)\cdot p_{i_\alpha}} \leq \frac{k-1}{(n-1)\cdot p_{min}}
\end{equation}
for all $i_{\alpha}$. Since $\tfrac{k}{n} \to 0$ and $k \to \infty$ in the $k$-nn limit and since $p_{i_\alpha}$ is independent of $k$ and $n$, we find that $\tfrac{k-1}{(n-1)\cdot p_{i_\alpha}} \to 0^+$ for all $i_\alpha$ and  $\tfrac{k-1}{(n-1)\cdot p_{min}} \to 0^+$. Thus, there is a positive integer $n_0$ such that for all $n \geq n_0$ and for all $i_\alpha$ the bound
\begin{equation}
\left[1 - \frac{k-1}{(n-1)\cdot p_{i_\alpha}}\right]^2 \geq \left[1 - \frac{k-1}{(n-1)\cdot p_{min}}\right]^2 
\end{equation}

holds. We then get

\begin{gather}
% \resizebox{\linewidth}{!}{
\begin{aligned}
&\nLimit \, \E{\left\vert\hat{I}^{0-\infty}(X;Y| Z) - \hat{I}^{MS}(X^\prime;Y^\prime| Z^\prime)\right\vert^q } \\
\leq &\nLimit \frac{4^q \cdot \ln(n)^q}{n^{q}} \cdot \\ 
 & \,\left(\sum_{i_1 \,=\,1}^n\sum_{i_2 \,=\,1}^n \dots \sum_{i_q \,=\,1}^n\right)\, \sum_{\alpha \,=\,1}^q e^{-\tfrac{(n-1)\cdot p_{min}}{2}\cdot\left[1 - \tfrac{k-1}{(n-1)\cdot p_{min}}\right]^2} \, .
\end{aligned}
% }
\end{gather}

We can now pull the sum of exponentials (that is, the sum over $\alpha$) out of the product-sum (that is, the sums over $i_1, \, \ldots, i_q$) and get

\begin{gather}
% \resizebox{\linewidth}{!}{
\begin{aligned}
& \nLimit \, \E{\left\vert\hat{I}^{0-\infty}(X;Y| Z) - \hat{I}^{MS}(X^\prime;Y^\prime| Z^\prime)\right\vert^q } \\
\leq & \!\begin{aligned}[t] & \nLimit \frac{4^q \cdot \ln(n)^q}{n^{q}} \cdot \\
& \,\left(\sum_{i_1 \,=\,1}^n\sum_{i_2 \,=\,1}^n \dots \sum_{i_q \,=\,1}^n\right)\, \sum_{\alpha \,=\,1}^q e^{-\tfrac{(n-1)\cdot p_{min}}{2}\cdot\left[1 - \tfrac{k-1}{(n-1)\cdot p_{min}}\right]^2} \end{aligned} \\
 =  & \!\begin{aligned}[t] & \nLimit  \, \,\frac{4^q \cdot \ln(n)^q}{n^{q}} \cdot \\
 & \underbrace{\sum_{\alpha \,=\,1}^q e^{-\tfrac{(n-1)\cdot p_{min}}{2}\cdot\left[1 - \tfrac{k-1}{(n-1)\cdot p_{min}}\right]^2}}_{= \, q \cdot \exp\left\{-\tfrac{(n-1)\cdot p_{min}}{2}\cdot\left[1 - \tfrac{k-1}{(n-1)\cdot p_{min}}\right]^2\right\}}\,\underbrace{\left(\sum_{i_1 \,=\,1}^n\sum_{i_2 \,=\,1}^n \dots \sum_{i_q \,=\,1}^n\right)}_{n^q} \end{aligned} \\
=  & \nLimit  \,4^q \cdot \ln(n)^q \cdot q \cdot e^{-\tfrac{(n-1)\cdot p_{min}}{2}\cdot\left[1 - \tfrac{k-1}{(n-1)\cdot p_{min}}\right]^2} \\
=  & \, 0
\end{aligned}
% }
\end{gather}

The last equality follows because the argument of the exponential goes to $-\infty$ in the $k$-nn limit.  \hfill $\qedsymbol$

\textbf{Proof of Theorem 2.}
The sequence $X_n = \hat{I}^{0-\infty}(X;Y| Z) - \hat{I}^{MS}(X^\prime;Y^\prime| Z^\prime)$ converges to $X = 0$ in $L^1$ according to Lemma~1 with $q = 1$, and the sequence $Y_n  = \hat{I}^{MS}(X^\prime;Y^\prime| Z^\prime) - I(X^\prime;Y^\prime|Z^\prime)$ converges to $Y = 0$ in $L^1$ according to the proof of Theorem 3.1 in \citep{mesner_shalizi}
\footnote{That theorem itself only claims asymptotic unbiasedness, which is strictly weaker than $L^1$-convergence, but the proof actually shows $L^1$-convergence.} Therefore, the sequence $Z_n = X_n + Y_n = \hat{I}^{0-\infty}(X;Y| Z) - I(X^\prime;Y^\prime|Z^\prime)$ converges to $X + Y = 0$ in $L^1$. We conclude the proof by noting the equality $I(X^\prime;Y^\prime|Z^\prime) = I(X;Y|Z)$, which follows because CMI is invariant under componentwise non-constant affine transformations.\hfill $\qedsymbol$

\textbf{Proof of Theorem 3.}
The sequence $X_n = \hat{I}^{0-\infty}(X;Y| Z) - \hat{I}^{MS}(X^\prime;Y^\prime| Z^\prime)$ converges to $X = 0$ in $L^2$ according to Lemma~1 with $q = 2$, and the sequence $Y_n  = \hat{I}^{MS}(X^\prime;Y^\prime| Z^\prime) - I(X^\prime;Y^\prime|Z^\prime)$ converges to $Y = 0$ in $L^2$ according to Theorem 3.2 in \citet{mesner_shalizi} in combination with Theorem 3.1 in \citet{mesner_shalizi}. Therefore, the sequence $Z_n = X_n + Y_n = \hat{I}^{0-\infty}(X;Y| Z) - I(X^\prime;Y^\prime|Z^\prime)$ converges to $X + Y = 0$ in $L^2$. We conclude the proof by noting the equality $I(X^\prime;Y^\prime|Z^\prime) = I(X;Y|Z)$, which follows because CMI is invariant under componentwise non-constant affine transformations.\hfill $\qedsymbol$

% \bibliography{library}
% \bibliographystyle{apalike}

% \end{document}

\section{Further Results of the Numerical Evaluation of the CMI Estimators}
\label{app:further_Estim}

Here, we present further results for the numerical evaluation of the CMI estimators. 

\textbf{Experimental setup} We keep the experimental setup as described in Sec. 4 of the main paper and generate further synthetic datasets according to the following models: 

\underline{"Chain structure'' (\cite{mesner_shalizi}):} Here, $X\sim \exp\left(10\right)$, $Z = (Z_1, \ldots, Z_{d})$ is multivariate with $Z_1\sim Poisson\left(X\right)$ and $Z_i \sim \mathcal{N}(0, 1)$ for $2 \leq i \leq d$, and $Y\sim Bin\left(Z_1,\ 0.5\right)$. The ground truth is $I(X;Y|Z) = 0$.

\underline{"Confounder with Gaussian $\X$ and $\Y$" (\citet{Zan2022ACM}):} This model describes a confounder structure with normally distributed $X$ and $Y$, where $X\sim\mathcal{N}\left(Z,1\right)$, $Y\sim\mathcal{N}\left(Z,\ 1\right)$, and $Z\sim\mathcal{U}\left(\{0,\ldots,m\}\right)$. The ground truth is $I(X,Y|Z) =0$. As in \citet{Zan2022ACM}, $m=9$.

\underline{"Confounder with uniform $\X$ and $\Y$"}: This model describes a confounder structure with uniformly distributed $X$ and $Y$, where $X\sim \mathcal{U}\left(0,Z\right)$, $Y\sim\mathcal{U}\left(Z, Z+1\right)$, and $Z\sim\mathcal{U}\left(\{0,1\}\right)$. The ground truth is $I(X, Y|Z)=0$.

\begin{figure}[ht]
\centering
\includegraphics[width=0.41\linewidth]{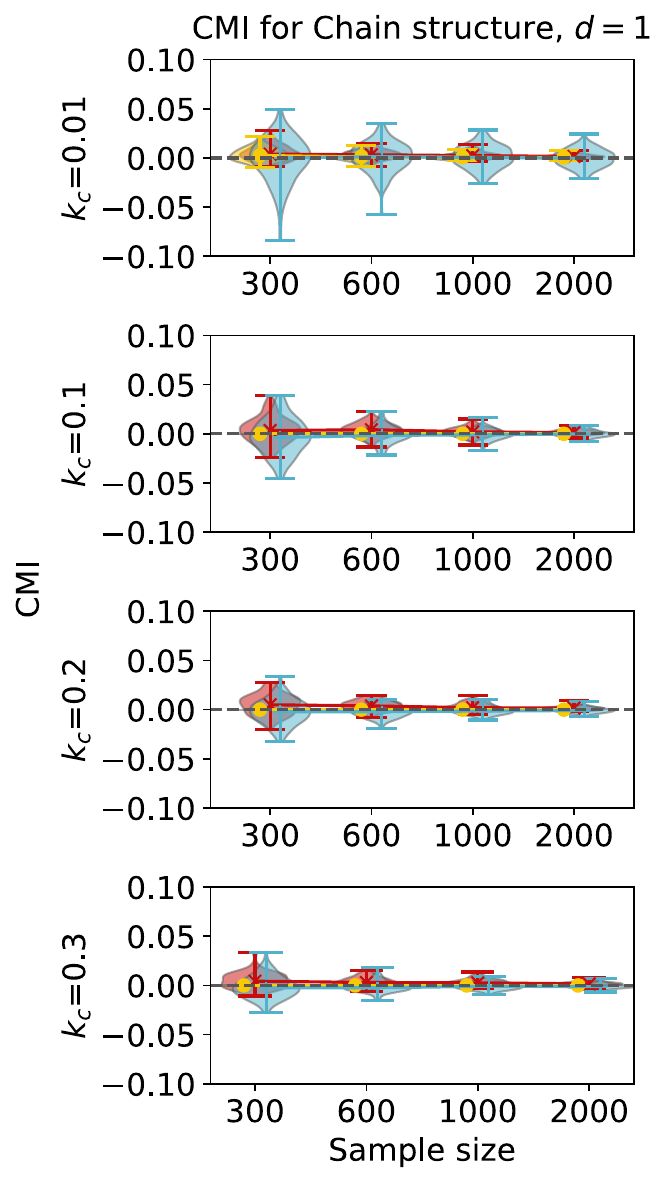}
\includegraphics[width=0.58\linewidth]{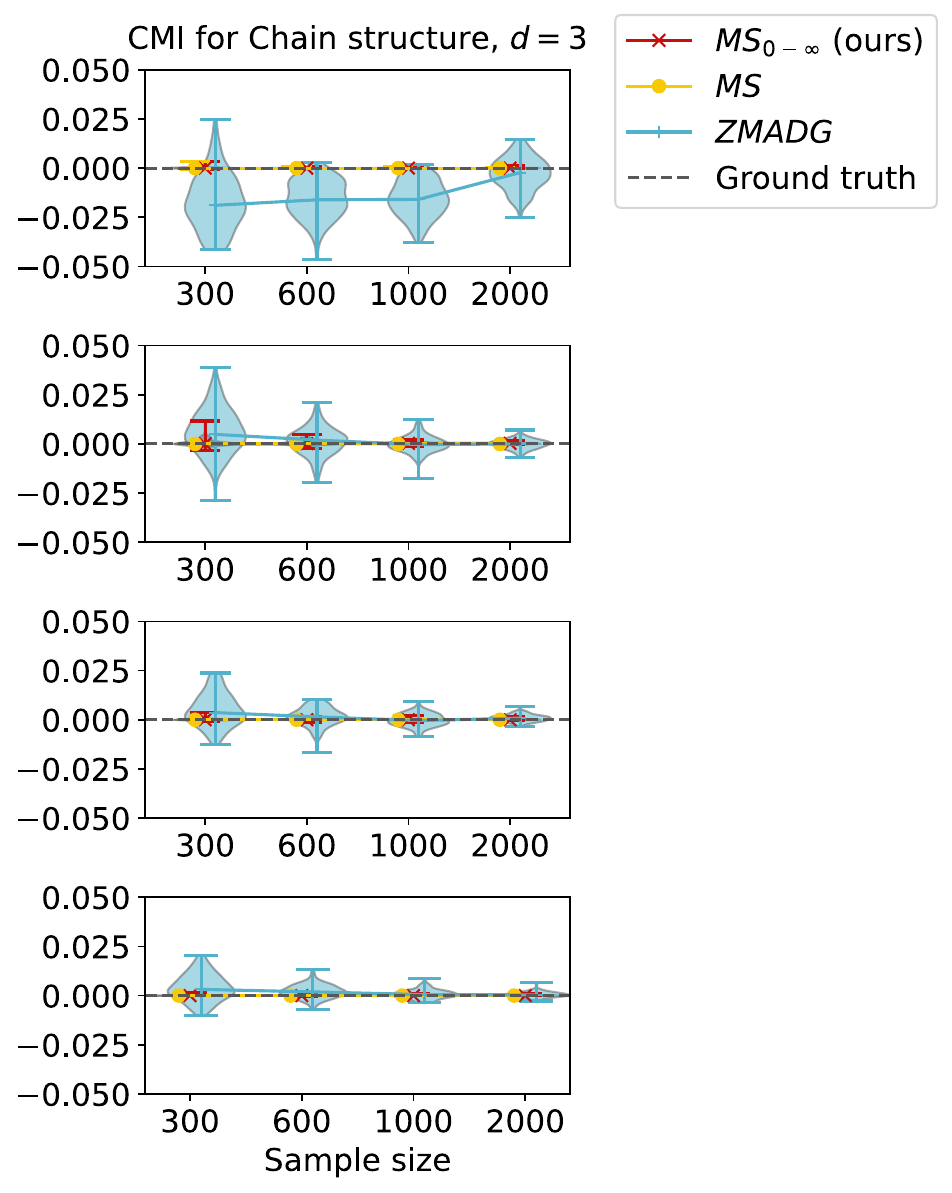}
\caption{Distribution of the CMI estimates for the \underline{"Chain structure"} model with $d=1$ on the left and $d=3$ on the right. Each row shows the results for different $k_c$. Ground truth shown as dashed line.}
\label{fig:cmi_t1_c1}
\end{figure}

\begin{figure}[H]
\centering
\includegraphics[width=0.42\linewidth]{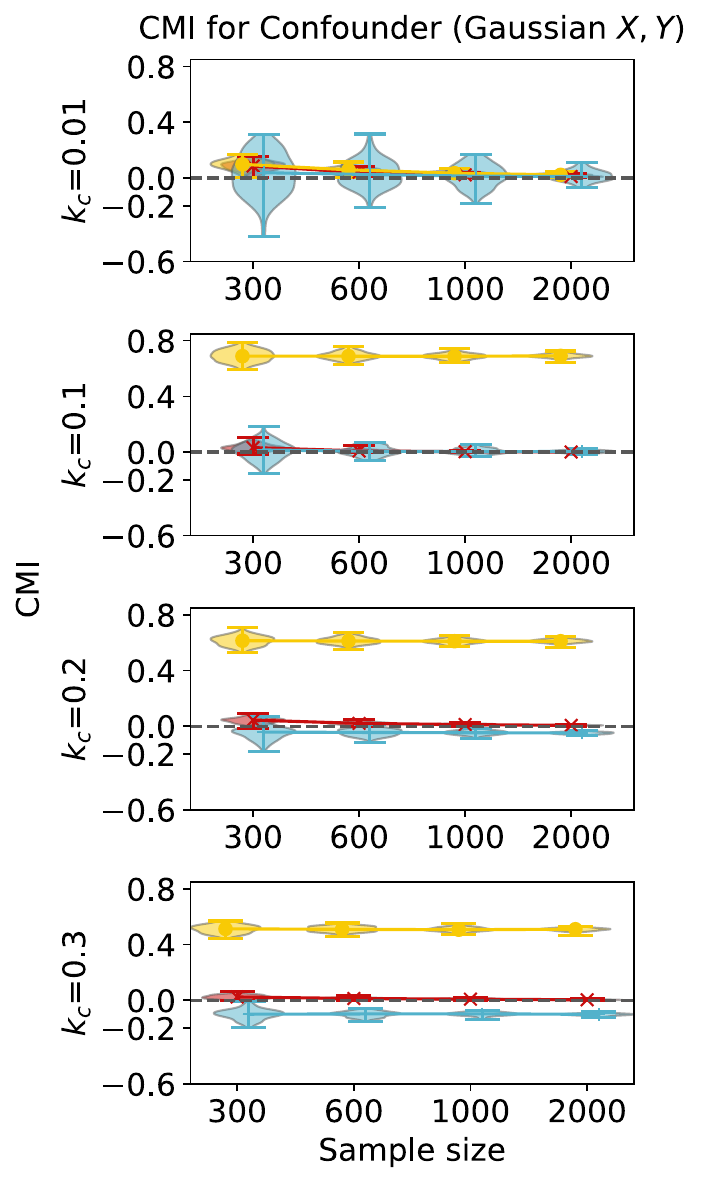}
\includegraphics[width=0.57\linewidth]{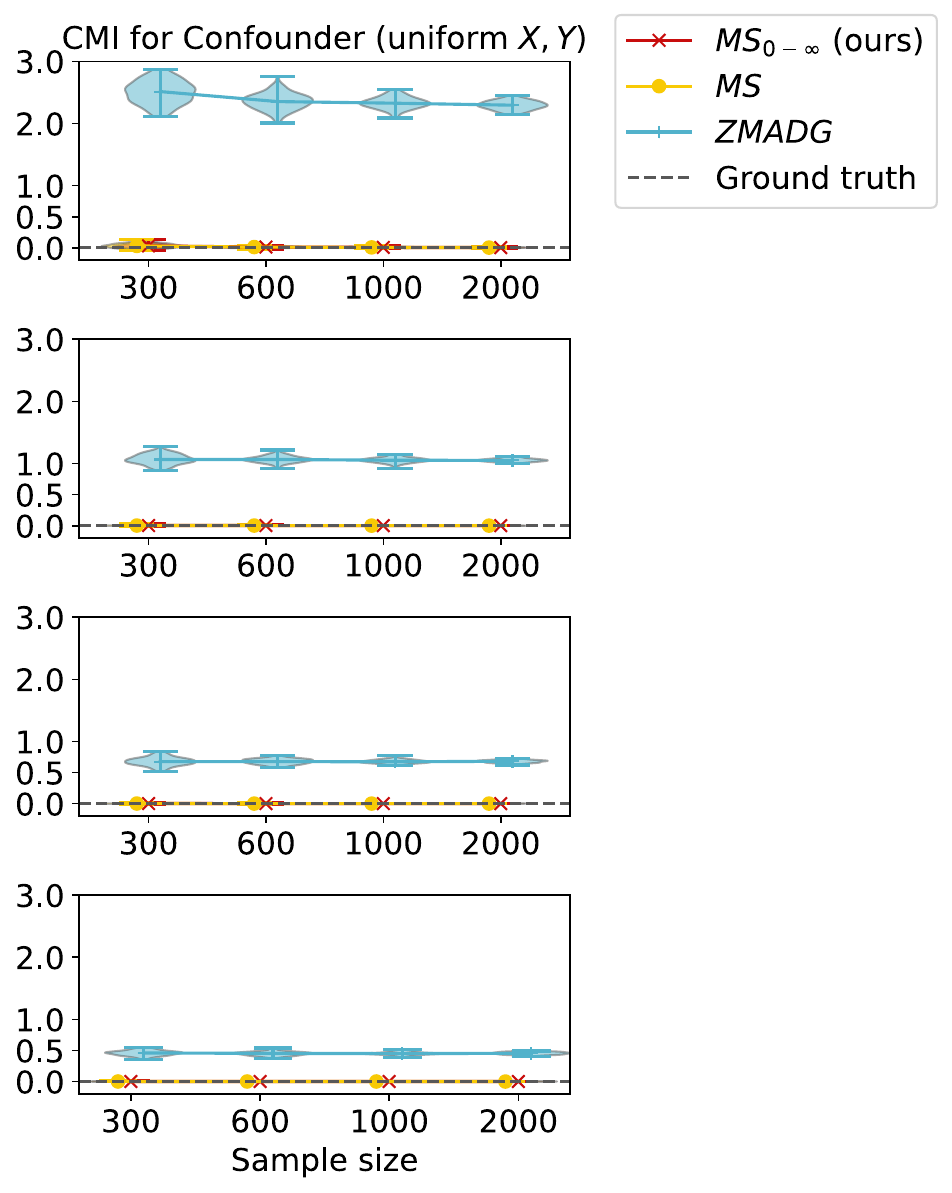}
\caption{Distribution of the CMI estimation for \underline{"Confounder with Gaussian $\X$ and $\Y$"} model on the left and \underline{"Confounder with uniform $\X$ and $\Y$"} model on the right. Ground truth values are shown as the dashed line. Each row shows results for a different $k_c$ value.}
\label{fig:cmi_34}
\end{figure}

\textbf{Results} For the \underline{"Chain structure"} model (Fig.~\ref{fig:cmi_t1_c1}) with $d=1$, the MS$_{0{-}\infty}$ and ZMADG estimators perform comparably well. Notably, the ZMADG estimator has slightly higher variance than MS$_{0{-}\infty}$ and MS, especially for smaller $n$ and higher $d=3$. For the \underline{"Confounder with Gaussian $\X$ and $\Y$"} model (Fig.~\ref{fig:cmi_34}, left), we observe that MS$_{0{-}\infty}$ performs best, while the MS estimator overestimates for $k_c \geq 0.1$. The ZMADG estimator again suffers from higher variance compared to our estimator.  For the \underline{"Confounder with uniform $\X$ and $\Y$"} (Fig.~\ref{fig:cmi_34}, right), besides having high variance, ZMADG wrongly finds a strong conditional dependency between $X$ and $Y$. It seems specific to the uniform distribution, as estimates are correct for the same model using normally distributed data.  For this model, both MS and our estimator perform well.

\subsection{Results of the Numerical Evaluation of the Estimators for the Mixture-type Variable Case}

As mentioned in the main paper, our study did not focus on the mixture-type variable case. However, our MS$_{0-\infty}$ estimator can be used with mixture-type variables. We present here a preliminary evaluation of the CMI estimation for this use case. We keep the experimental setup as for the previous experiments, but do not evaluate the ZMADG estimator, since this method was not designed for use with mixture-type variables. We thus evaluate only the MS and MS$_{0-\infty}$ estimators on data generated from a model inspired by \citet{mesner_shalizi}, defined as follows:

\underline{"Mixture" (adapted from \citet{mesner_shalizi}):} Here, $Z$ is discrete with $Z\sim Bin(1, p=0.3)$. With probability $1-p$, $X$ and $Y$ are drawn from a multivariate Gaussian with correlation coefficient of $0.6$, and with probability $p$, $X\sim \mathcal{U}(\{0,...,4\})$ and $Y\sim\mathcal{U}\left([X, X+2]\right)$. The ground truth is $I(X;Y| Z)=-(1-p) \cdot \ln{(1-0.36)} \cdot 0.5 + p \cdot (\ln{5} - \frac{4}{5} \cdot \ln{2}) = 0.472$. 
\begin{figure}[H]
\centering
\includegraphics[width=0.8\linewidth]{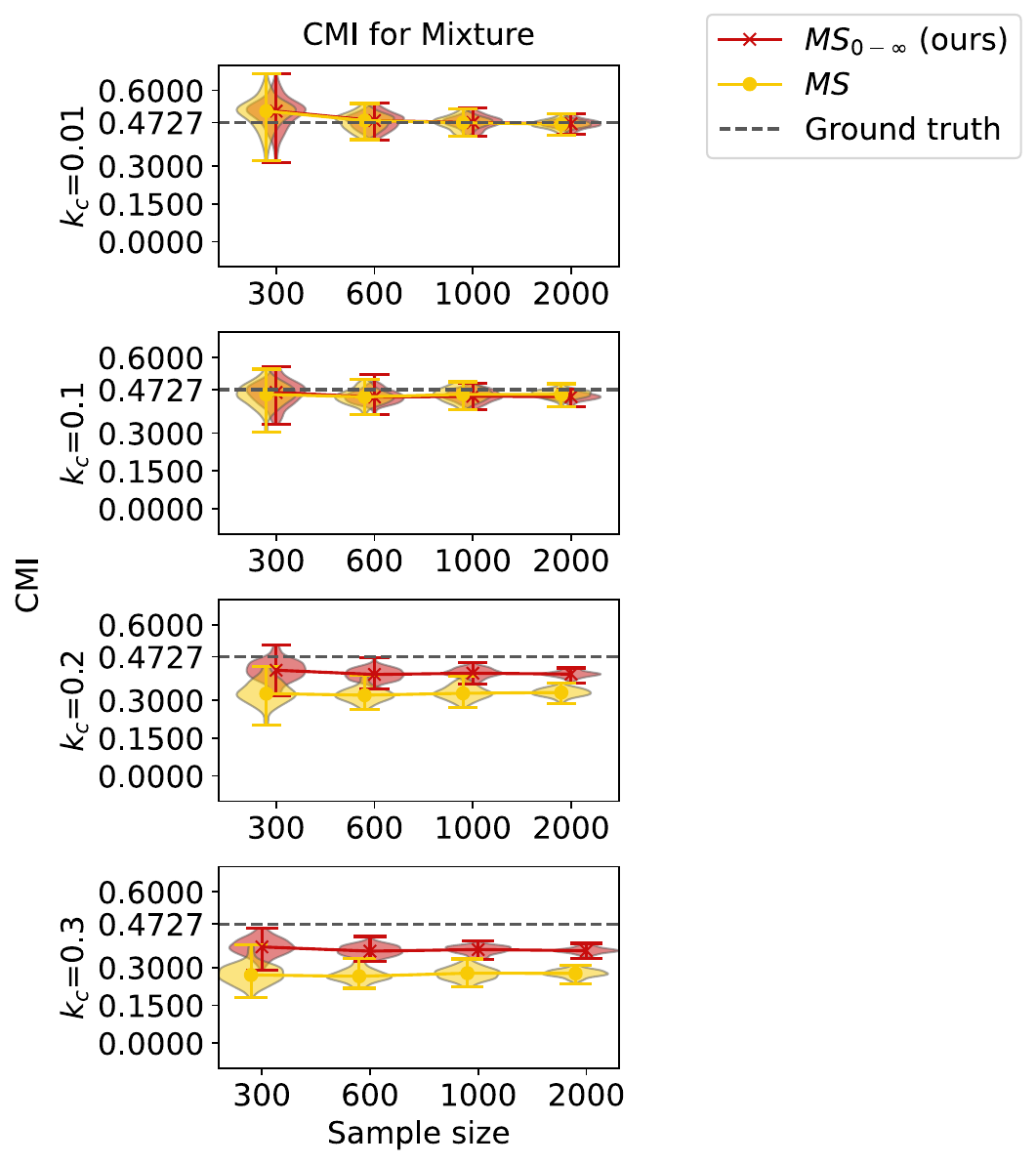}%
\caption{Distribution of CMI estimates for the \underline{"Mixture"} model. Each row shows the results for different $k_c$. Ground truth shown as dashed line.}
\label{fig:cmi_mixture}
\end{figure}

\textbf{Results} As Fig.~\ref{fig:cmi_mixture} shows, our approach performs best across all $k_c$ values. Both MS and MS$_{0-\infty}$ suffer from bias towards 0 for $k_c \geq 0.2$, however, our  estimator is considerably less affected than the MS estimator.
% and consistently returns CMI values closer to the ground truth. 

\section{Choice of Heuristic for $k$}
\label{sec:heuristic}

In contrast to the theoretical setting, where we assume infinite samples are available, we only have access to a finite number of samples in practice. Thus, for large enough $k$ with fixed $n$, it is probable that some of the $k$-th nearest neighbors in the dataset are at $\infty$ distance, i.e., originate from a different cluster. Because  our approach does not allow neighbors from different clusters, a heuristic was necessary to deal with this particular case. We defined and tested three different heuristics. In this Section, we describe the other two heuristics besides the heuristic described in Sec. 3 of the main paper. We then motivate our choice for the heuristic in the main paper, which we based on an empirical comparison of the MS$_{0-\infty}$ estimator's performance using the different heuristics.

\subsection{Two Alternative Heuristics}

We start with the presentation of the other two heuristics, namely the "global" and "cluster-size" heuristics. Note that from now on, we will refer to the heuristic described in Sec. 3 of the main paper as the "local" heuristic. 

\subsubsection{"Global" heuristic}

The "global" heuristic defines a "global" $k$ as a fraction of the number of samples $n$. If the distance to the $k$-th nearest neighbor is $\infty$, then, instead of enforcing $k$ nearest neighbours for all sample points, we allow for the following adaptiveness: If the $k$-th NN of $w_i$ is at distance $\infty$ from $w_i$ (that is, if $n_{cl}^i \leq k$ with $n_{cl}^i$ the number of points in the cluster of $w_i$), then for this $i$ we replace $k$ by $k_i = \lfloor k_c \cdot n^i_{cl}\rfloor$. Explicitly: For all $i$ let $k^{0-\infty}_i = k$ if $k+ 1 \leq n^i_{cl}$ and $k^{0-\infty}_i = \lfloor k_c \cdot n^i_{cl}\rfloor$ else. Thus, in effect, all considered nearest neighbours of $w_i$ come from the same cluster as $w_i$. 

\subsubsection{"Cluster-size" heuristic}

This heuristic still uses a "global" $k$ as a fraction of the number of samples $n$. However, if the distance to the $k$-th nearest neighbor is $\infty$, the "cluster-size" heuristic deals with this case by simply setting $k_i = n^i_{cl}$, where $n^i_{cl}$ is the number of samples in the cluster of point $i$, defined as previously described.

\subsection{Numerical Evaluation of the Three Heuristics}

\textbf{Experimental setup} We run numerical experiments to compare the bias and variance of our MS$_{0-\infty}$ estimator using the three different heuristics, alongside the bias and variance of the MS and ZMADG estimators. We keep the same experimental setting as in the previous experiments (described in the Sec. 4 of the main paper and Sec.~\ref{app:further_Estim} of the SM) and evaluate results on the "Independent $Z$" and "Chain structure" models.

\begin{figure}[H]
% \centering
\includegraphics[width=0.55\linewidth]{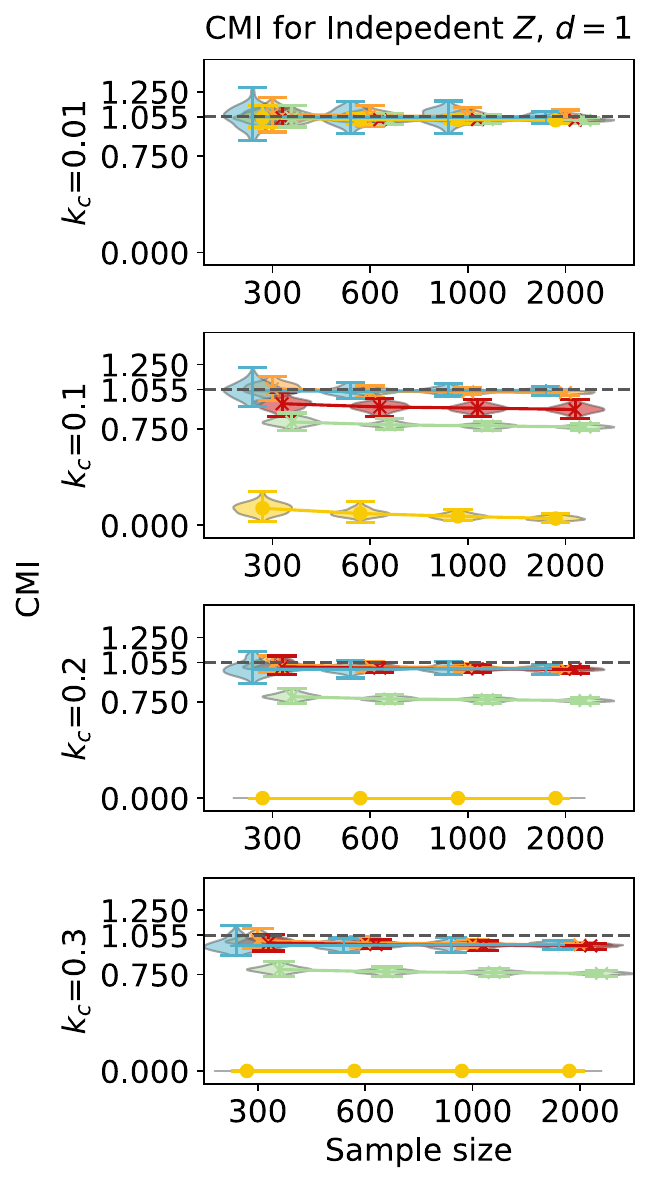}
\\
\includegraphics[width=0.95\linewidth]{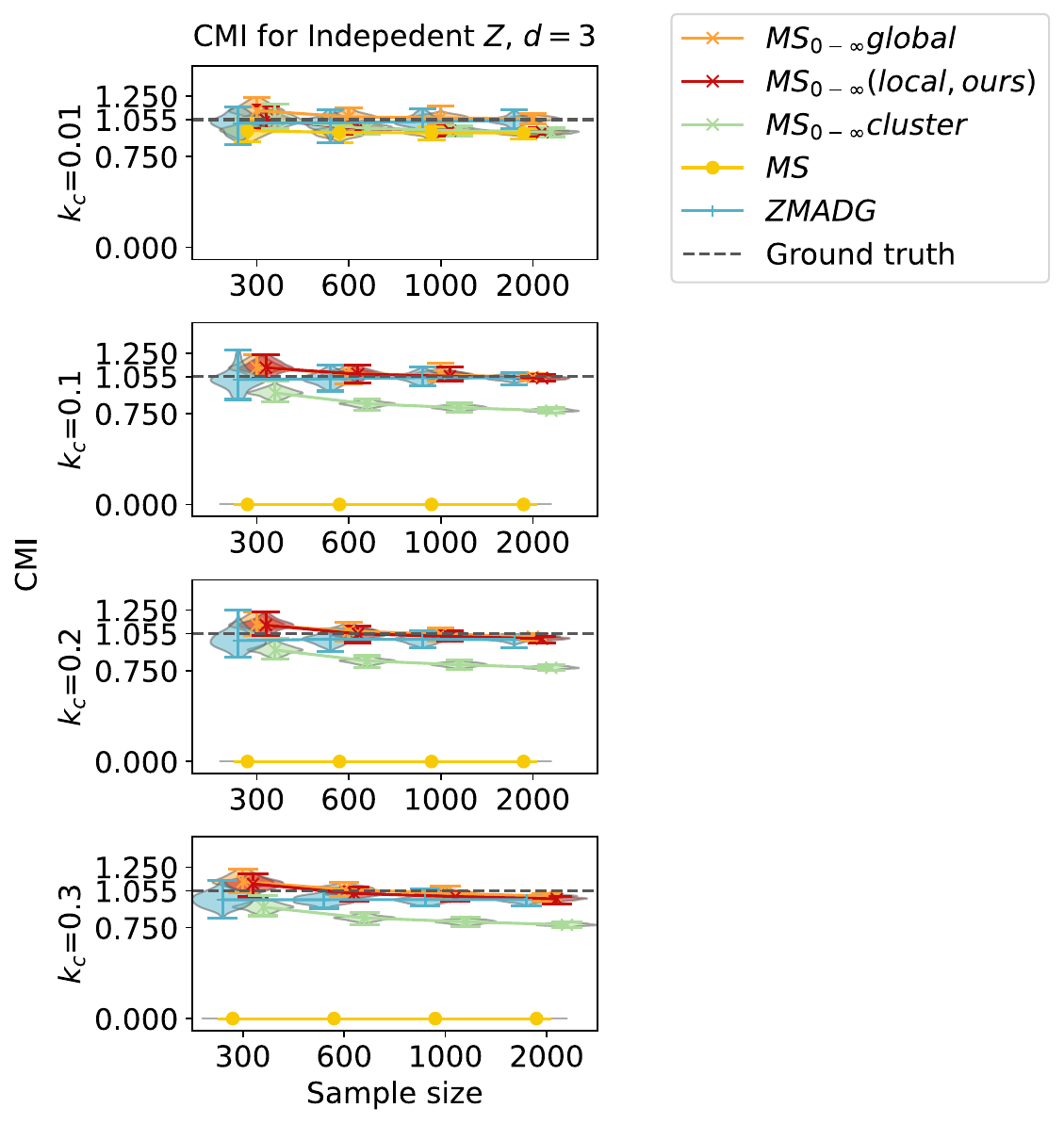}  

\caption{CMI estimation results for the \underline{"Independent $Z$"} model using the three different heuristics for setting the hyperparameter $k$: the "local" heuristic (used in the main paper, MS$_0-\infty$ local), the "global" heuristic (MS$_0-\infty$ global), and the "cluster-size" heuristic (MS$_0-\infty$ cluster). The ground truth CMI values are indicated by the dashed line. The first four rows show results for different $k_c$ values for the model with $d=1$, and the last four rows show results for different $k_c$ values for the model with $d=3$.}
\label{fig:k_heuristic_1}
\end{figure}

\begin{figure}[H]
% \centering
\includegraphics[width=0.55\linewidth]{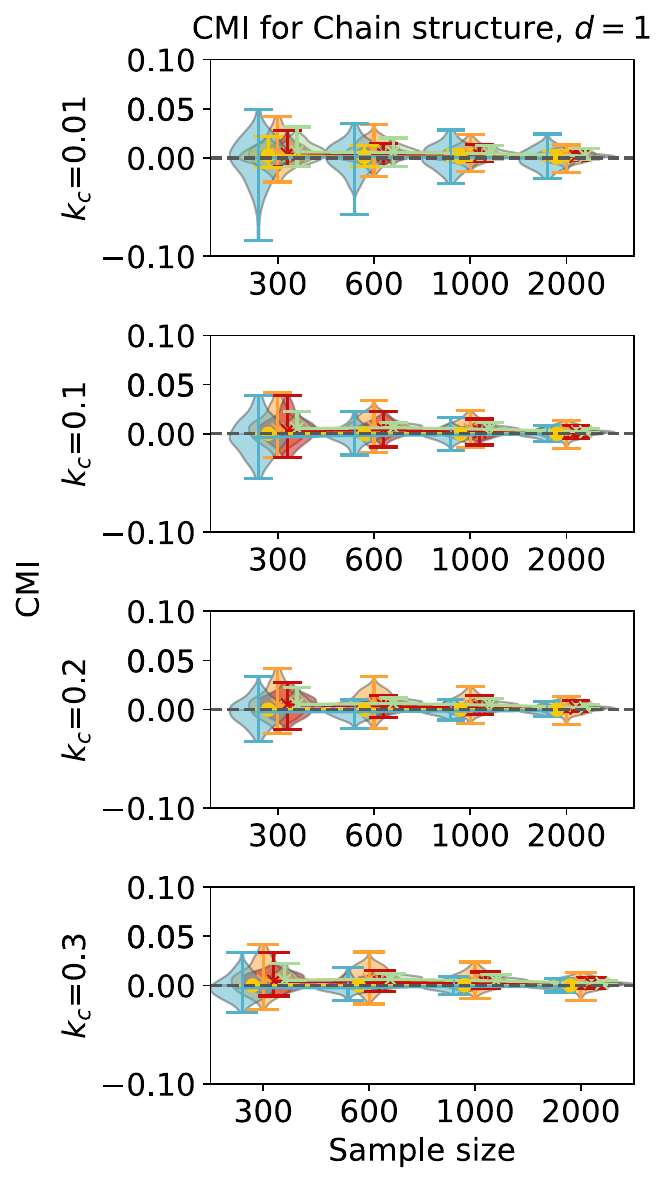}
\\
\includegraphics[width=0.95\linewidth]{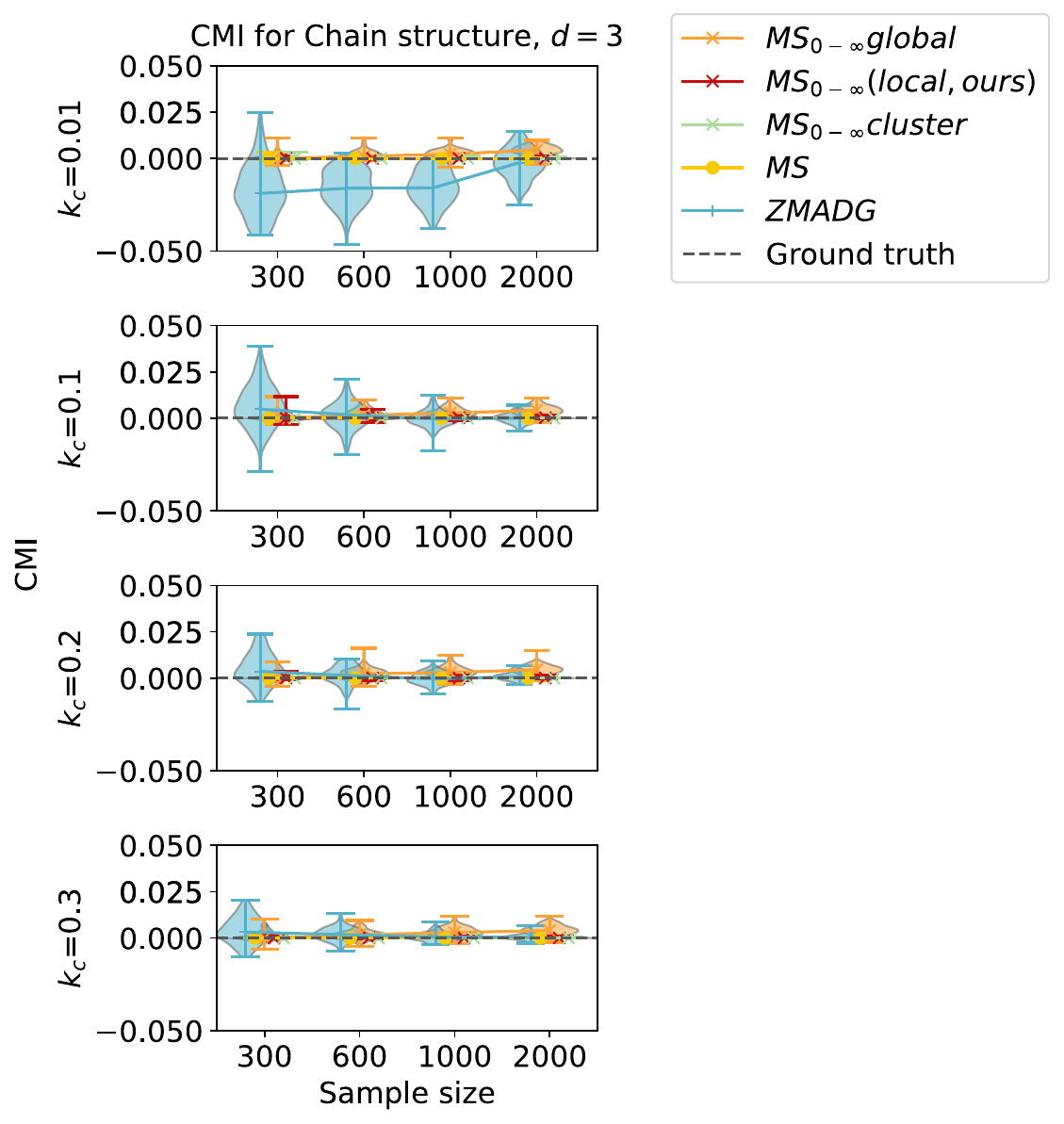}

\caption{CMI estimation results for the \underline{"Chain structure"} model using the three different heuristics for setting the hyperparameter $k$: the "local" heuristic (MS$_0-\infty$ local), the "global" heuristic (MS$_0-\infty$ global), and the "cluster-size" heuristic (MS$_0-\infty$ cluster). The ground truth CMI values are indicated by the dashed line.  The first four rows show results for different $k_c$ values for the model with $d=1$, and the last four rows show results for different $k_c$ values for the model with $d=3$.}
\label{fig:k_heuristic_2}
\end{figure}

\textbf{Results} The violin plots in Figures~\ref{fig:k_heuristic_1} and ~\ref{fig:k_heuristic_2} show the results of the CMI estimation using the three heuristics presented above. We observe that the "cluster-size" heuristic suffers from bias towards zero. This is expected because, when $k=n^i_{cl}$, the distance to the $k$-th nearest neighbor is equal to the distance from point $i$ to the farthest point in its respective cluster. Thus, for the subspaces $\X\Y, \X\Z$ and $\Z$, the number of counted neighbors is equal to $n^i_{cl}$ with high probability, which results in an estimation equal to or close to 0. The "global" heuristic has the highest variance across the different heuristics, yet still has lower variance compared to the ZMADG estimator. The "global" approach also slightly suffers from bias for higher dimesionality, for example, for the "Chain structure" model with $d=3$. The best bias-variance trade-off is obtained using the "local" heuristic, where both bias and variance are smaller compared to the other two heuristics. This motivates our choice to use this heuristic for the experiments in the main paper. 

\subsection{Discussion on the Choice of $k_c$}

Generally, $k$-NN methods benefit from a value of $k$ that is neither too high or too low: A small $k$ leads to lower bias at the cost of increased variance, while larger $k$ values lead to low variance but increased bias \cite{Kraskov2004EstimatingMI}. From the CMI estimation results and the outcome of the conditional independence tests, we observe that a larger $k_c$, e.g. $k_c > 0.2$ is beneficial for our method, especially for the case of weak dependency and smaller sample size. However, as the sample size increases, e.g. $n=2000$, we observe that a smaller $k_c$, e.g. $k_c=0.2$ performs better. We thus generally recommend $k_c \geq 0.1$, with larger $k_c$ when the number of samples is small or there are many clusters, as an appropriate choice that reduces the bias and variance of the estimation.

\section{Further Results of the Numerical Evaluation of the CIT}

We present further results of the CIT evaluation for the configurations presented in Sec. 5 of the main paper. As in Sec. 5, we show plots of the FPR and TPR for different $k_c$ values. In the following sections, we also present additional plots of the true null and the permuted CMI distributions for the different $k_c$ values. These distribution plots allow us to investigate whether the FPR/TPR reflects the desired behavior of the tests: The true null and the permuted distributions should have CMI values distributed around 0. In contrast, in the dependent case, the CMI values should be larger than 0, and their distribution should have minimal overlap if the null hypothesis does not hold. While we investigated the CMI distribution plots of all configurations, we refrain from adding all plots in the SM for length reasons and exemplify using one set of plots. 

\subsection{Computation of the Confidence Intervals for the TPR/FPR Plots}

Before presenting the results, we first describe how we compute the error bars of the CIT plots, which represent the $95\%$ confidence interval of the false positive rate (FPR) and true positive rate (TPR).

The confidence intervals are obtained by modelling the false and true positives as distributed according to the binomial distribution. We describe the computation of the confidence interval for the FPR, and obtain the confidence interval for the TPR analogously. For a given model and a set of values of the CIT parameters, the probability to obtain a false positive in the $n_{rep}=100$ repetitions of our experiments is $p_{FP}$ (ideally, $p_{FP}=\alpha$). Under the assumption that the repetitions are independent, the random variable that describes the number of false positives, $FP$, is distributed according to the binomial distribution:

\begin{equation}FP \sim Bin(n_{rep}, p_{FP}).
\end{equation}

We can estimate $p_{FP}$ as the empirical fraction of false positives that we have obtained in our repetitions: $\hat{p}_{FP}=\frac{\#FP}{n_{rep}}=FPR$.

We now wish to obtain a confidence interval for $\hat{p}_{FP}$, i.e., find the lower and upper bounds $p_L, p_U$ of the confidence interval such that $P(p_L(FP) < \hat{p}_{FP} < p_U(FP)) = 1 - \alpha$. Since $FP \sim Bin(n_{repetitions}, p_{FP})$, and supposing we have observed $FP=fp$, we obtain $p_L$ and $p_U$ by numerically solving the following two equations:

\begin{equation}
\begin{aligned}
    \sum_{k=fp}^{n_{repetitions}} & (p_L)^k \cdot (1-p_L)^{(n_{repetitions}-k)} \\ = & 1 - CDF(fp, n_{repetitions}, p_L)\\
     = &\frac{1-\alpha}{2}
\end{aligned}
\end{equation}

\begin{equation}
\begin{aligned}
    \sum_{k=0}^{fp} & (p_U)^k \cdot (1-p_U)^{(n_{repetitions}-k)} \\ 
    = & CDF(fp, n_{repetitions}, p_U) \\
    = & \frac{1-\alpha}{2}
\end{aligned}
\end{equation}

Here $CDF$ is the cumulative distribution function of the binomial distribution.

\subsection{"Confounder" Model}

Here, we present further results for the "Confounder" model.

\begin{figure}[H]
\centering
\includegraphics[width=0.93\linewidth]{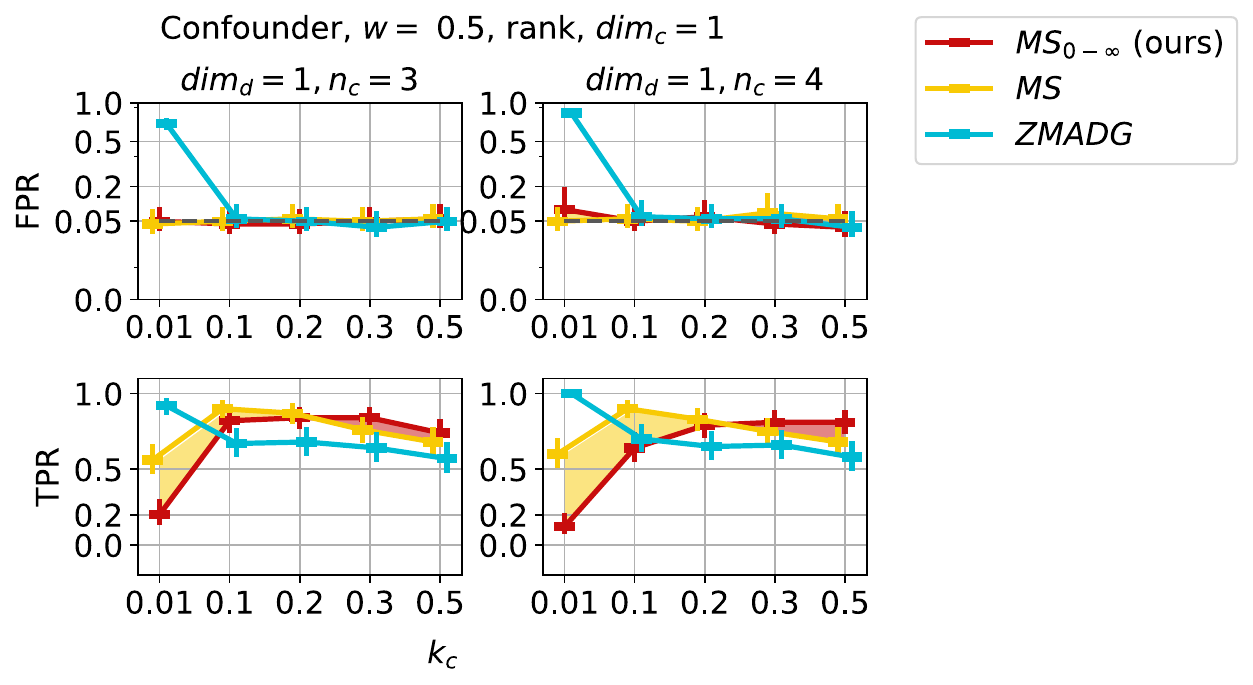}%

\caption{False positive rate (FPR, ideally under 0.05) and true positive rate (TPR, higher is better, with 1 best) for the \underline{"Confounder"} model where $Z$ has $dim_c=1, dim_d=1$, with rank preprocessing for the continuous variables and coupling factor in the dependent case $w=0.5$.}
\label{fig:cit_cf_1}
\end{figure}

\begin{figure}[H]
\centering
\includegraphics[width=0.48\linewidth]{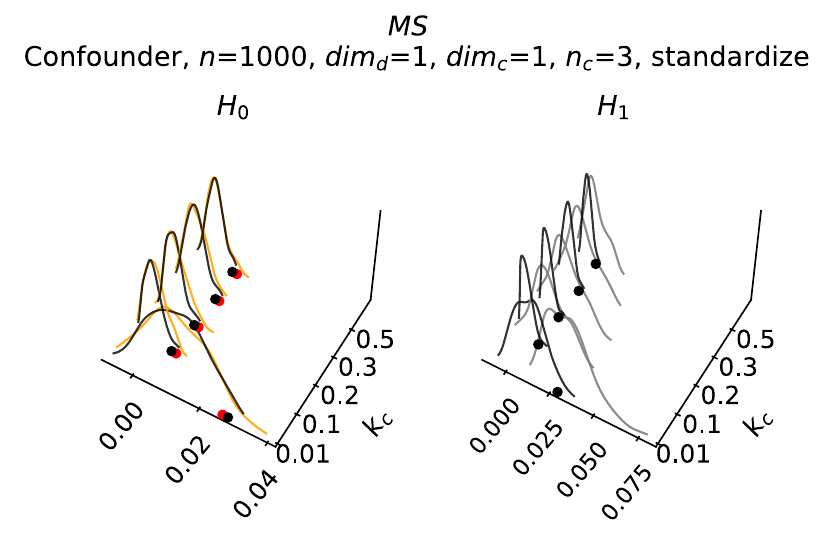}%
\includegraphics[width=0.48\linewidth]{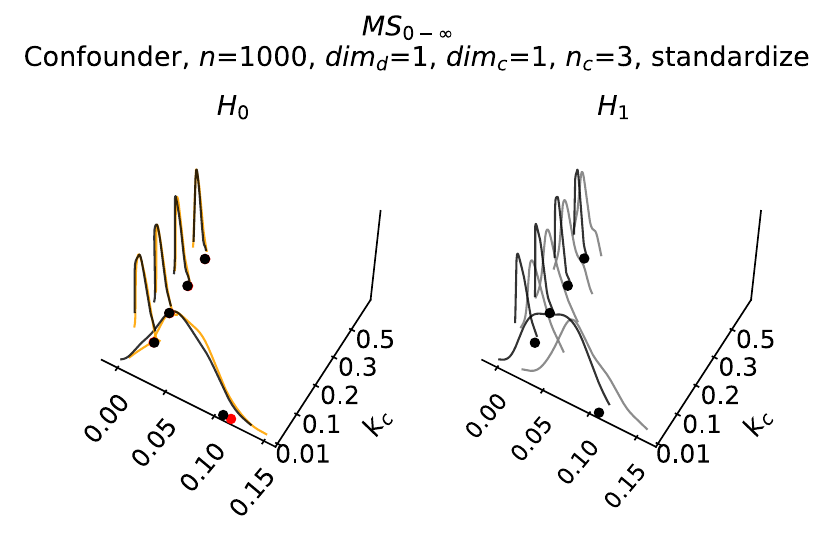}%
\\
\includegraphics[width=0.48\linewidth]{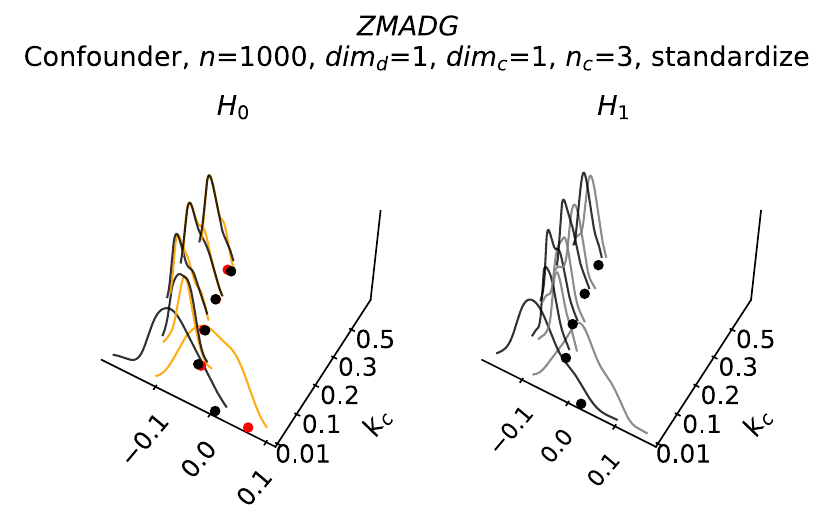}%
\caption{The true null and permuted distributions of the CMI estimated with the MS, MS$_{0-\infty}$ and ZMADG estimators for the \underline{"Confounder"}-model with $n=1000$ where $Z$ has $dim_c=1, dim_d=1$ and $n_c=3$ with standardization preprocessing. The left figure of each plot pair shows the true null distribution under $H_0$ as the orange line with the red dot indicating the $95\%$ quantile, and the permuted null distribution is shown as the black line, with the black dot indicating the $95\%$ quantile. On the right, we show the true distribution under $H_1$ as the grey line, and the permuted distributions as the black, with the black dot indicating the $95\%$ quantile.}
\label{fig:dist_1}
\end{figure}

\begin{figure}[H]
\centering
\includegraphics[width=0.48\linewidth]{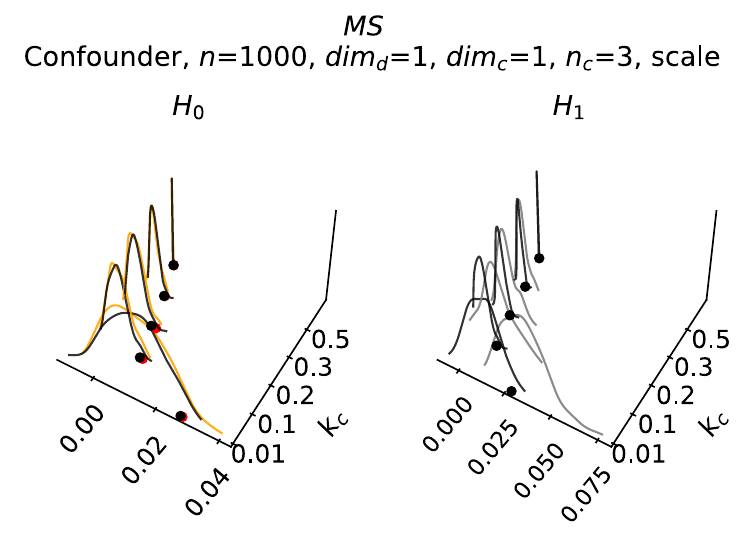}%
\includegraphics[width=0.48\linewidth]{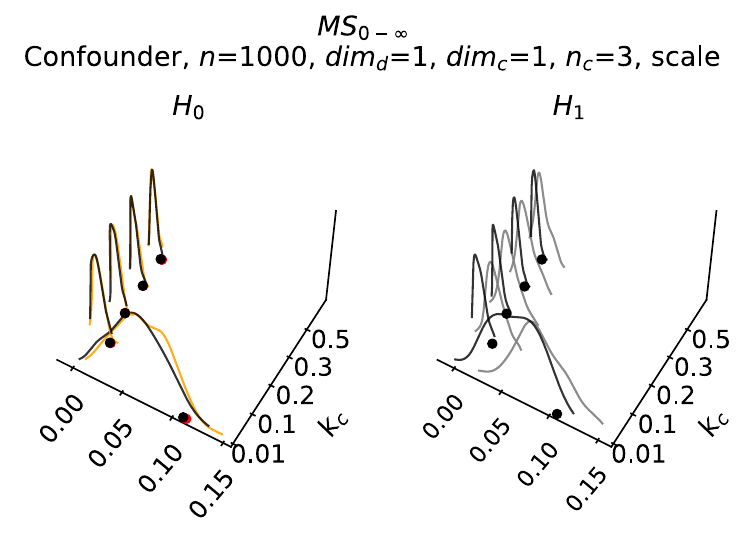}%
\\
\includegraphics[width=0.48\linewidth]{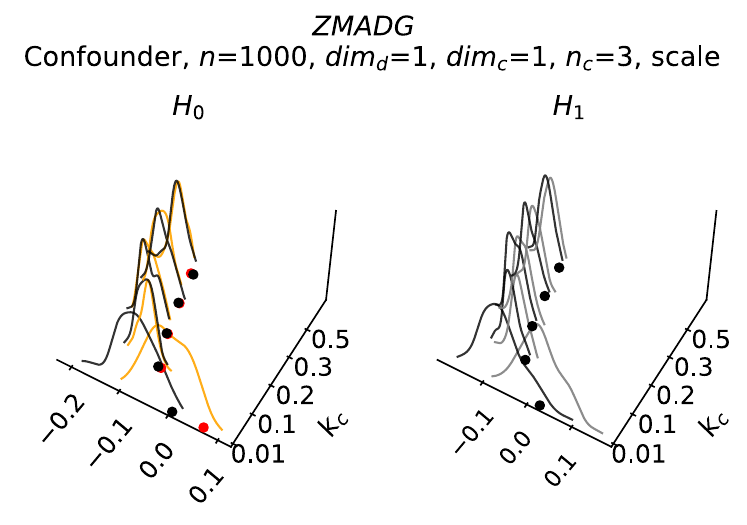}%
\\
\includegraphics[width=0.48\linewidth]{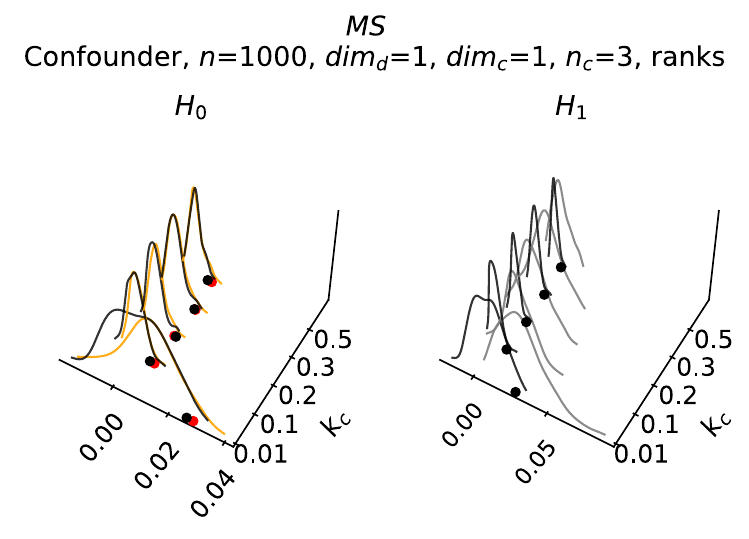}%
\includegraphics[width=0.48\linewidth]{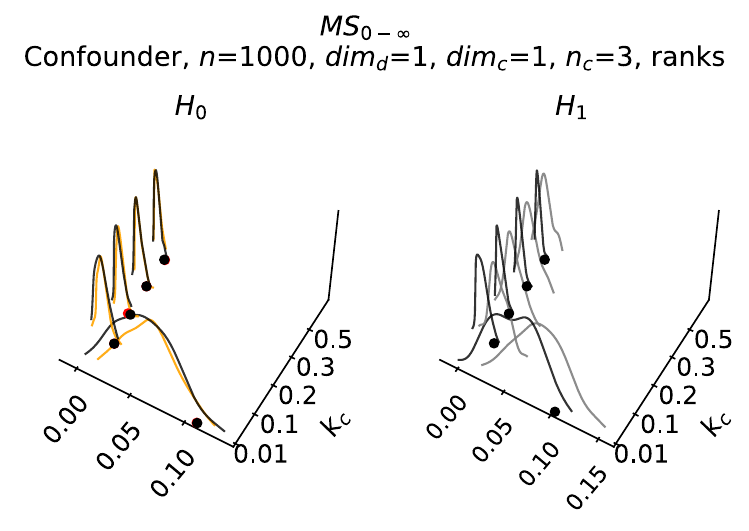}%
\\
\includegraphics[width=0.48\linewidth]{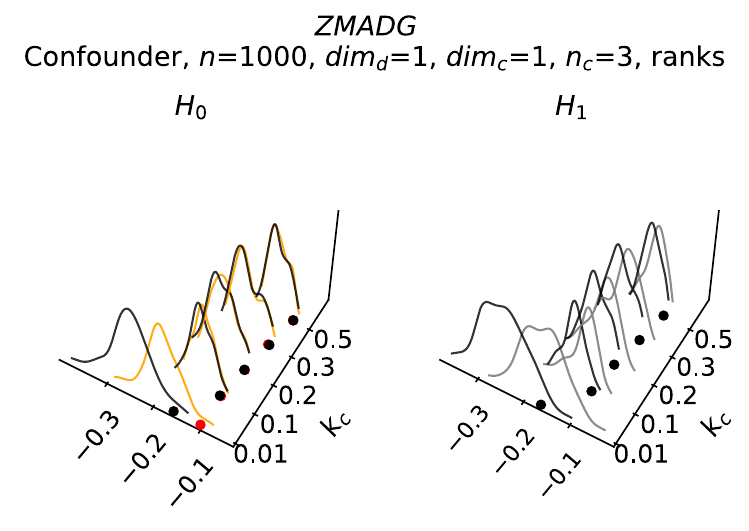}%
\caption{The true null and permuted distributions of the CMI estimated with the MS, MS$_{0-\infty}$ and ZMADG estimators for the \underline{"Confounder"}-model with $n=1000$ where $Z$ has $dim_c=1, dim_d=1$ and $n_c=3$ with scaling to $(0,1)$ and rank preprocessing. The distributions are depicted as described in Fig.~\ref{fig:dist_1}.}
\label{fig:dist_1_2}
\end{figure}

\begin{figure}[H]
\centering
\includegraphics[width=0.47\linewidth]{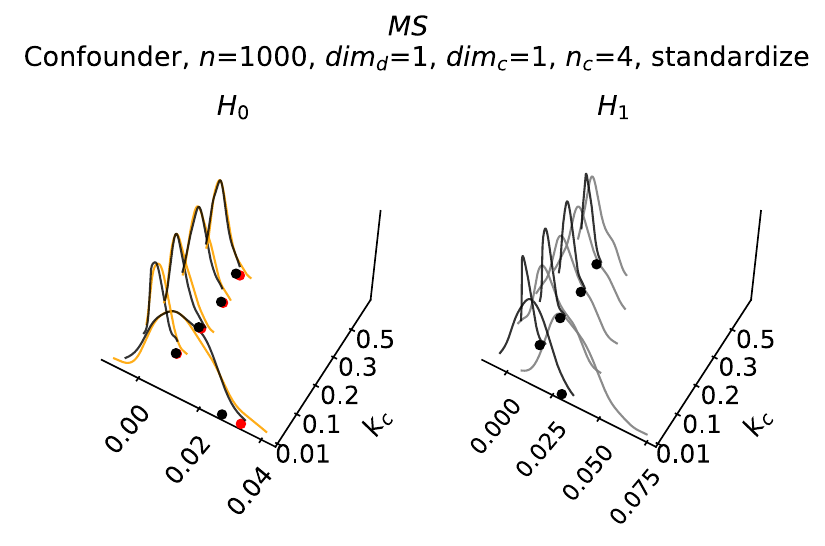}%
\includegraphics[width=0.48\linewidth]{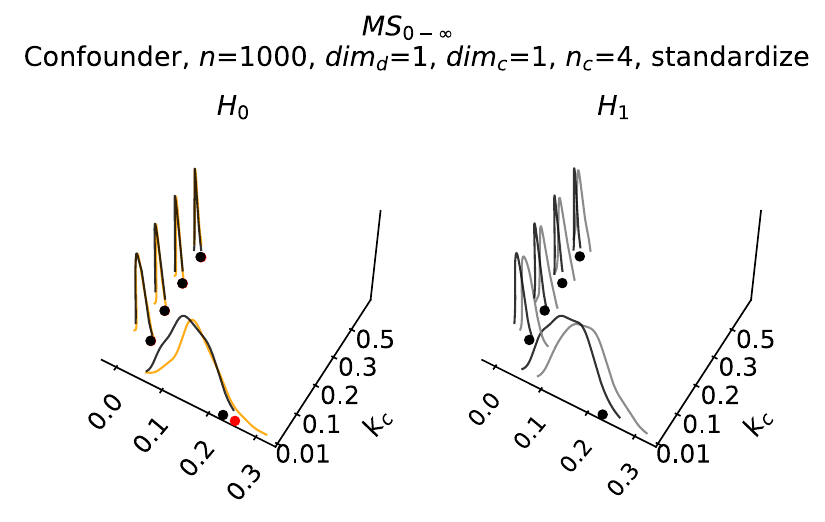}%
\\
\includegraphics[width=0.48\linewidth]{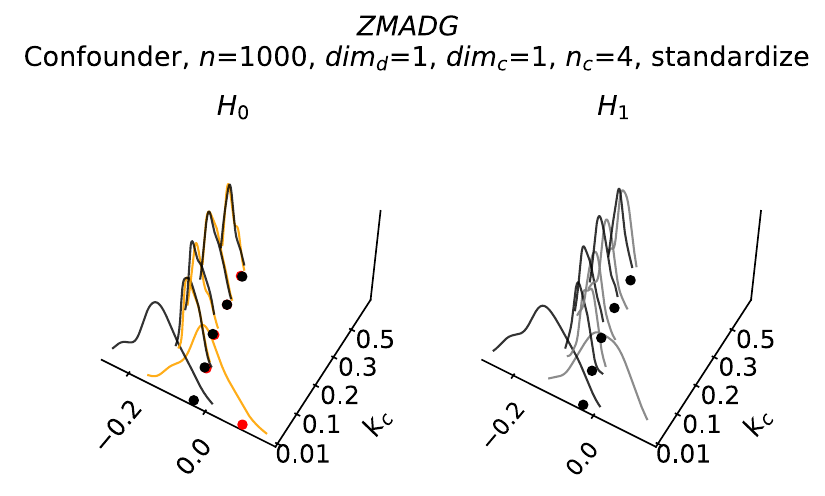}%

\caption{The true null and permuted distributions of CMI estimated with the MS, MS$_{0-\infty}$ and ZMADG estimators for the \underline{"Confounder"}-model where $Z$ has $dim_c=1, dim_d=1$ and $n_c=4$ with standardization preprocessing. The distributions are depicted as described in Fig.~\ref{fig:dist_1}.}
\label{fig:dist_1_3}
\end{figure}

\begin{figure}[H]
\centering
% \includegraphics[width=0.48\linewidth]{figures/dists/confounder/dists_MS_1000_1_1_0.5_3_standardize.pdf}%
% \includegraphics[width=0.5\linewidth]{figures/dists/confounder/dists_MSinf_1000_1_1_0.5_3_standardize.pdf}%
% \\
% \includegraphics[width=0.5\linewidth]{figures/dists/confounder/dists_ZMADG_1000_1_1_0.5_3_standardize.pdf}%
% \\
\includegraphics[width=0.47\linewidth]{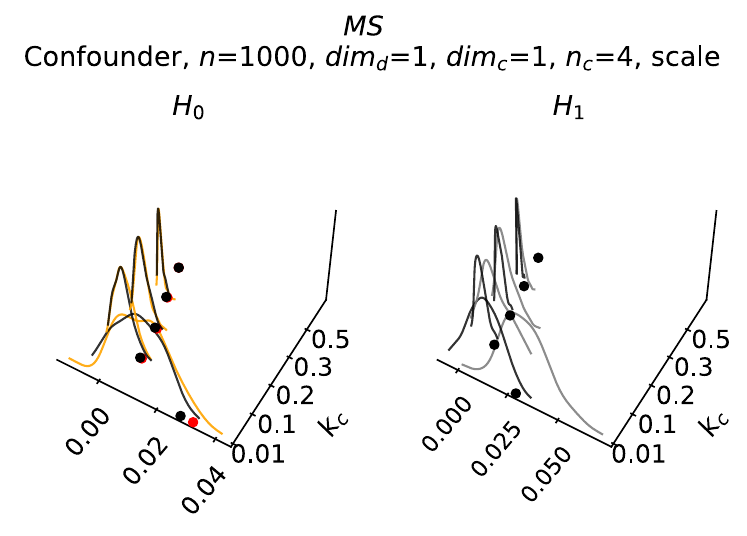}%
\includegraphics[width=0.48\linewidth]{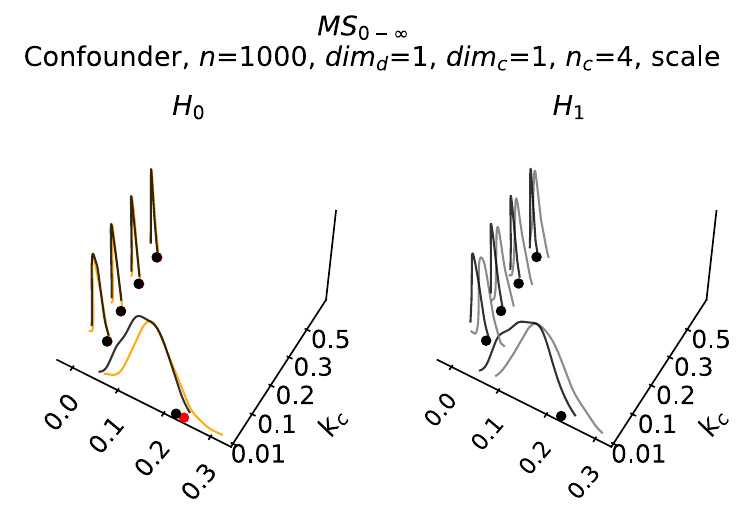}%
\\
\includegraphics[width=0.48\linewidth]{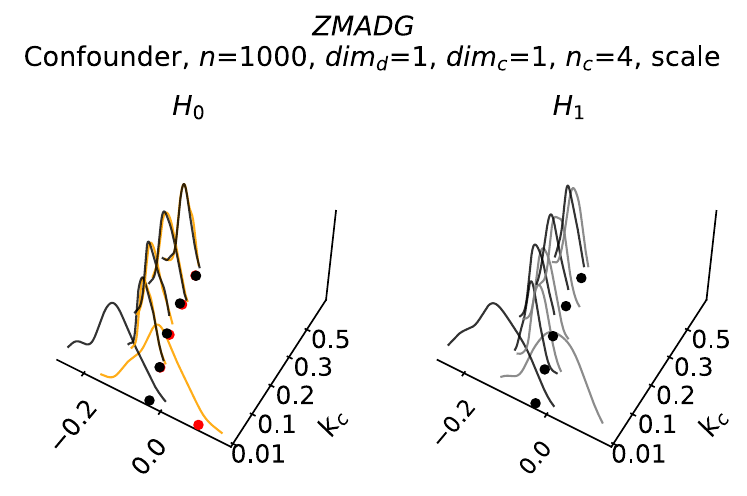}%
\\
\includegraphics[width=0.47\linewidth]{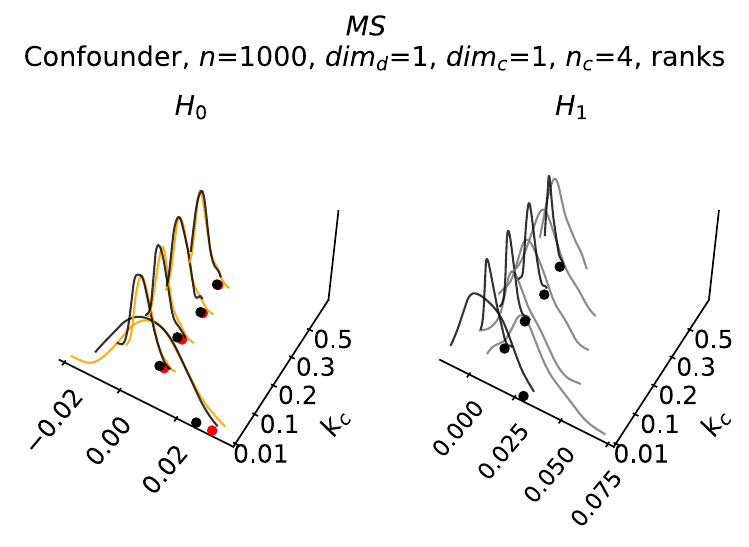}%
\includegraphics[width=0.48\linewidth]{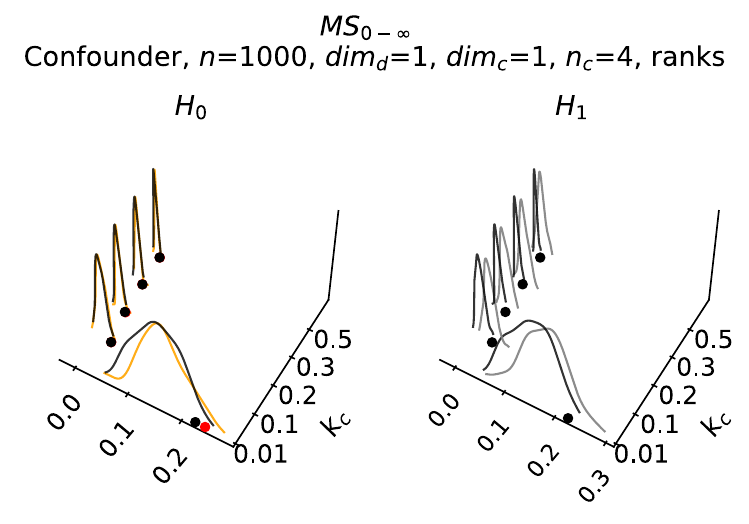}%
\\
\includegraphics[width=0.48\linewidth]{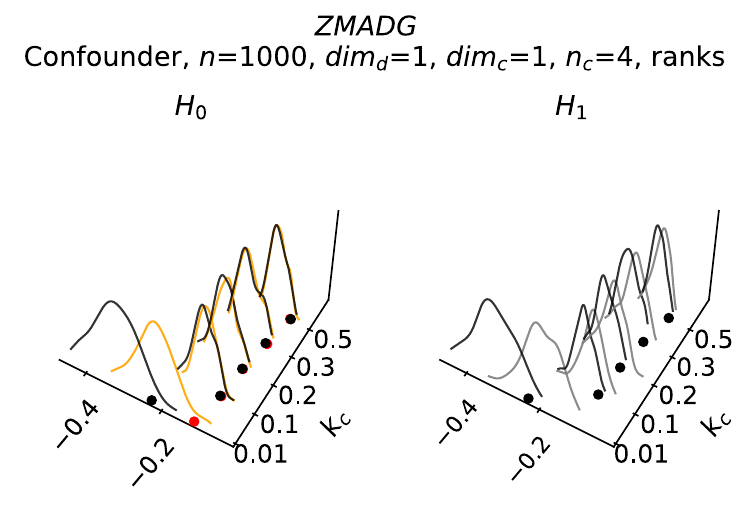}%
\caption{The true null and permuted distributions of CMI estimated with the MS, MS$_{0-\infty}$ and ZMADG estimators for the \underline{"Confounder"}-model with $n=1000$ where $Z$ has $dim_c=1, dim_d=1$ and $n_c=4$ with scaling to $(0, 1)$ and rank preprocessing. The distributions are depicted as described in Fig.~\ref{fig:dist_1}.}
\label{fig:dist_1_4}
\end{figure}

\textbf{For the "Confounder"-model with $dim_c=1$ and $dim_d=1$ and sample size $n=1000$}, the CIT with rank transformation (Fig.~\ref{fig:cit_cf_1}) behaves similarly to the CIT with standardization. Inspecting the distributions of the CMI values (see Fig.~\ref{fig:dist_1} and \ref{fig:dist_1_2} for $n_c=3$ and Fig.~\ref{fig:dist_1_3} and ~\ref{fig:dist_1_4} for $n_c=4$), we observe that MS and MS$_{0-\infty}$ estimators behave similarly for $k_c > 0.1$, while ZMADG suffers from slight negative bias.

\begin{figure}[H]
\centering
\includegraphics[width=0.47\linewidth]{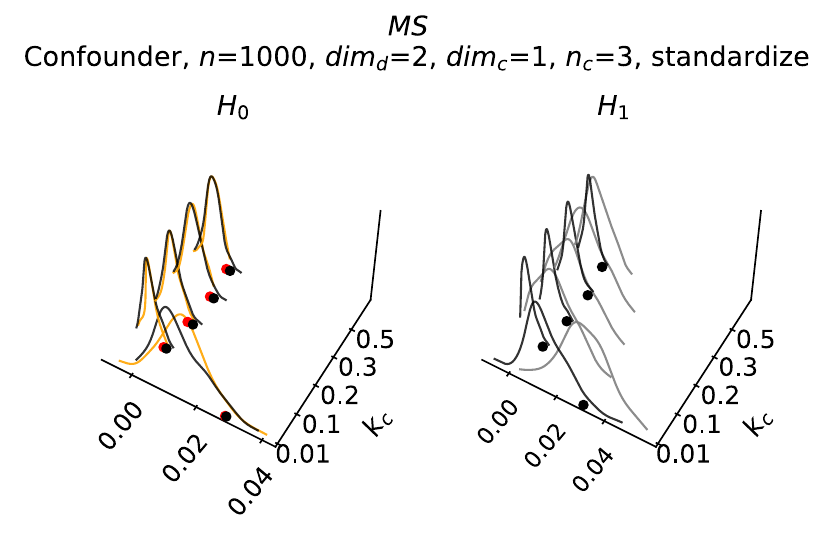}%
\includegraphics[width=0.48\linewidth]{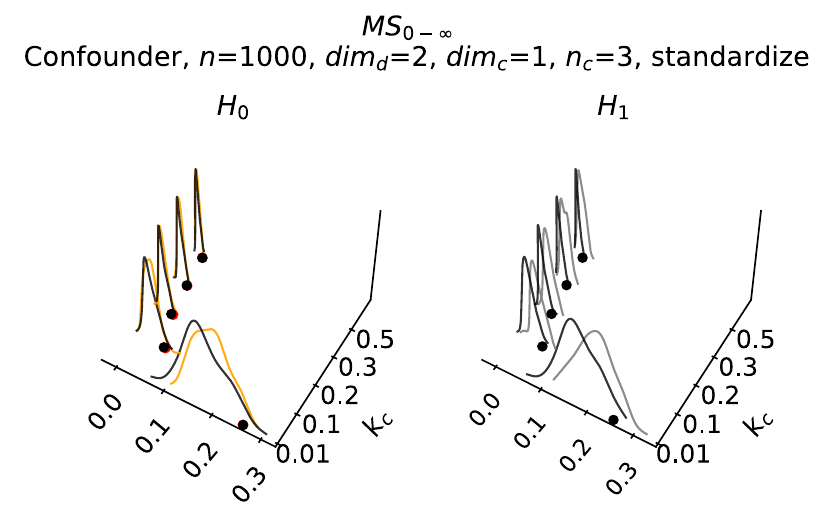}%
\\
\includegraphics[width=0.48\linewidth]{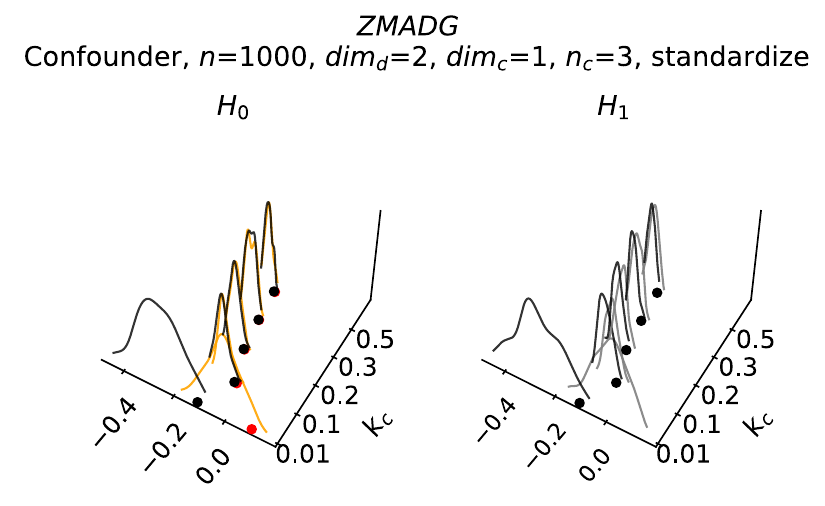}%
\caption{The true null and permuted distributions of CMI estimated with the MS, MS$_{0-\infty}$ and ZMADG estimators for the \underline{"Confounder"}-model with $n=1000$ where $Z$ has $dim_c=1, dim_d=2$ and $n_c=3$ with standardization preprocessing. The distributions are depicted as described in Fig.~\ref{fig:dist_1}.}
\label{fig:dist_2_1}
\end{figure}

\begin{figure}[H]
% \centering
\includegraphics[width=0.65\linewidth]{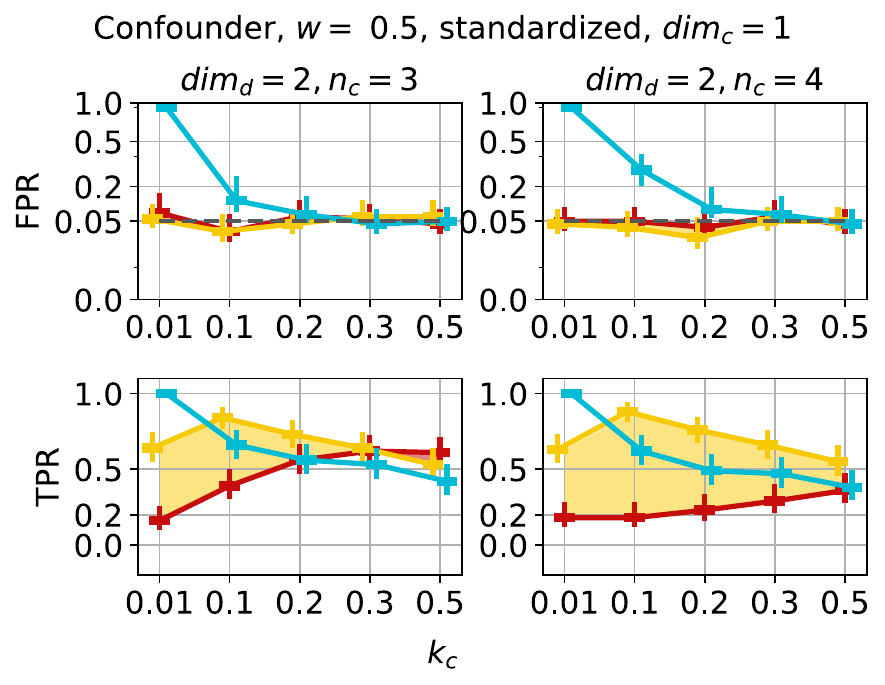}
\\ 
\includegraphics[width=0.93\linewidth]{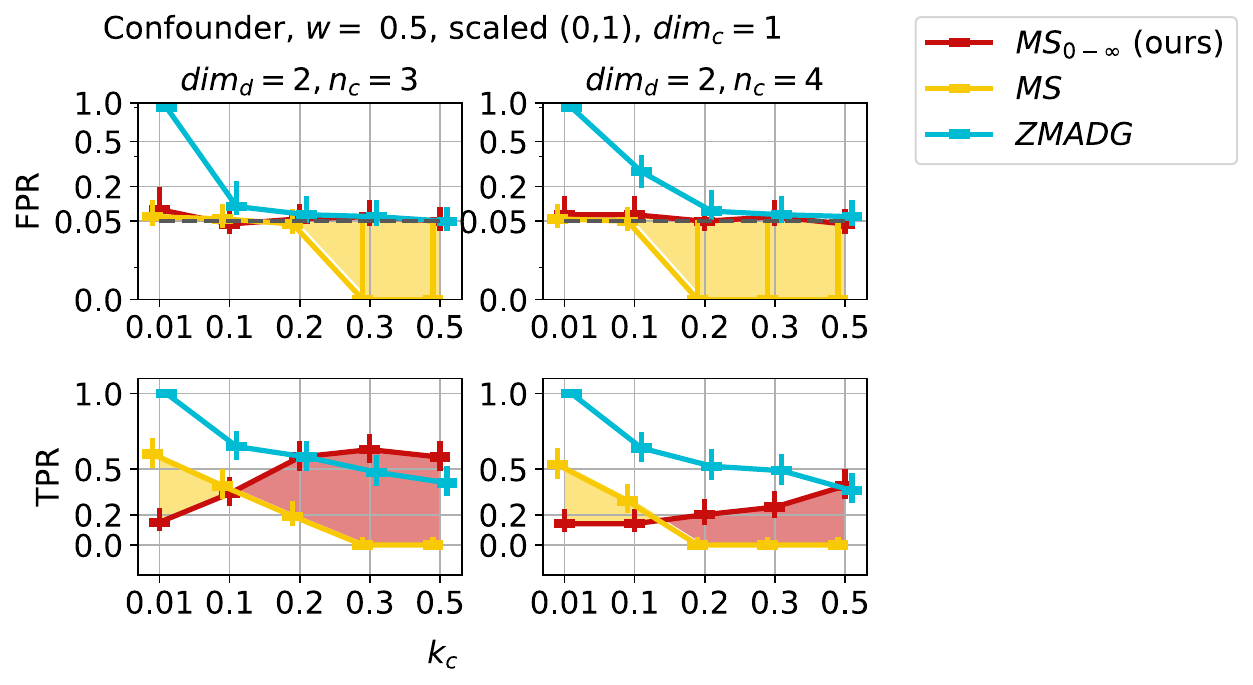}%
\\
\includegraphics[width=0.65\linewidth]{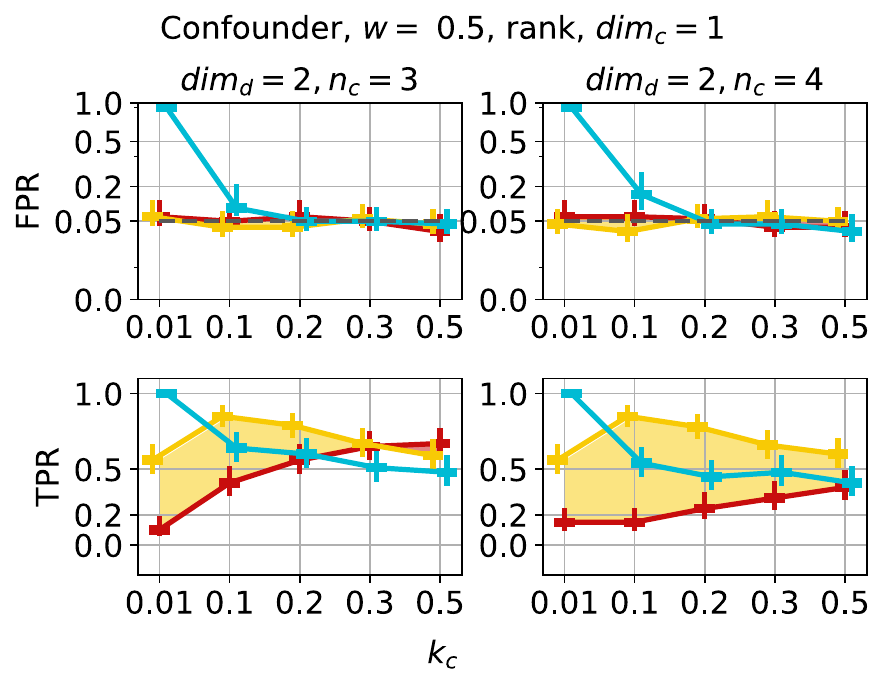}%

\caption{False positive rate (FPR, ideally under 0.05) and true positive rate (TPR, higher is better, with 1 best) for the \underline{"Confounder"}-model with $n=1000$ where $Z$ has $dim_c=1, dim_d=2$, with standardization, scaling to $(0, 1)$ and rank preprocessing for the continuous variables and coupling factor in the dependent case $w=0.5$.}
\label{fig:cf2}
\end{figure}

\begin{figure}[H]
\centering

\includegraphics[width=0.47\linewidth]{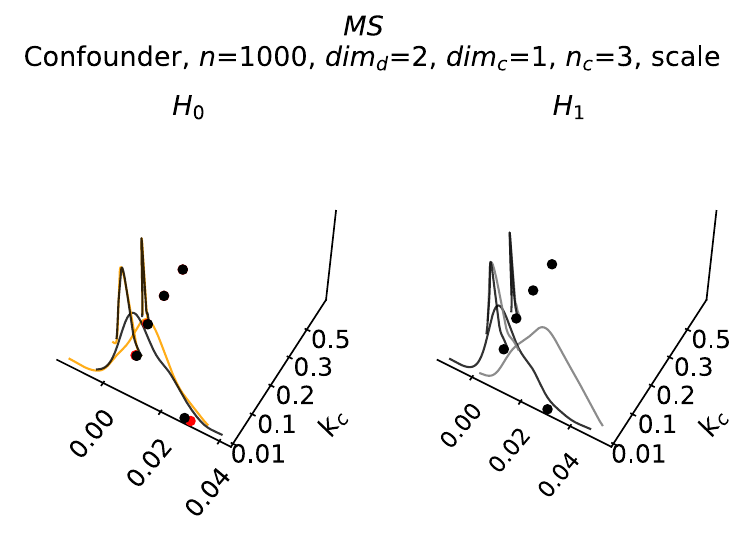}%
\includegraphics[width=0.48\linewidth]{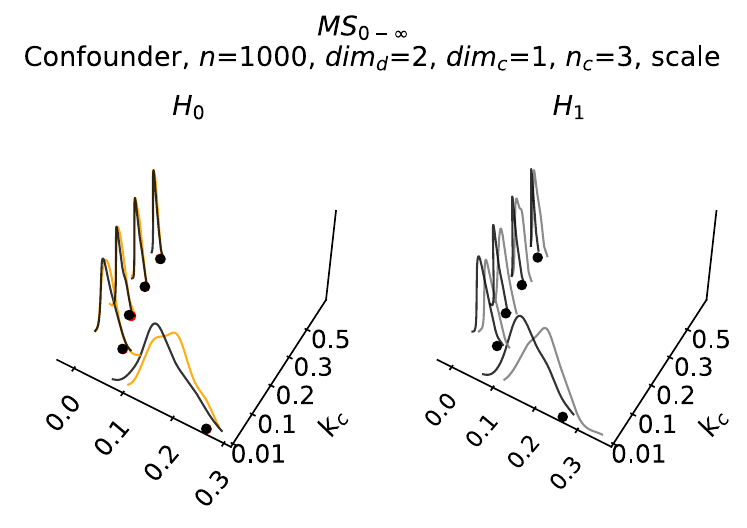}%
\\
\includegraphics[width=0.48\linewidth]{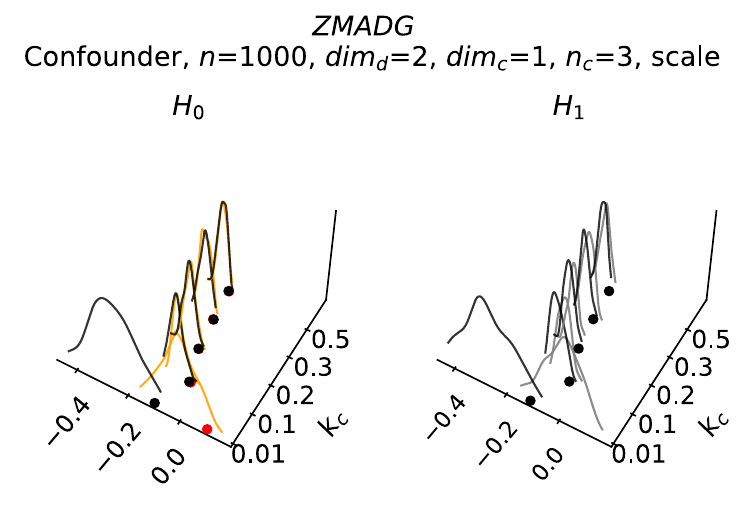}%
\\
\includegraphics[width=0.48\linewidth]{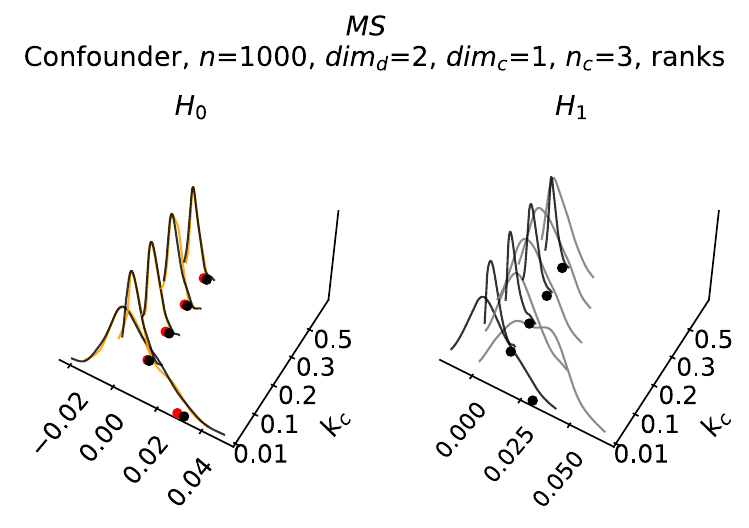}%
\includegraphics[width=0.5\linewidth]{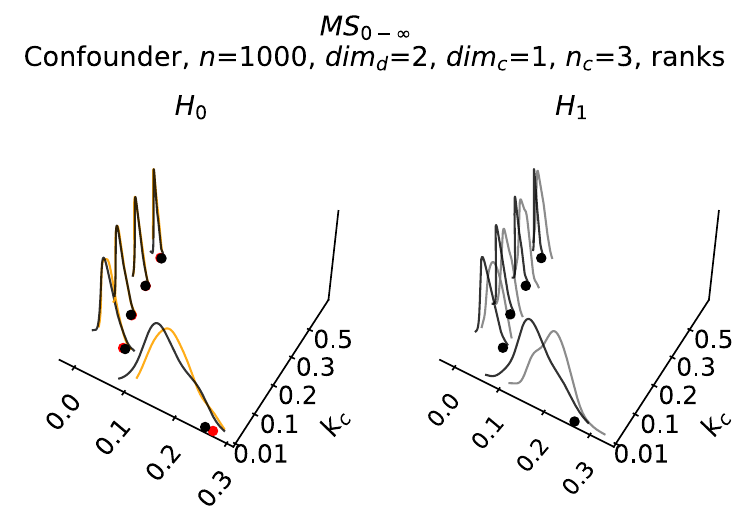}%
\\
\includegraphics[width=0.5\linewidth]{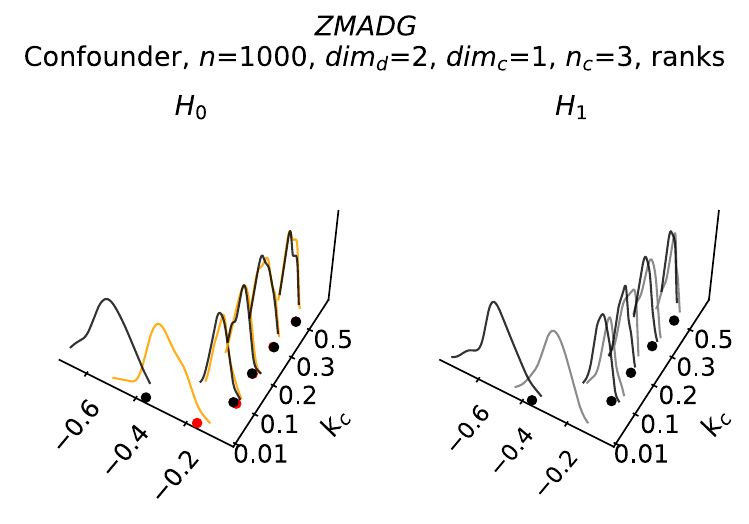}%
\caption{The true null and permuted distributions of CMI estimated with the MS, MS$_{0-\infty}$ and ZMADG estimators for the \underline{"Confounder"}-model with $n=1000$ where $Z$ has $dim_c=1, dim_d=2$ and $n_c=3$ with scaling to $(0, 1)$ and rank preprocessing. The distributions are depicted as described in Fig.~\ref{fig:dist_1}.}
\label{fig:dist_2_2}
\end{figure}

\begin{figure}[H]
\centering
\includegraphics[width=0.47\linewidth]{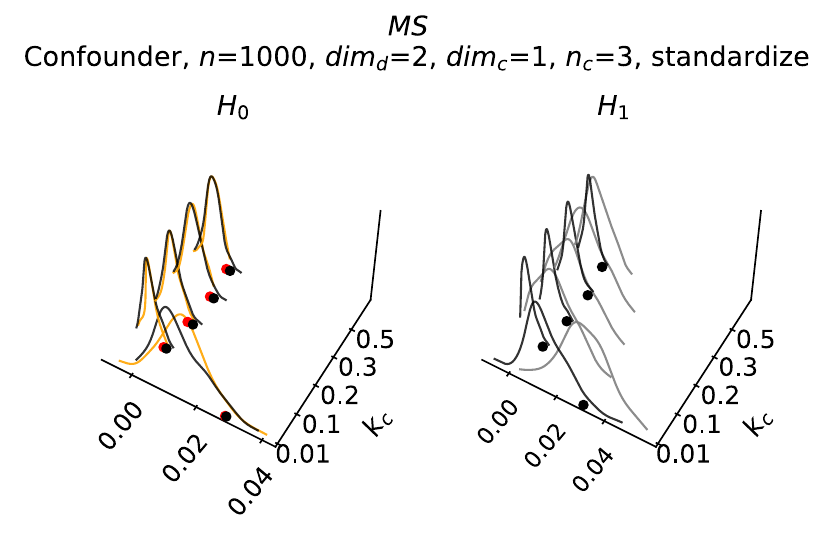}%
\includegraphics[width=0.48\linewidth]{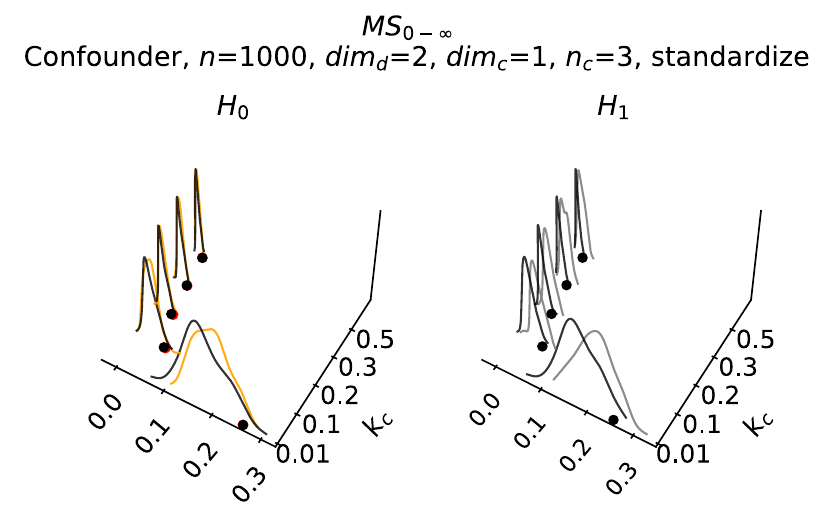}%
\\
\includegraphics[width=0.48\linewidth]{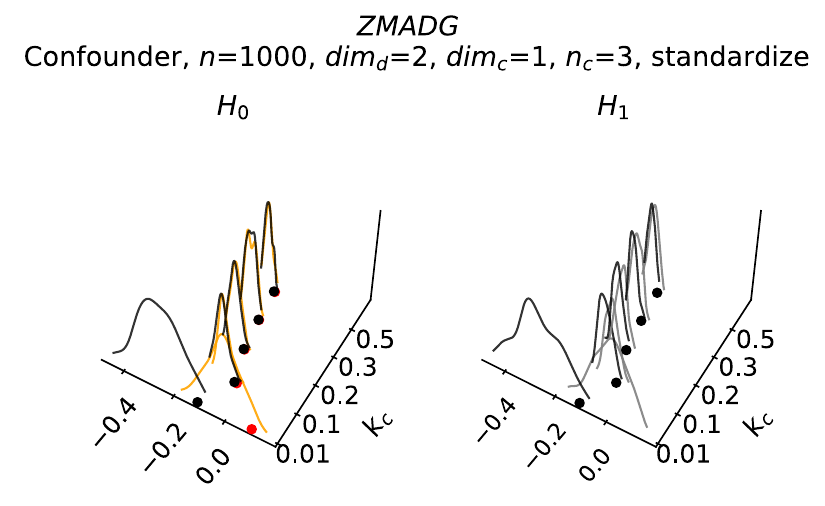}%
\caption{The true null and permuted distributions of CMI estimated with the MS, MS$_{0-\infty}$ and ZMADG estimators for the \underline{"Confounder"}-model with $n=1000$ where $Z$ has $dim_c=1, dim_d=2$ and $n_c=3$ with standardization preprocessing. The distributions are depicted as described in Fig.~\ref{fig:dist_1}.}
\label{fig:dist_2_3}
\end{figure}

\begin{figure}[H]
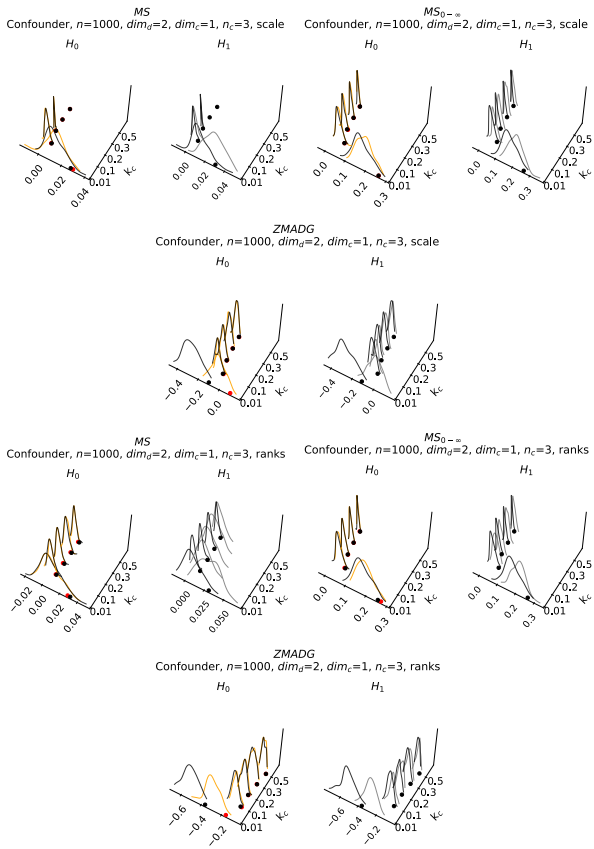

\centering
\includegraphics[width=0.47\linewidth]{figures/dists/confounder/dists_MS_1000_2_1_0.5_2_scale.pdf}%
\includegraphics[width=0.48\linewidth]{figures/dists/confounder/dists_MSinf_1000_2_1_0.5_2_scale.pdf}%
\\
\includegraphics[width=0.48\linewidth]{figures/dists/confounder/dists_ZMADG_1000_2_1_0.5_2_scale.pdf}%
\\
\includegraphics[width=0.47\linewidth]{figures/dists/confounder/dists_MS_1000_2_1_0.5_2_ranks.pdf}%
\includegraphics[width=0.48\linewidth]{figures/dists/confounder/dists_MSinf_1000_2_1_0.5_2_ranks.pdf}%
\\
\includegraphics[width=0.48\linewidth]{figures/dists/confounder/dists_ZMADG_1000_2_1_0.5_2_ranks.pdf}%
\caption{The true null and permuted distributions of CMI estimated with the MS, MS$_{0-\infty}$ and ZMADG estimators for the \underline{"Confounder"}-model with $n=1000$ where $Z$ has $dim_c=1, dim_d=2$ and $n_c=3$ with scaling to $(0, 1)$ and rank preprocessing. The distributions are depicted as described in Fig.~\ref{fig:dist_1}.}
\label{fig:dist_2_4}
\end{figure}

As the dimensionality of the discrete variable increases to $dim_d=2$ (Fig.~\ref{fig:cf2}), we observe that our approach performs better with higher $k_c$, e.g., $k_c \geq 0.3$. The CMI distribution plots (see Fig.~\ref{fig:dist_2_1}, \ref{fig:dist_2_2}, \ref{fig:dist_2_3} and \ref{fig:dist_2_4}) indicate that our approach suffers from positive bias in the case of small $k_c$ and high dimensionality. For $n_c=3$ with both standardization and rank preprocessing, our approach performs slightly worse than MS , but performs slightly better than ZMADG, which generally suffers from negative bias. For $n_c=4$, our approach performs worse than MS and slightly worse than ZMADG. Due to the definition of the data model, dependence holds in every cluster. Thus, the MS estimator performs well in the standardized and rank preprocessing cases, despite the fact that points from other clusters are considered neighbors. However, when continuous variables are scaled to $(0, 1)$, the scaling-related problems of MS lead to a rapid decline in performance as $k_c$ increases.

\begin{figure}[H]
% \centering
\includegraphics[width=0.65\linewidth]{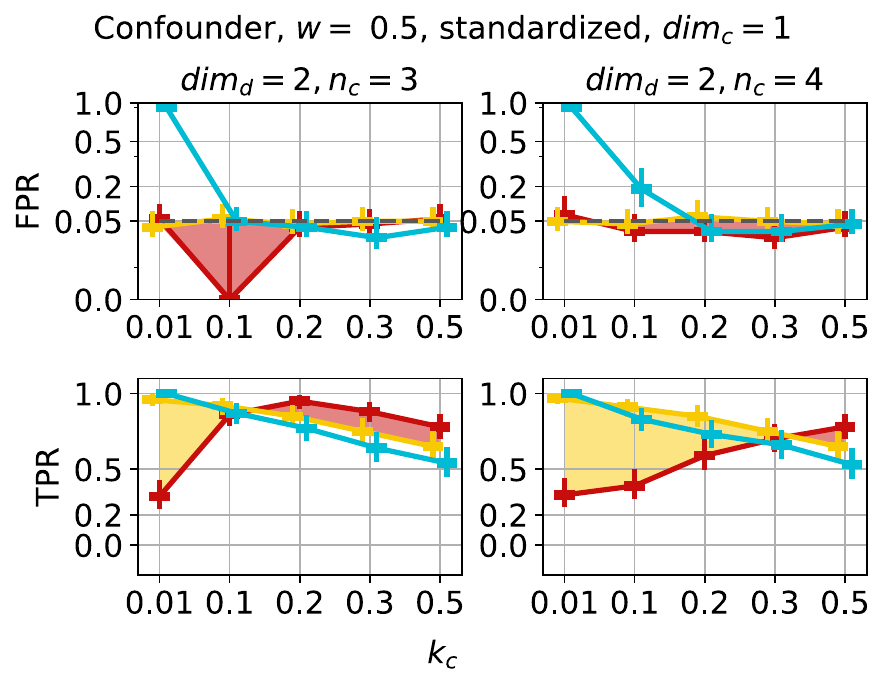}
\\ 
\includegraphics[width=0.93\linewidth]{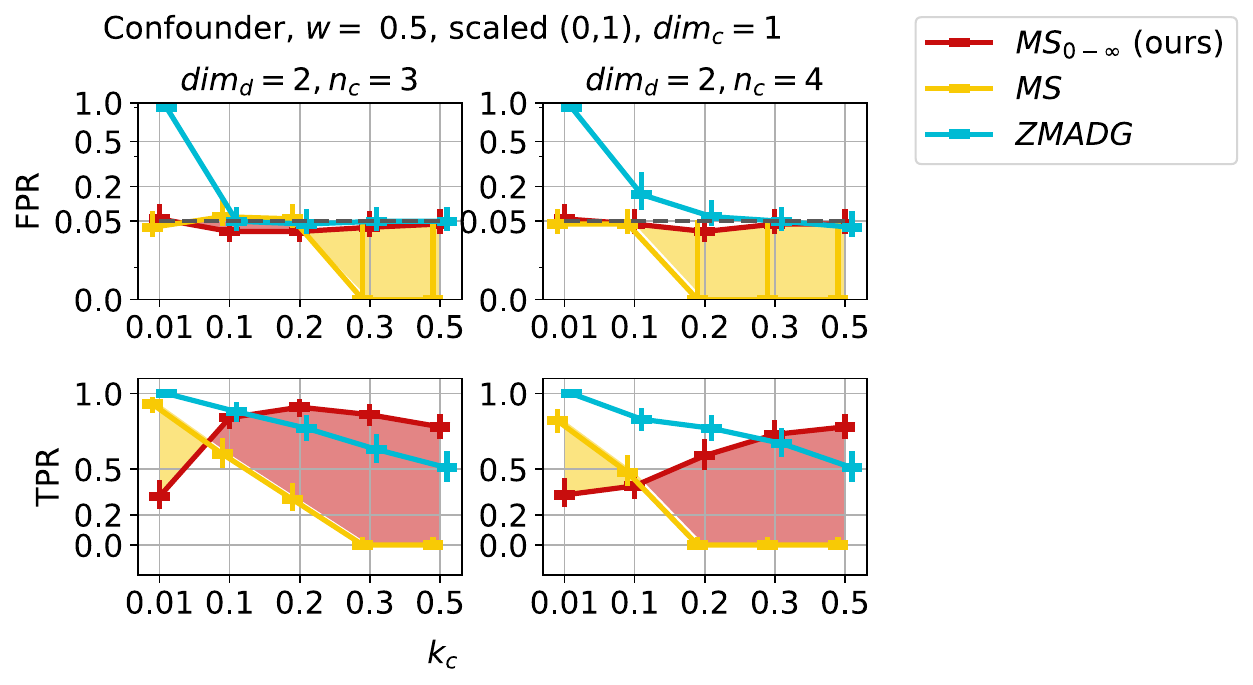}%
\\
\includegraphics[width=0.65\linewidth]{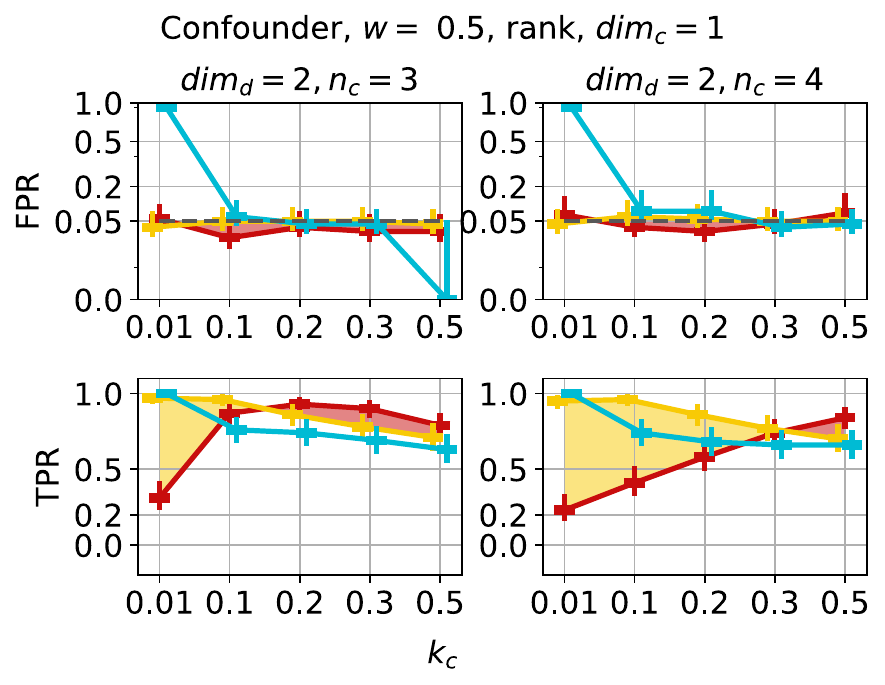}%

\caption{False positive rate (FPR, ideally under 0.05) and true positive rate (TPR, higher is better, with 1 best) for the \underline{"Confounder"}-model with $n=2000$, where $Z$ has $dim_c=1, dim_d=2$, with standardization, scaling to $(0, 1)$ and rank preprocessing for the continuous variables and coupling factor in the dependent case $w=0.5$.}
\label{fig:cf2000}
\end{figure}

Nevertheless, results for the same model with $dim_d=2$ and a larger sample size of $n=2000$ (see Fig.~\ref{fig:cf2000}) indicate that the performance of our approach considerably increases given enough samples. Noticeably, in the case of a larger sample size, a smaller $k_c=0.2$ gives the optimal results. Thus, when applying our estimator, we recommend users to consider both the number of samples and the dimensionality (i.e., the number of clusters). To exemplify, consider the case where $n=1000$ and $dim_d=2$: There are approximately $1000/(2 \cdot 3) \approx 166$ samples per cluster for $n_c=3$ and approximately $1000/(2 \cdot 4)=125$ samples for $n_c=4$. In this case, first of all, $k_c$ should be high enough such that $k$ has is large enough. Otherwise, as previously discussed in Sec.~\ref{sec:heuristic}, the estimation can suffer from high variance. Second, the number of samples per cluster should be high enough to obtain reliable results: Although our approach improves bias towards 0 compared to MS, it is still subject to the curse of dimensionality. 

\begin{figure}[H]
% \centering
\includegraphics[width=0.65\linewidth]{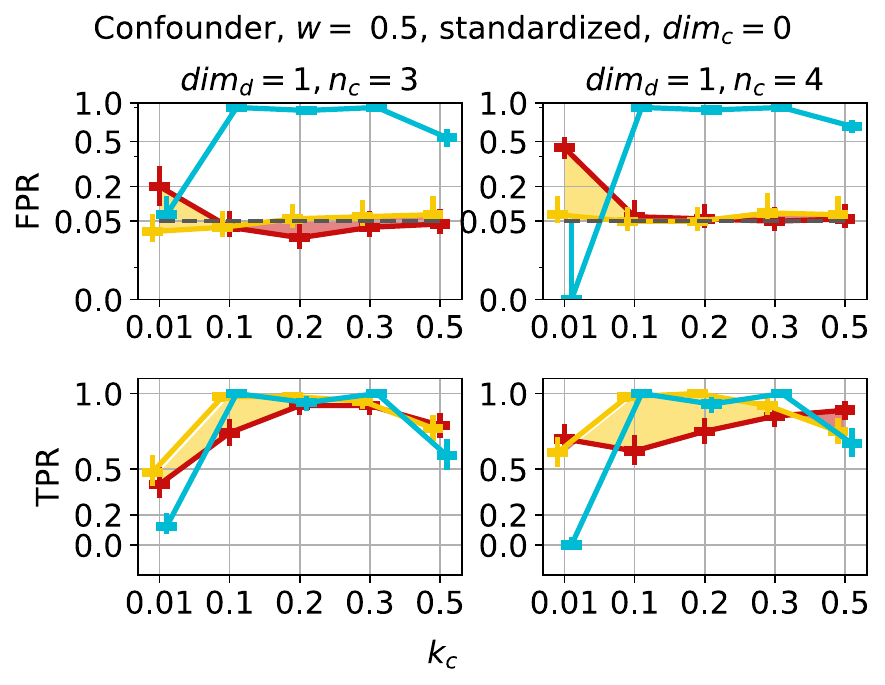}%
\\
\includegraphics[width=0.93\linewidth]{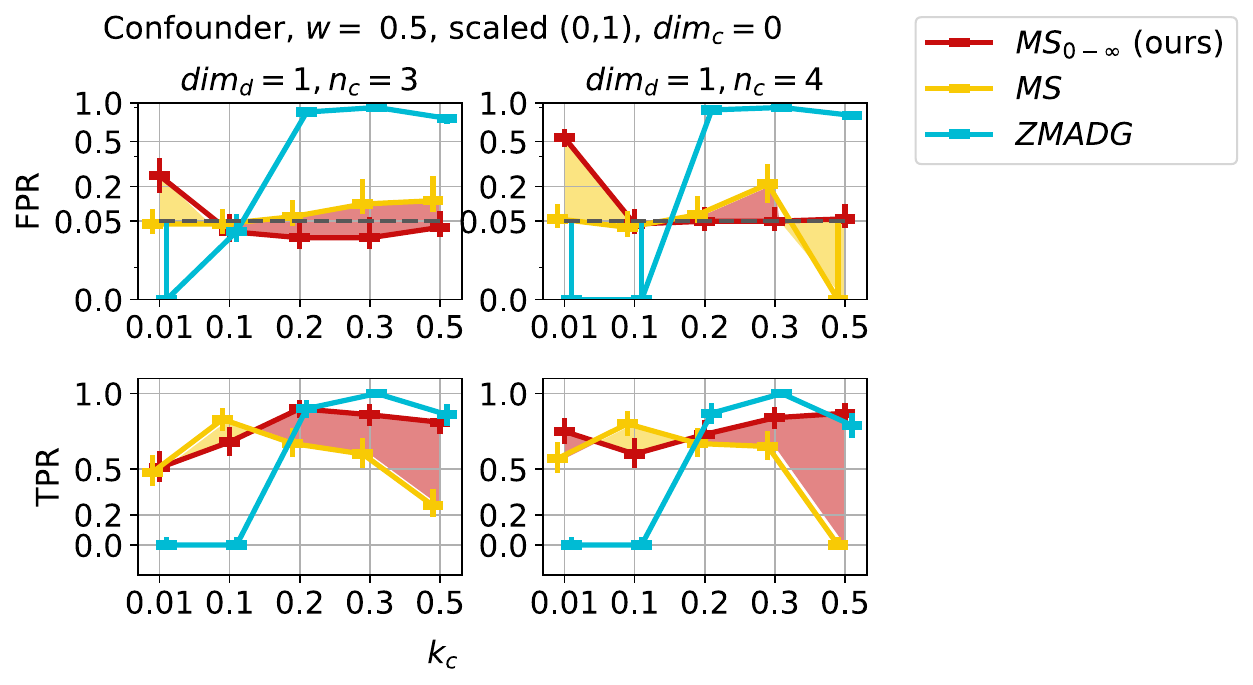}%
\\
\includegraphics[width=0.65\linewidth]{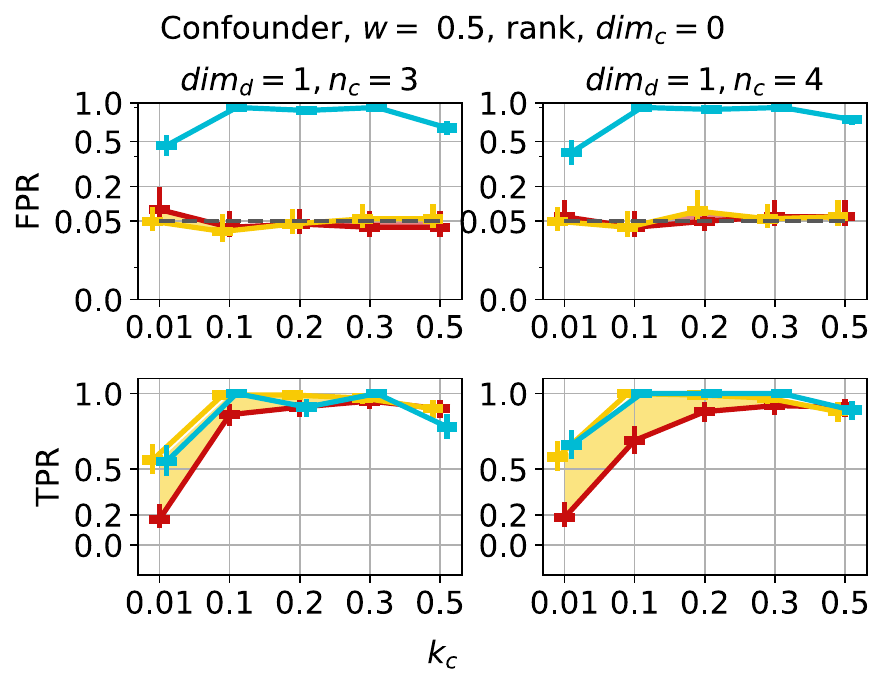}%
\\
\includegraphics[width=0.65\linewidth]{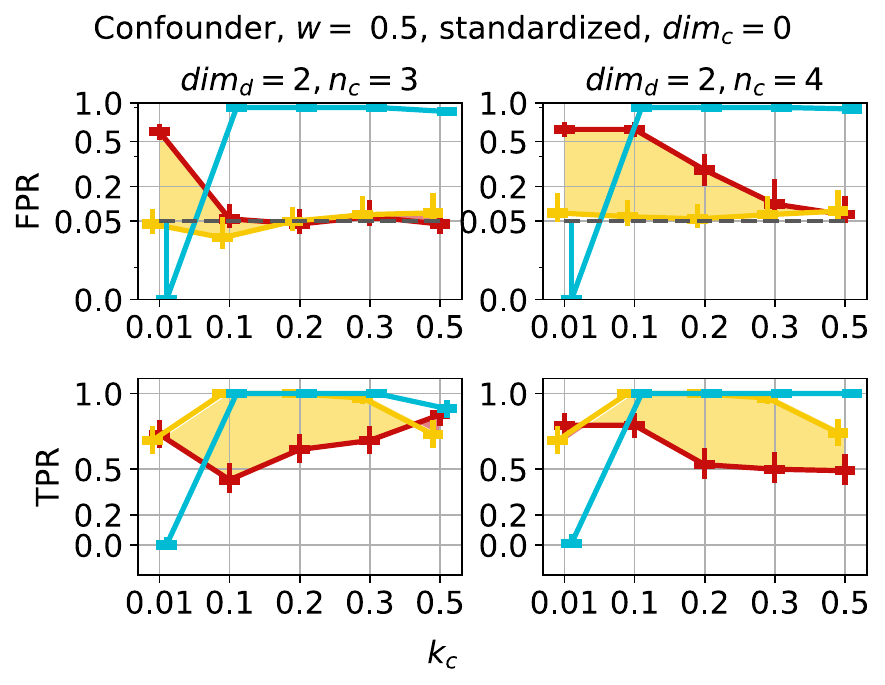}%

\caption{False positive rate (FPR, ideally under 0.05) and true positive rate (TPR, higher is better, with 1 best) for the \underline{Confounder}-model with $n=1000$ where $Z$ has $dim_c=0, dim_d=1$, with standardization, scaling to $(0, 1)$ and rank preprocessing for the continuous variables and where $Z$ has $dim_c=0, dim_d=2$ with standardization preprocessing for the continuous variables. Both configurations have coupling factor in the dependent case $w=0.5$.}
\label{fig:cf_0_2_1}
\end{figure}

\begin{figure}[H]
% \centering
% \includegraphics[width=0.65\linewidth]{figures/cit/Confounder_2_c0_T1000_c0.5_standardize.pdf}%
% \\
\includegraphics[width=0.93\linewidth]{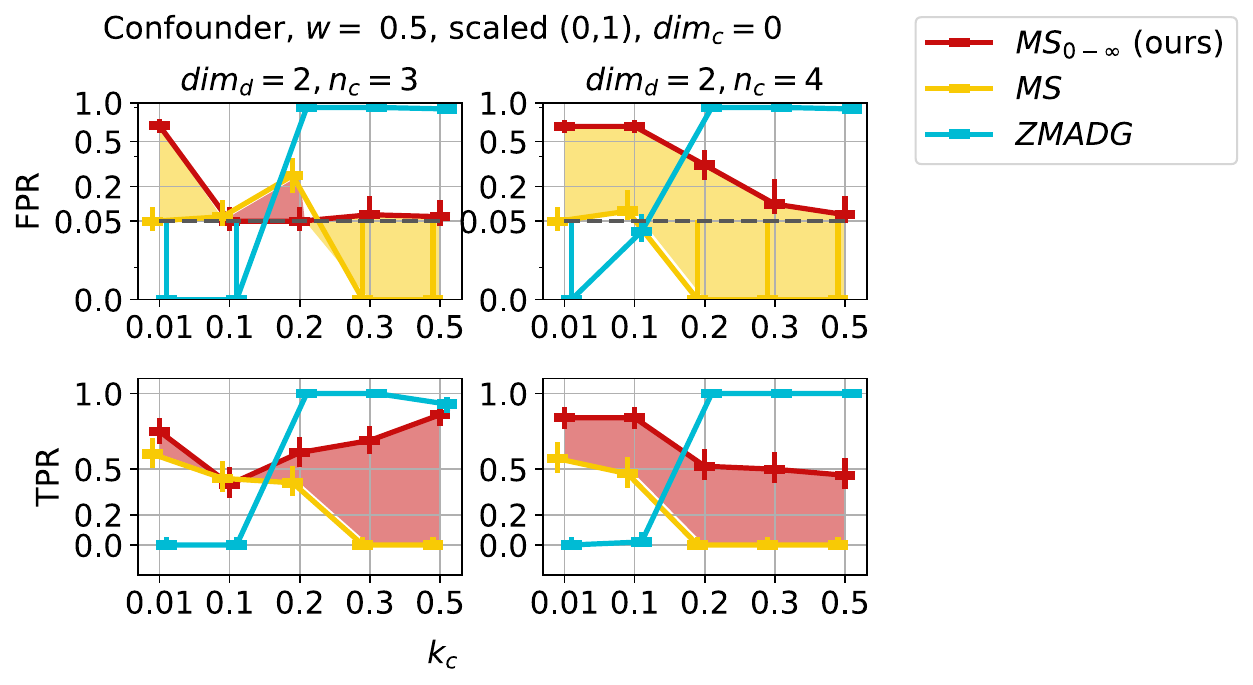}%
\\
\includegraphics[width=0.65\linewidth]{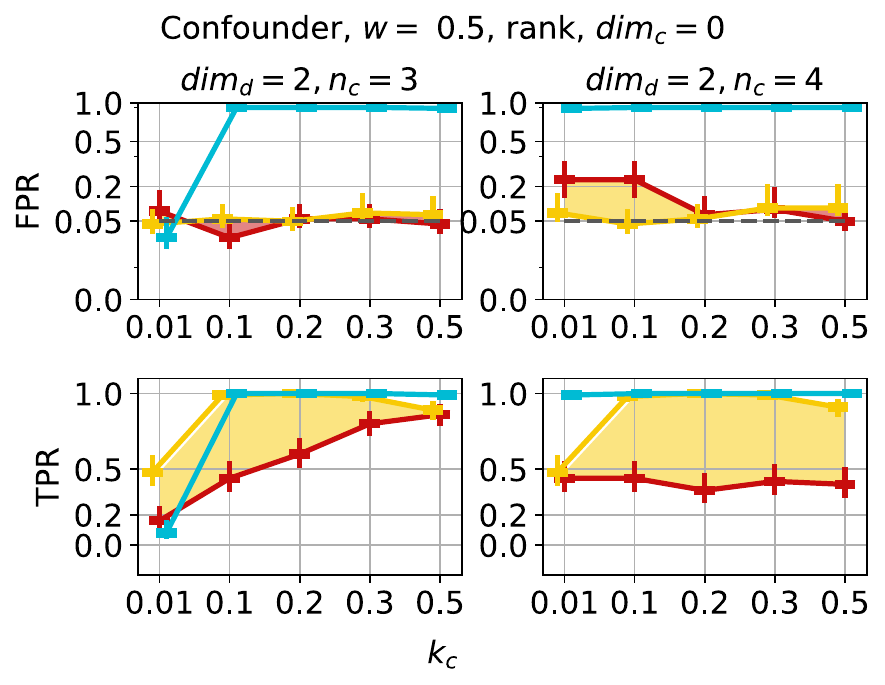}%

\caption{False positive rate (FPR, ideally under 0.05) and true positive rate (TPR, higher is better, with 1 best) for the \underline{"Confounder"}-model with $n=1000$ where $Z$ has $dim_c=0, dim_d=2$, with standardization, scaling to $(0, 1)$ and rank preprocessing for the continuous variables and coupling factor in the dependent case $w=0.5$.}
\label{fig:cf_0_2_2}
\end{figure}

\textbf{For the \underline{"Confounder"}-case where $Z$ has only discrete dimension, i.e., $dim_c=0$ and sample size $n=1000$} (Fig. \ref{fig:cf_0_2_1} and \ref{fig:cf_0_2_2}), we observe that ZMADG underperforms, either having high FPR or low TPR. MS and MS$_{0-\infty}$ perform similarly for the case when $dim_d=1$ and $dim_d=2$ and $n_c=3$ (considering their respective optimal $k_c$). Our method suffers from the curse of dimensionality for $dim_d=2$ and $n_c=4$. As always, the scaling-related problems of MS lead to underpeformance when variables are scaled to $(0, 1)$ and $k_c$ has larger values.  

\begin{figure}[H]
\centering
\includegraphics[width=0.95\linewidth]{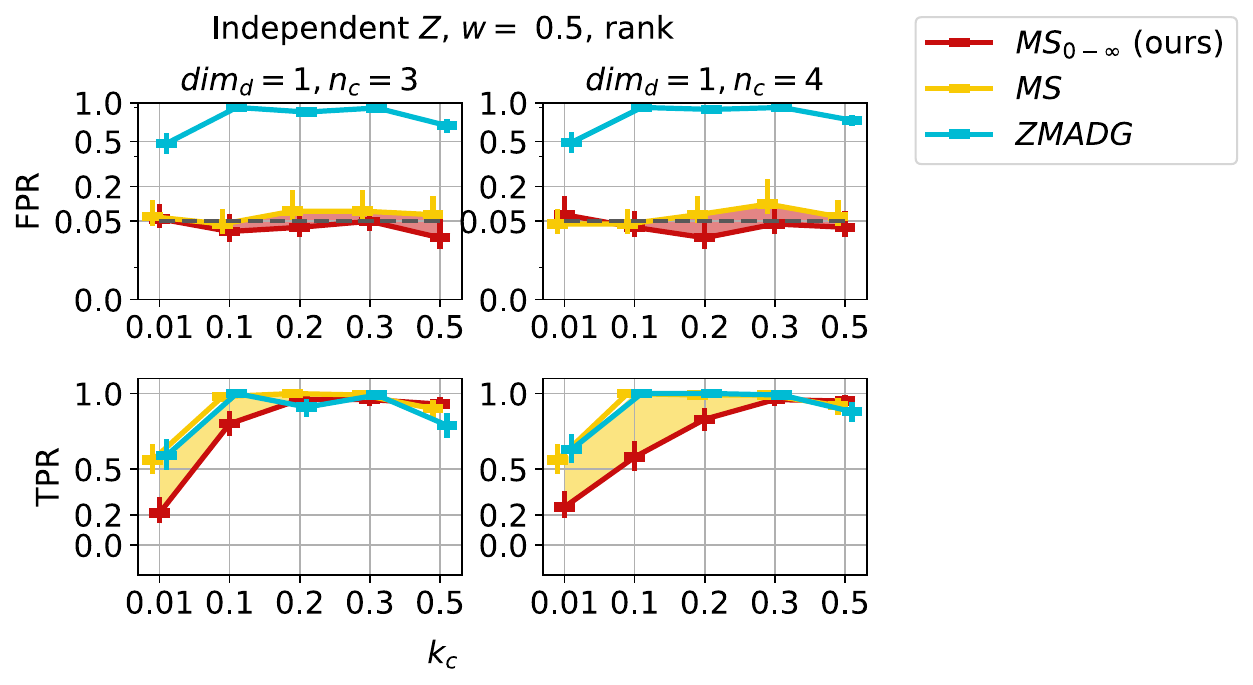}%
\caption{False positive rate (FPR, ideally under 0.05) and true positive rate (TPR, higher is better, with 1 best) for the \underline{"Independent $Z$"}-model with $n=1000$ where $Z$ has $dim_d=1$. Here we present results with rank preprocessing for the continuous variables and coupling factor in the dependent case $w=0.5$.}
\label{fig:indepz_d1}
\end{figure}

\subsection{"Independent $Z$" Model}

Here we present further results for the \underline{"Independent $Z$"}-model with $n=1000$. 

For the model with one discrete dimension, the MS and MS$_{0-\infty}$ CITs using rank preprocessing for the continuous variables (Fig.~\ref{fig:indepz_d1}) behave similarly to the CITs with standardizatio. Considering their optimal $k_c$ values, the CIT using our estimator performs similarly to the CIT using the MS estimator.

\begin{figure}[H]
\includegraphics[width=0.65\linewidth]{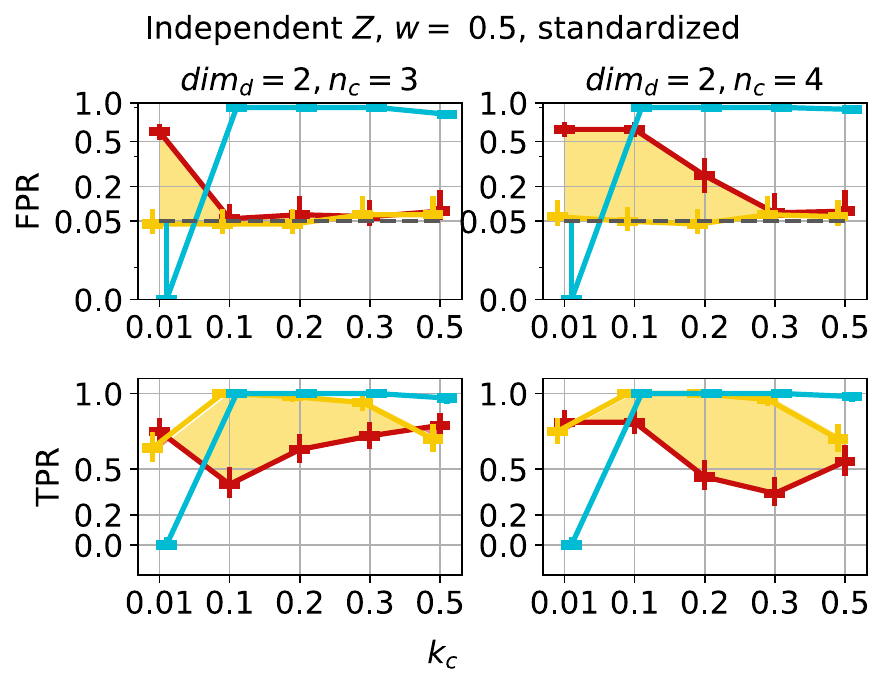}%
\\
\includegraphics[width=0.93\linewidth]{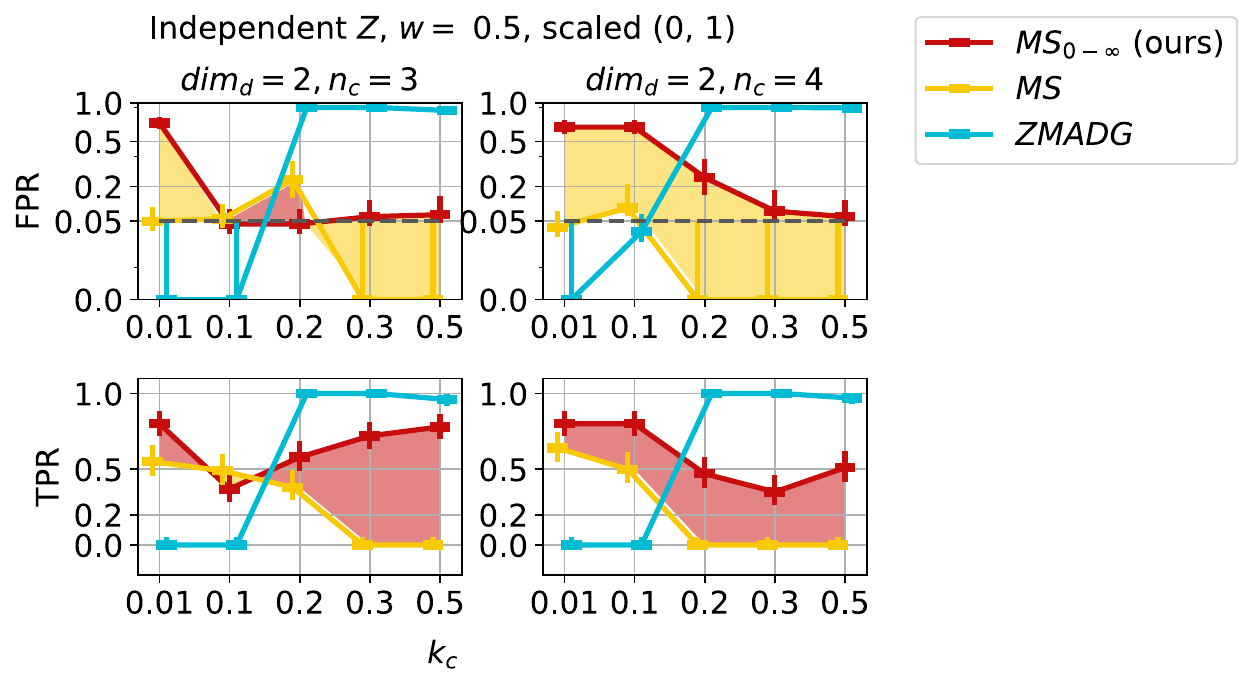}%
\\
\includegraphics[width=0.65\linewidth]{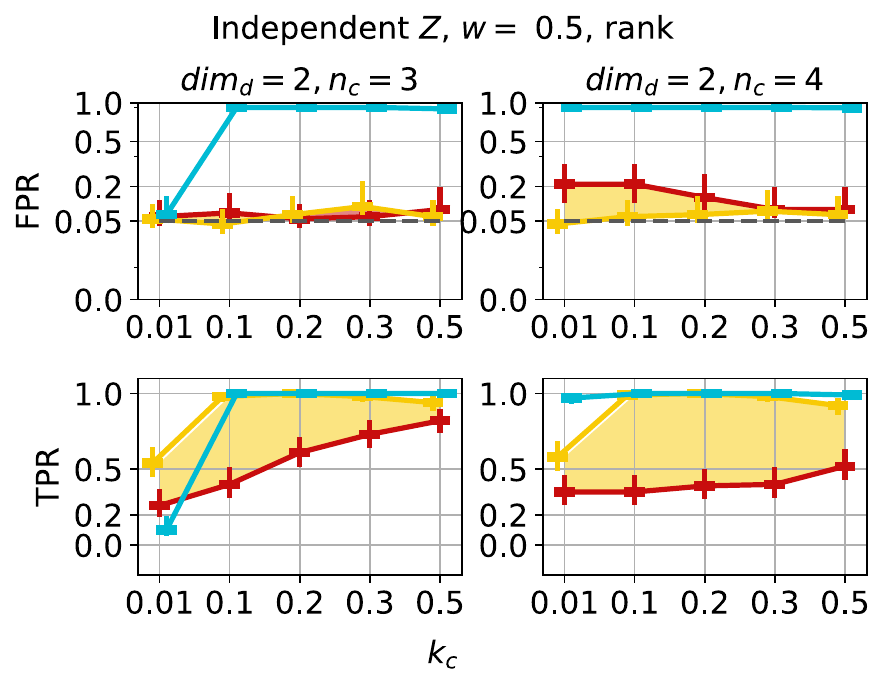}%
\caption{False positive rate (FPR, ideally under 0.05) and true positive rate (TPR, higher is better, with 1 best) for the \underline{"Independent $Z$"} model where $Z$ has $dim_d=2$, with standardization, scaling to $(0, 1)$ and rank preprocessing for the continuous variables and coupling factor in the dependent case $w=0.5$.}
\label{fig:indepz_d2}
\end{figure}

For the model with higher dimensionality (Fig.~\ref{fig:indepz_d2}), we observe that the behavior of the CITs is aligned with the behavior of the CITs for the "Confounder"-model: our MS$_{0-\infty}$ CIT has slightly lower performance than MS, and the performance gap increases for $n_c=4$ due to the curse of dimensionality. Similar to the "Confounder"-model, the "Independent $Z$"-model also has the characteristic of  dependence across clusters. Consequently, the performance of MS with standardization or rank preprocessing is not affected if neighbors are taken from distinct clusters. However, due to the scaling-related problems of MS, our estimator performs better for the scaling to $(0, 1)$ case with $dim_d=2$.

\subsection{"Cluster-dependent Confounder" Model}

\begin{figure}[H]
\centering
\includegraphics[width=\linewidth]{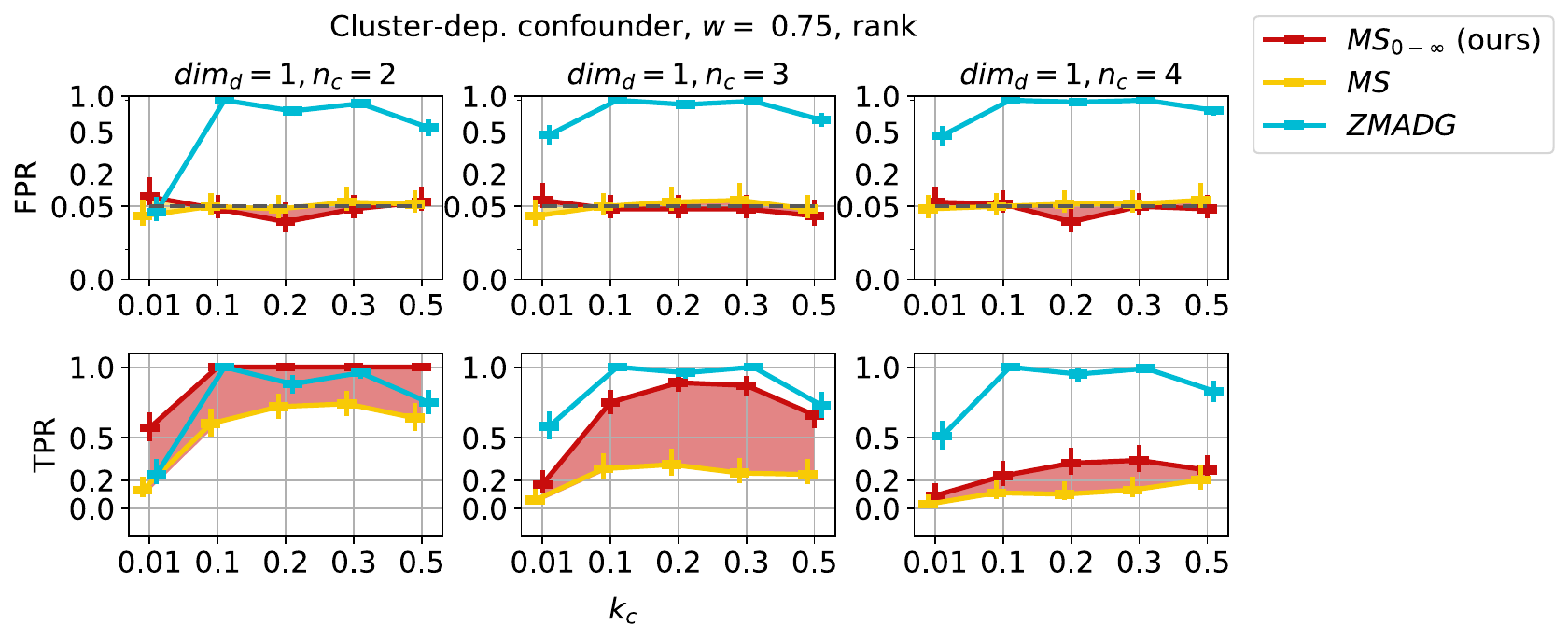}%
\caption{False positive rate (FPR, ideally under 0.05) and true positive rate (TPR, higher is better, with 1 best) for the \underline{"Cluster-dependent"}-model where $Z$ has $dim_d=1$. Here we present results with rank preprocessing for the continuous variables and coupling factor in the dependent case $w=0.75$.}
\label{fig:cluster_1}
\end{figure}

Here, we present additional results for the \underline{"Cluster-dependent Confounder"} model with $n=1000$. Consistent with the observations from previous models, the CITs with rank preprocessing (Fig.~\ref{fig:cluster_1}) behave similarly to the CITs using standardization. Nevertheless, there is a slightly more significant performance gap between MS and our MS$_{0-\infty}$, with our method showcasing superior performance, especially for $n_c=2$ and $n_c=3$.

\subsection{"Chain" Model}

\begin{figure}[H]
% \centering
\includegraphics[width=0.93\linewidth]{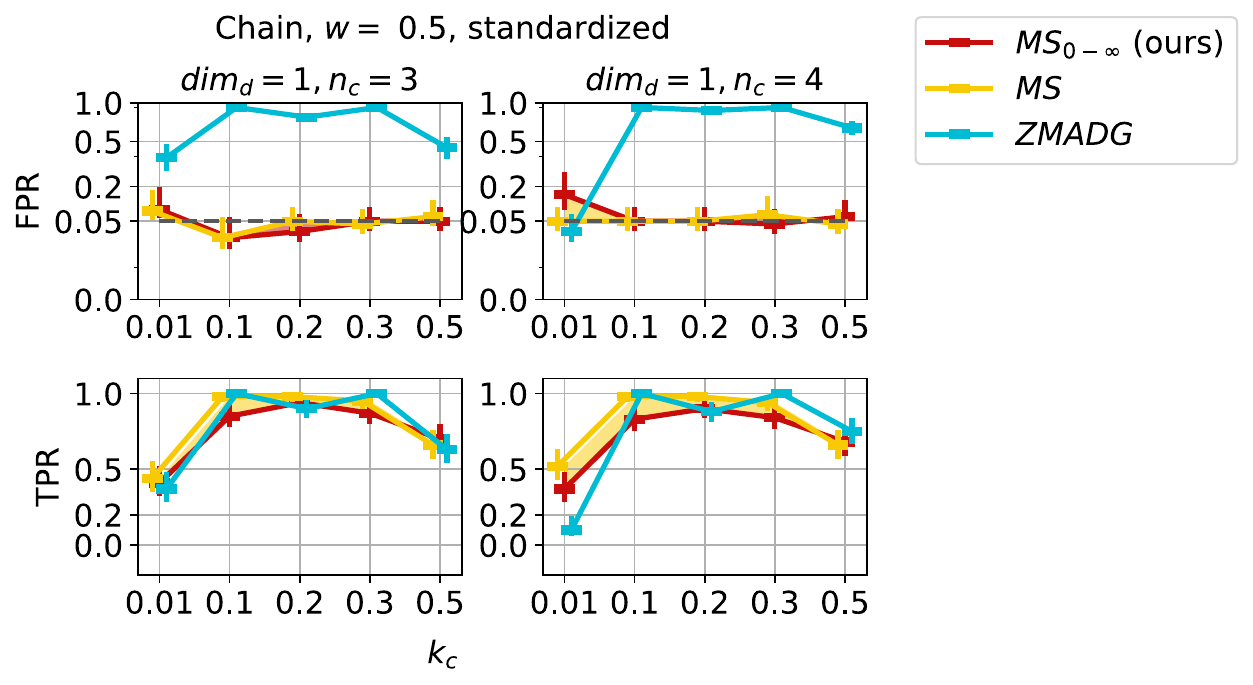}%
\caption{ False positive rate (FPR, ideally under 0.05) and true positive rate (TPR, higher is better, with 1 best) for the \underline{"Chain"}-model, with standardization preprocessing for the continuous variables and coupling factor in the dependent case $w=0.5$.}
\label{fig:chain_cit1}
\end{figure}

\begin{figure}[H]
\includegraphics[width=0.93\linewidth]{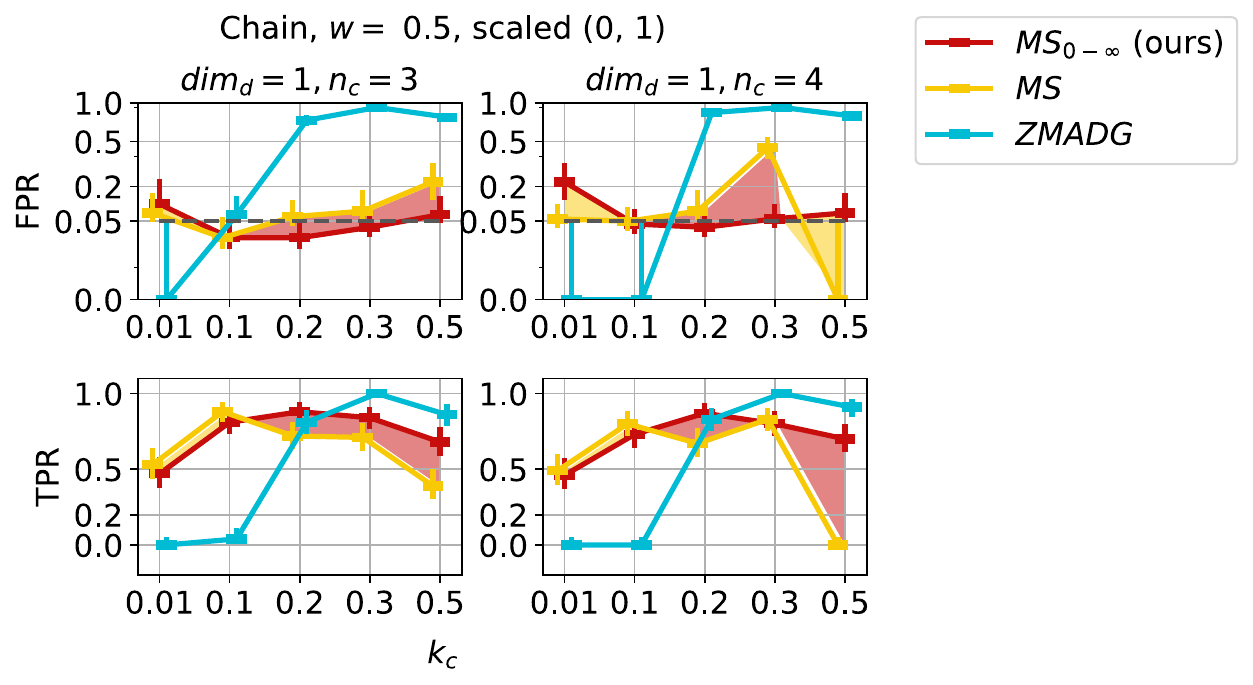}%
\\
\includegraphics[width=0.65\linewidth]
{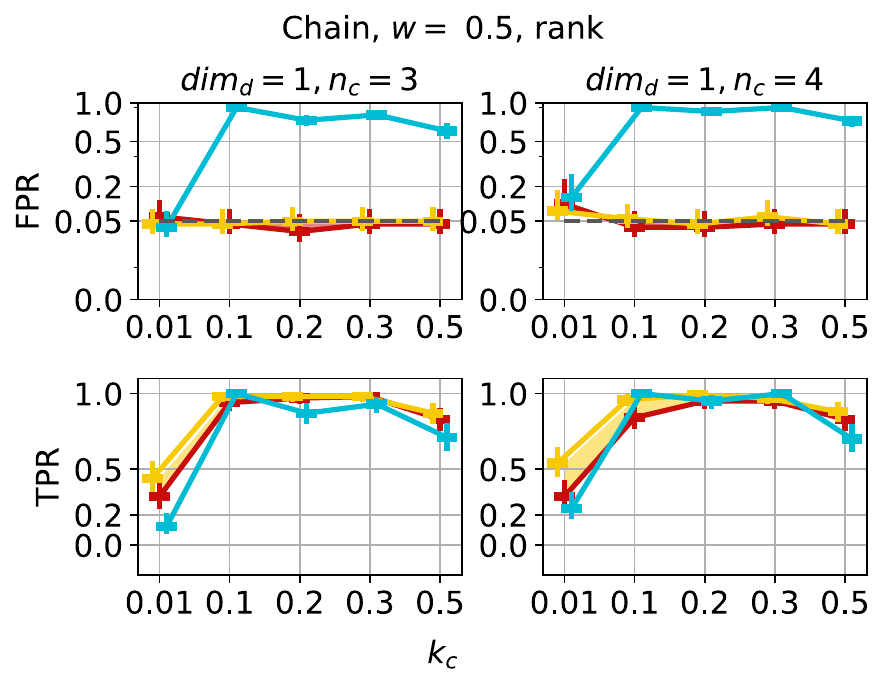}
\caption{ False positive rate (FPR, ideally under 0.05) and true positive rate (TPR, higher is better, with 1 best) for the \underline{"Chain"}-model, with scaling to $(0, 1)$ and rank preprocessing for the continuous variables and coupling factor in the dependent case $w=0.5$.}
\label{fig:chain_cit2}
\end{figure}

For the \underline{"Chain"}-model (Fig.~\ref{fig:chain_cit1} and \ref{fig:chain_cit2}), ZMADG consistently suffers from high FPR. In contrast, MS and MS$_{0{-}\infty}$ demonstrate good performance across varying $k_c$ values and dimensionalities regarding TPR. The MS and MS$_{0-\infty}$ CITs perform similarly for the standardization and rank preprocessing. In cases where $n_c=4$, our approach has slightly lower TPR. Nonetheless, both CITs control FPR effectively, except when $k_c=0.01$. The scaling-related problems of MS lead to performance issues when variables are scaled to $(0, 1)$: an increase in $k_c$ leads to elevated FPR and decreased TPR. Our approach demonstrates more robust performance across varying $k_c$ values. 

\section{Mean, Variance and Computational Time for the CMI Estimation Experiments}

% Wir testen weil I skalieret große sind und sonst lasst es sich nicht sagen dass es unterschiedlich ist, skalieren ist nicht klar wie. 
\subsection{Comparison of the Mean and Variances of the CMI}

As outlined in Sec. 3 of the main paper, our estimator addresses the challenges confronted by the MS and ZMADG estimators. Consequently, we expect the following outcomes
\begin{itemize}
    \item for some of the models, there will likely be significant differences between the bias of the MS and MS$_{0-\infty}$ estimators due to our estimator's capacity to reduce the bias towards 0, 
    \item and significant differences between the variance of the MS$_{0-\infty}$ estimator and the variance of the ZMADG estimator for most of the models. These differences arise from the definition of our estimator, which, in contrast to the ZMADG estimator, is not an aggregation of multiple estimators.
\end{itemize}

Since bias and variance are not scaled metrics, we perform statistical tests to investigate whether there are statistically significant differences in the bias and variance among the various CMI estimators when measured on the models presented in Sec.4 of the main paper and Sec.~\ref{app:further_Estim} of the SM. However, it is not immediately apparent how exactly to conduct these tests. One possible approach would be to compare the three estimators pairwise per $k_c$ value and sample size. However, the MS and ZMADG estimators perform very differently across $k_c$ values; thus, it would not make for a fair comparison. 

Hence, we conduct statistical tests to compare the estimators' bias and variances as follows: For each model, we identify the $k_c$ value with optimal performance regarding bias and variance for the $100$ repetitions of the experiment across all sample sizes. Subsequently, to test our hypothesis, we compare the bias of the MS estimate versus the bias of the MS$_{0-\infty}$ estimate and the variance of the MS$_{0-\infty}$ versus the variance of the ZMADG estimator.

To compare the bias, for each data model and each estimator $estim \in \{MS, MS_{0-\infty}\}$, we compute the mean absolute error (in short, MAE) $MAE_{estim, j}$ for each repetition of the experiment $j \in \{1, \ldots, 100\}$ as the absolute difference between the estimated CMI using estimator $estim$ and the actual ground truth CMI value $I(X; Y|Z)$ corresponding to the given data model:

\begin{equation}
    MAE_{estim,j} = |\hat{I}^{estim}_j(X;Y|Z) - I_j(X;Y|Z)|.
\end{equation}

We then apply the Wilcoxon-Signed-Rank Test \cite{Wilcoxon1945IndividualCB} to compare bias between estimators. As previously stated, we expect that for some of the data models, the MAE of the MS estimator is higher than the MAE of our MS$_{0-\infty}$ estimator. We thus perform one-tailed tests. For each model and each sample size, we formulate the hypothesis for the comparison of the two estimators as follows: 

\textit{
$H_0$: The median of the difference between the $MAE_{MS}$ and $MAE_{MS_{0-\infty}}$ is negative.}

Vs.

\textit{$H_1$:  The median of the difference between the $MAE_{MS}$ and $MAE_{MS_{0-\infty}}$ is non-negative.
}

To compare the variance of the $100$ estimates $\hat{I}^{estim}_j(X;Y|Z)$ obtained with the MS$_{0-\infty}$ and ZMADG estimators, we test for equality of variance using Levene's Test \cite{levene}. All performed tests are two-tailed tests. Thus, for each model and each sample size, we test the following hypothesis: 

\textit{$H_0$: The  variance of the MS$_{0-\infty}$ estimator and the variance of the ZMADG estimator are equal.}

Vs.

\textit{$H_1$: The variance of the MS$_{0-\infty}$ estimator and the variance of the ZMADG estimator are not equal.}

To account for repeated testing, we apply Bonferroni correction \cite{Bonferroni1935IlCD} by splitting the significance level by the number of hypotheses, in our case $48$, and thus reject the null hypothesis for p-values under $\frac{0.05}{48}=0.001$.

In the Tables \ref{tab:first_sig} to \ref{tab:last_sig} below, we report the obtained p-values for the statistical tests for each model, measurement, and sample size. We observe that the MAE of the MS estimator is significantly greater than the MAE of our MS$_{0-\infty}$ estimator for the "Independent $Z$" with $d=3$ and the "Confounder with uniform $X$ and $Y$" models . This aligns with our expectations, as we have hypothesized in the Sec. 3 of the main paper: Our estimator should suffer from less bias towards $0$. Furthermore, for almost all models and all sample sizes, the hypothesis of equality of variances of our estimator and the ZMADG estimator can be rejected, as expected and discussed in Sec. 3 of the main paper. 

\begin{table}[H]
\resizebox{\linewidth}{!}{%
\begin{tabular}{ccc}
\multicolumn{3}{c}{\textbf{p-values for the "Independent $Z$" Model with $d=1$}}             \\
\textbf{$n$} & \textbf{Bias MS vs. MS$_{0-\infty}$} & \textbf{Var MS$_{0-\infty}$ vs. ZMADG} \\
300  & 0.9993 & 0.0000 \\
600  & 1.0000    & 0.0000 \\
1000 & 1.0000   & 0.0000 \\
2000 & 1.0000    & 0.0000
\end{tabular}%
}
\caption{Results of the statistical tests for the "Independent $Z$" model with $d=1$. We select $k_c$ for the individual estimators as follows: $k_{c, MS}=0.01$, $k_{c, MS_{0-\infty}}=0.2$ and $k_{c, ZMADG}=0.1$.}
\label{tab:first_sig}
\end{table}

\begin{table}[H]
\resizebox{\linewidth}{!}{%
\begin{tabular}{ccc}
\multicolumn{3}{c}{\textbf{p-values for the "Independent $Z$" Model with $d=3$}}             \\
\textbf{$n$} & \textbf{Bias MS vs. MS$_{0-\infty}$} & \textbf{Var MS$_{0-\infty}$ vs. ZMADG} \\
300  & 0.0146 & 0.0000 \\
600  & 0.0000    & 0.0000 \\
1000 & 0.0000    & 0.0000 \\
2000 & 0.0000    & 0.0000
\end{tabular}%
}
\caption{Results of the statistical tests for the "Independent $Z$" model with $d=3$. We select $k_c$ for the individual estimators as follows: $k_{c, MS}=0.01$, $k_{c, MS_{0-\infty}}=0.2$ and $k_{c, ZMADG}=0.1$.}
\end{table}

\begin{table}[H]
\resizebox{\linewidth}{!}{%
\begin{tabular}{ccc}
\multicolumn{3}{c}{\textbf{p-values for the "Chain structure" Model with $d=1$}}             \\
\textbf{$n$} & \textbf{Bias MS vs. MS$_{0-\infty}$} & \textbf{Var MS$_{0-\infty}$ vs. ZMADG} \\
300  & 1.0000 & 0.0000 \\
600  & 1.0000 & 0.0000 \\
1000 & 1.0000 & 0.0000 \\
2000 & 1.0000 & 0.0000
\end{tabular}%
}
\caption{Results of the statistical tests for the "Chain structure" model with $d=1$. We select $k_c$ for the individual estimators as follows: $k_{c, MS}=0.01$, $k_{c, MS_{0-\infty}}=0.2$ and $k_{c, ZMADG}=0.2$.}
\end{table}

\begin{table}[H]
\resizebox{\linewidth}{!}{%
\begin{tabular}{ccc}
\multicolumn{3}{c}{\textbf{p-values for the "Chain structure" Model with $d=3$}}             \\
\textbf{$n$} & \textbf{Bias MS vs. MS$_{0-\infty}$} & \textbf{Var MS$_{0-\infty}$ vs. ZMADG} \\
300  & 0.9996 & 0.0000 \\
600  & 1.0000 & 0.0000 \\
1000 & 1.0000 & 0.0000 \\
2000 & 1.0000 & 0.0000
\end{tabular}%
}
\caption{Results of the statistical tests for the "Chain structure" model with $d=3$. We select $k_c$ for the individual estimators as follows: $k_{c, MS}=0.01$, $k_{c, MS_{0-\infty}}=0.2$ and $k_{c, ZMADG}=0.2$.}
\end{table}

\begin{table}[H]
\resizebox{\linewidth}{!}{%
\begin{tabular}{ccc}
\multicolumn{3}{c}{\textbf{p-values for the "Confounder with uniform $X$, $Y$" Model}}       \\
\textbf{$n$} & \textbf{Mean MS vs. MS$_{0-\infty}$} & \textbf{Var MS$_{0-\infty}$ vs. ZMADG} \\
300  & 0.0000 & 0.0000 \\
600  & 0.0000 & 0.0000 \\
1000 & 0.0000 & 0.0000 \\
2000 & 0.0000 & 0.0000
\end{tabular}%
}
\caption{Results of the statistical tests for the "Confounder with uniform $X$ and $Y$" model. We select $k_c$ for the individual estimators as follows: $k_{c, MS}=0.1$, $k_{c, MS_{0-\infty}}=0.1$ and $k_{c, ZMADG}=0.3$.}
\end{table}

\begin{table}[H]
\resizebox{\linewidth}{!}{%
\begin{tabular}{ccc}
\multicolumn{3}{c}{\textbf{p-values for the "Confounder with Gaussian $X, Y$" Model}}   \\
\textbf{$n$} & \textbf{Bias MS vs. MS$_{0-\infty}$} & \textbf{Var MS$_{0-\infty}$ vs. ZMADG} \\
300  & 0.5000 & 0.0000 \\
600  & 0.5000 & 0.0000 \\
1000 & 0.5000 & 0.0000 \\
2000 & 0.5000 & 0.0000
\end{tabular}%
}
\caption{Results of the statistical tests for the "Confounder with Gaussian $X$ and $Y$" model. We select $k_c$ for the individual estimators as follows: $k_{c, MS}=0.01$, $k_{c, MS_{0-\infty}}=0.3$ and $k_{c, ZMADG}=0.1$.}
\label{tab:last_sig}
\end{table}

\subsection{Report on Computational Runtimes}

Here, we report the computational runtimes for each model in the CMI estimation experiments,. The experiments were performed on an Intel(R) Core(TM) i7-6600U CPU. We recorded the runtimes for each run of the $100$ repetitions of the CMI experiments using the different $k_c$ values and sample sizes $n$. In the Tables \ref{tab:comp_first} to \ref{tab:comp_last}, we present the average runtime for each sample size $n$ in seconds, averaged over the $100$ runs and all $k_c$ values. We also include the average runtime of the different heuristics for our approach: "local", "global", and "cluster". We note that in the tables below, MS$_{0-\infty}$ stands for the MS$_{0-\infty}$ in combination with the "MS$_{0-\infty}$ local" heuristic for our estimator in combination with the "local" heuristic, and "MS$_{0-\infty}$ 
 cluster" stands for our estimator in combination with the "cluster-size" heuristic.

\begin{table}[H]
\resizebox{\linewidth}{!}{%
\begin{tabular}{cccccc}
\textbf{$n$} & \textbf{MS$_{0-\infty}$} & \textbf{MS$_{0-\infty}$ global} & \textbf{MS$_{0-\infty}$ cluster} & \textbf{MS} & \textbf{ZMADG} \\
300  & 0.013 & 0.016 & 0.013 & 0.029 & 0.036 \\
600  & 0.033 & 0.023 & 0.034 & 0.03  & 0.046 \\
1000 & 0.069 & 0.046 & 0.076 & 0.05  & 0.046 \\
2000 & 0.249 & 0.142 & 0.271 & 0.143 & 0.05 
\end{tabular}%
}
\caption{Computational runtimes (in seconds) for the "Independent $Z$" model (Sec. 4 of the main paper) with $d=1$. }
\label{tab:comp_first}
\end{table}

\begin{table}[H]
\resizebox{\linewidth}{!}{%
\begin{tabular}{cccccc}
\textbf{$n$} & \textbf{MS$_{0-\infty}$} & \textbf{MS$_{0-\infty}$ global} & \textbf{MS$_{0-\infty}$ cluster} & \textbf{MS} & \textbf{ZMADG} \\
300  & 0.021 & 0.018 & 0.02  & 0.025 & 0.059 \\
600  & 0.052 & 0.043 & 0.051 & 0.037 & 0.087 \\
1000 & 0.107 & 0.081 & 0.111 & 0.061 & 0.118 \\
2000 & 0.327 & 0.215 & 0.333 & 0.165 & 0.164
\end{tabular}%
}
\caption{Computational runtimes (in seconds) for the "Independent $Z$" model (Sec. 4 of the main paper) with $d=3$. }
\end{table}

\begin{table}[H]
\resizebox{\linewidth}{!}{%
\begin{tabular}{cccccc}
\textbf{$n$} & \textbf{MS$_{0-\infty}$} & \textbf{MS$_{0-\infty}$ global} & \textbf{MS$_{0-\infty}$ cluster} & \textbf{MS} & \textbf{ZMADG} \\
300  & 0.012 & 0.015 & 0.012 & 0.029 & 0.014 \\
600  & 0.036 & 0.023 & 0.035 & 0.033 & 0.021 \\
1000 & 0.082 & 0.05  & 0.08  & 0.053 & 0.027 \\
2000 & 0.332 & 0.171 & 0.306 & 0.136 & 0.039
\end{tabular}%
}
\caption{Computational runtimes (in seconds) for the "Chain structure" model (Sec.~\ref{app:further_Estim}) with $d=1$. }
\end{table}

\begin{table}[H]
\resizebox{\linewidth}{!}{%
\begin{tabular}{lllllll}
  & $n$  & MS$_{0-\infty}$ & MS$_{0-\infty}$ global & MS$_{0-\infty}$ cluster & MS    & ZMADG \\
0 & 300  & 0.019           & 0.011                  & 0.019                   & 0.024 & 0.027 \\
1 & 600  & 0.057           & 0.024                  & 0.055                   & 0.035 & 0.041 \\
2 & 1000 & 0.116           & 0.041                  & 0.113                   & 0.052 & 0.054 \\
3 & 2000 & 0.433           & 0.097                  & 0.399                   & 0.131 & 0.079
\end{tabular}%
}
\caption{Computationals runtime (in seconds) for the "Chain structure" model (Sec.~\ref{app:further_Estim}) with $d=3$. }
\end{table}

\begin{table}[H]
\resizebox{\linewidth}{!}{%
\begin{tabular}{cccccc}
\textbf{$n$} & \textbf{MS$_{0-\infty}$} & \textbf{MS$_{0-\infty}$ global} & \textbf{MS$_{0-\infty}$ cluster} & \textbf{MS} & \textbf{ZMADG} \\
300  & 0.01  & 0.012 & 0.01  & 0.025 & 0.023 \\
600  & 0.03  & 0.023 & 0.029 & 0.027 & 0.023 \\ 
1000 & 0.068 & 0.051 & 0.068 & 0.042 & 0.026 \\
2000 & 0.253 & 0.183 & 0.252 & 0.115 & 0.037 \\
\end{tabular}%
}
\caption{Computational runtimes (in seconds) for the "Confounder with uniform $X$ and $Y$" model (Sec.~\ref{app:further_Estim}).}
\end{table}

\begin{table}[H]
\resizebox{\linewidth}{!}{%
\begin{tabular}{cccccc}
\textbf{$n$} & \textbf{MS$_{0-\infty}$} & \textbf{MS$_{0-\infty}$ global} & \textbf{MS$_{0-\infty}$ cluster} & \textbf{MS} & \textbf{ZMADG} \\
300  & 0.009 & 0.013 & 0.009 & 0.027 & 0.022 \\
600  & 0.027 & 0.021 & 0.026 & 0.029 & 0.023 \\
1000 & 0.06  & 0.046 & 0.059 & 0.043 & 0.028 \\
2000 & 0.221 & 0.156 & 0.216 & 0.098 & 0.04 
\end{tabular}%
}
\caption{Computational runtimes (in seconds) for the "Confounder with Gaussian $X$ and $Y$" model (Sec.~\ref{app:further_Estim}).}
\label{tab:comp_last}
\end{table}

\section{Remarks on Reproducibility}

We intentionally reduced the number of plots and tables in this supplementary material for length reasons. However, all code to obtain the CIT plots and the measurements for the statistical significance tests, as well as the computational time reports can be found in the .zip file accompanying this supplementary, including the details necessary for replicating our experiments, such as random seeds.

\end{document}